\documentclass{article}

\PassOptionsToPackage{numbers, compress, sort}{natbib}

 \usepackage[preprint]{neurips_2026}

\usepackage[utf8]{inputenc} 
\usepackage[T1]{fontenc}    

\usepackage{url}            
\usepackage{booktabs}       
\usepackage{amsfonts}       
\usepackage{nicefrac}       
\usepackage{microtype}      
\usepackage{xcolor}         
\usepackage{multirow}
\usepackage{algorithm}
\usepackage{algorithmic}
\usepackage{amsmath}
\usepackage{graphicx}
\usepackage{multirow, booktabs, colortbl}
\usepackage{amsthm}
\usepackage{wrapfig}

\usepackage{listings}
\usepackage{xcolor}

\usepackage{titletoc}  
\usepackage{hyperref}       

\newcommand{\CUT}[1]{}
\newcommand{\NOTE}[1]{\textcolor{red}{[NOTE: #1]}}

\title{Pan-FM: A Pan-Organ Foundation Model with Saliency-Guided Masking for Missing Robustness}

\CUT{
\author{%
  David S.~Hippocampus\thanks{Use footnote for providing further information
    about author (webpage, alternative address)---\emph{not} for acknowledging
    funding agencies.} \\
  Department of Computer Science\\
  Cranberry-Lemon University\\
  Pittsburgh, PA 15213 \\
  \texttt{hippo@cs.cranberry-lemon.edu} \\
}
}

\author{%
  Qiangqiang Wu$^{1, 2}$ \quad
  Grace McIlvain$^{2}$ \quad
  Zhou Yu$^{2}$ \quad
  Junhao Wen$^{1, 2}$\thanks{Corresponding author} \\[0.5em]
  $^{1}$Laboratory of AI and Biomedical Science (LABS) \\
  $^{2}$Columbia University \\[0.3em]
  \texttt{\{qw2487, gm3128, zy2461\}@columbia.edu}, \texttt{jw4745@cumc.columbia.edu}
}

\begin{document}

\maketitle

\begin{abstract}

Foundation models (FMs) have shown great promise in medical imaging, but most FMs are trained on unimodal data within isolated domains, such as brain MRI alone. Human aging and disease arise through coordinated biological processes across organs, therefore motivating multimodal FMs that learn whole-body representations. A key challenge, however, is that real-world multimodal biomedical data are often missing not at random, which can reduce power, limit generalizability, and introduce bias. We propose Pan-FM, a pan-organ foundation model pre-trained on imaging from seven organs (\textit{Brain}, \textit{Heart}, \textit{Adipose}, \textit{Liver}, \textit{Kidney}, \textit{Spleen}, and \textit{Pancreas}) under realistic missing-organ scenarios. Pan-FM uses a unified backbone that handles organ missingness during both training and inference, and is pre-trained with masking-based self-distillation. We find that naive multimodal pre-training leads to dominant-organ shortcut learning bias, with the model over-relying on dominant organs such as adipose and heart. To address this, we introduce Saliency-Guided Masking (SGM), which uses the model attention distribution to adaptively mask dominant organs during pre-training, thus encouraging more balanced cross-organ, whole-body learning. 
Notably, SGM introduces negligible computational overhead and can be seamlessly integrated into existing self-supervised learning frameworks to improve multi-organ representation learning. 
On the UK Biobank, Pan-FM achieves stronger prediction across 13 disease categories and 14 single disease entities than single-organ and multi-organ baselines, with improved robustness under missing-organ settings. Pan-FM serves as a scalable solution to realistic modality-missingness in multimodal learning in system neuroscience and as a step toward more generalizable whole-body FMs. Code and model will be made publicly available upon acceptance.

\end{abstract}

\section{Introduction}
Foundation models (FMs) \cite{chexzero,mei2022radimagenet,wang2025triad,beeche2025pan,agrawal2025pillar} pre-trained on large-scale biomedical data via self-supervised learning (SSL) have emerged as a powerful paradigm in medical AI, enabling transferable representations across a wide range of downstream clinical tasks. For example, recent efforts have developed FMs for the brain~\cite{brainiac} and the heart~\cite{shad2026generalizable}, demonstrating strong performance on organ-specific disease prediction. However, most prior FMs have been trained in isolated domains using imaging data from a single organ system, thereby i) overlooking cross-organ interactions in aging and disease \cite{multi2026multi,multi2026mri,wen2025refining,wen2025metbag,wen2024genetic,wen2025towards,wen2023multiscale} and neglecting the recent notion that "No Organ System Is An Island" \cite{wen2024biological}, ii) reducing generalizability under missing-not-at-random settings (e.g., individuals undergoing brain MRI are typically healthier), and iii) limiting training efficiency and power by constraining to a single modality given the data-hungry nature of FMs. This motivates the development of multimodal FMs that integrate multiple organ modalities to more fully capture disease etiology and aging biology.

\begin{figure}
\begin{center}
   \includegraphics[width=1.0\linewidth]{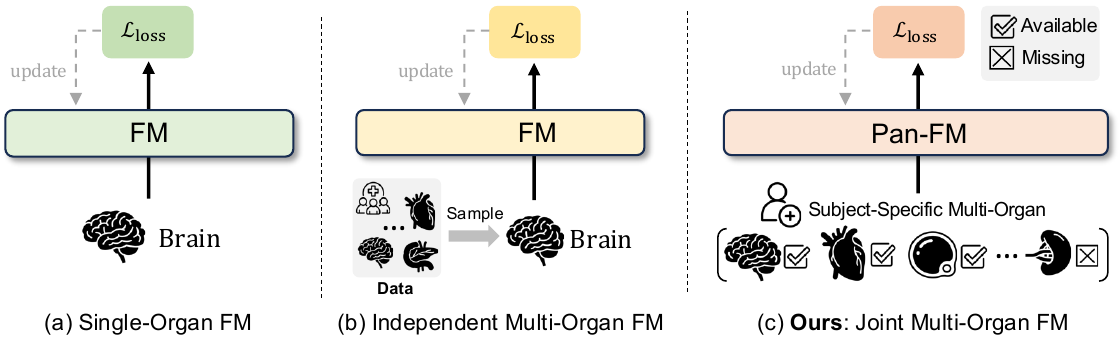} 
\end{center}
 \caption{(a) Organ-specific FMs \cite{brainiac,shad2026generalizable} trained on data from a single organ system (e.g., brain); (b) Independent multi-organ FMs \cite{yang2026decipher,wang2025triad} learned by sampling one organ sample per iteration from a multi-organ dataset, lacking subject-level multi-organ modeling; (c) Our proposed Pan-FM jointly learns cross-organ representations from subject-level multi-organ data, explicitly designed to handle missing-organ scenarios prevalent in biomedical research and clinical settings.}
\label{concep_compare}
\end{figure}

As shown in Fig.~\ref{concep_compare}, recent multi-organ learning methods \cite{yang2026decipher,mei2022radimagenet,wang2025triad} process each organ independently, learning multi-organ representations via single-organ training iterations rather than jointly modeling structured inter-organ dependencies. Moreover, they largely overlook missing-organ scenarios (missing-not-at-random), a clinically prevalent setting where imaging data for certain organs is unavailable due to incomplete protocols or acquisition constraints. Existing methods either assume full modality availability or address missingness post-hoc, yielding representations that are neither inherently robust nor biologically meaningful under organ absence. This motivates a unified pan-organ FM that jointly models cross-organ interactions and handles missing organ problems.


To achieve this, we propose Pan-FM, a multi-organ FM that jointly learns cross-organ representations from \textit{in vivo} MRI data across 7 organ systems. We design a unified backbone that supports cross-organ missingness during both training and inference, and pre-train it via masking-based self-distillation. To mitigate dominant-organ shortcut learning bias during the na\"{i}ve pre-training, where the model over-relies on a subset of organs, we introduce Saliency-Guided Masking (SGM), a simple and effective strategy that adaptively selects organs to mask based on the model’s attention distribution. Specifically, SGM identifies organs that the model currently over-relies on and preferentially masks them during pre-training. This self-regulating mechanism progressively balances the model's reliance across more organs, while introducing negligible computational overhead and no architectural changes.

We use UK Biobank \cite{ukbb} as our pre-training data source, which includes MRI-derived organ-specific features across seven organ systems in 41,969 participants, including \textit{Brain}, \textit{Heart}, \textit{Adipose}, \textit{Liver}, \textit{Kidney}, \textit{Spleen}, and \textit{Pancreas}. To demonstrate the pre-training effectiveness of the proposed Pan-FM, we perform downstream classification for organ-specific multi-disease categories (e.g., all diseases related to the central nervous system) and single disease prediction (e.g., Alzheimer's disease). Pan-FM consistently outperforms single-organ and multi-organ baselines, with particularly pronounced advantages on systemic diseases that closely interact with multiple organ systems. Under missing-organ evaluation settings, Pan-FM shows superior robustness compared to multi-organ baselines.

In summary, the main contributions of our work are:
\begin{itemize}
\item We present Pan-FM, to the best of our knowledge, the first multi-organ FM for jointly learning cross-organ representations from 7 organ systems in the presence of missing organs. 
\item We identify dominant-organ shortcut learning bias as a key limitation in multi-organ SSL, and propose Saliency-Guided Masking (SGM), a plug-and-play strategy that facilitates  more balanced cross-organ representation learning with negligible overhead and no architectural changes, and integrates seamlessly into existing SSL frameworks.
\item We demonstrate that Pan-FM achieves improved performance and missing-organ robustness on downstream disease prediction tasks, including prediction for single disease endpoints and organ-specific disease categories, compared to various baseline approaches.
\item We enable a flexible foundation framework that supports both hypothesis-driven and hypothesis-free research in system and clinical neuroscience. In the hypothesis-driven setting, prior knowledge can guide model construction; for example, if Alzheimer’s disease is thought to involve the brain-heart-eye axis, Pan-FM can be instantiated on these three organs to test related hypotheses. In the hypothesis-free setting, Pan-FM can be scaled to learn an organ-agnostic, whole-body foundation model with broad clinical generalizability, as demonstrated in the current work. 
\end{itemize}


\section{Related Work}

\textbf{Multi-Organ Aging and Disease Analysis.} Prior work characterizes multi-organ aging and disease primarily through organ-specific \emph{biological age gaps} (BAGs), defined as the difference between predicted biological age and chronological age derived from imaging or physiological features. Early studies established BAGs across brain and body systems and linked them to disease and mortality~\cite{tian2023heterogeneous, wen2024genetic}, with follow-up work extending the framework to MRI-based multi-organ age gaps~\cite{multi2026mri}, proteomics~\cite{wen2025refining, wen2025pnas}, metabolomics~\cite{wen2025metbag}, and the recent \emph{pan-disease} paradigm~\cite{multi2026multi}, alongside cross-organ axis studies~\cite{mccracken2022heart, wen2023multiscale}. Together these efforts outline a roadmap toward a multi-organ \emph{medical digital twin}~\cite{wen2025towards}. Yet they treat BAGs as intermediate variables combined via post-hoc late fusion, which cannot model inter-organ dependencies at the representation level, and assume complete multi-organ acquisitions, misaligned with clinical practice where organ missingness is common. This motivates Pan-FM, which performs multi-organ learning in a unified backbone and remains robust under missing-organ scenarios.

\textbf{Self-Supervised Learning.} Self-supervised learning (SSL) serves as the foundation for FM development. 
Contrastive learning methods use positive and negative pairs~\cite{simclr,moco,mocov3,clip,wu2021progressive}. Negative-free SSL methods rely on architectural asymmetry~\cite{byol,simsiam,assran2023self,assran2025v} or covariance regularization~\cite{vicreg, barlowtwins}. Recent methods use teacher--student momentum updates~\cite{dino, dinov2} with self-distillation, and masked image modeling reconstructs masked patches~\cite{mae, beit, simmim, ibot, wu2023dropmae}. While recent extensions explore adaptive image-patch masking strategies~\cite{kakogeorgiou2022hide, li2022semmae, wang2023hard}, Pan-FM extends them to the organ level via organ saliency-guided multinomial sampling, which mitigates dominant-organ shortcut learning under missing-organ scenarios. 
Building on these SSL methods, many biomedical FMs have emerged for brain MRI~\cite{brainiac,mazher2025towards,barbano2025anatomical,scholz2025mm}, heart MRI~\cite{zhang2024foundation}, chest imaging~\cite{chexzero, biomedclip}, retinal imaging~\cite{retfound, wei2024visionclip, shi2025multimodal}, pathology~\cite{uni, virchow, ctranspath, hipt}, ultrasound~\cite{usfm}, echocardiography~\cite{echoclip}, and dermatology~\cite{monet}, together with general-purpose medical foundation models~\cite{medsam, biomedgpt, medclip, radfm}. Despite these advances, most FMs remain organ-specific. While some works \cite{yang2026decipher,wang2025triad} extend to multi-organ MRI, they rely on independent organ training and fail to capture cross-organ dependencies. Moreover, current FMs assume complete and homogeneous inputs, which is unrealistic in clinical settings with organ missingness. To the best of our knowledge, Pan-FM is the first multi-organ FM that learns whole-body representations under missing-organ scenarios.

\section{Methodology}
This section presents Pan-FM, a multi-organ foundation model that jointly learns cross-organ representations across seven organ systems and is designed to handle biologically meaningful missing modalities (miss-not-at-random), a common challenge in system neuroscience and large-scale biobanks where heterogeneous organ data are only partially collected at the population level. 

\subsection{Architecture}
\begin{figure}
\begin{center}
   \includegraphics[width=1.0\linewidth]{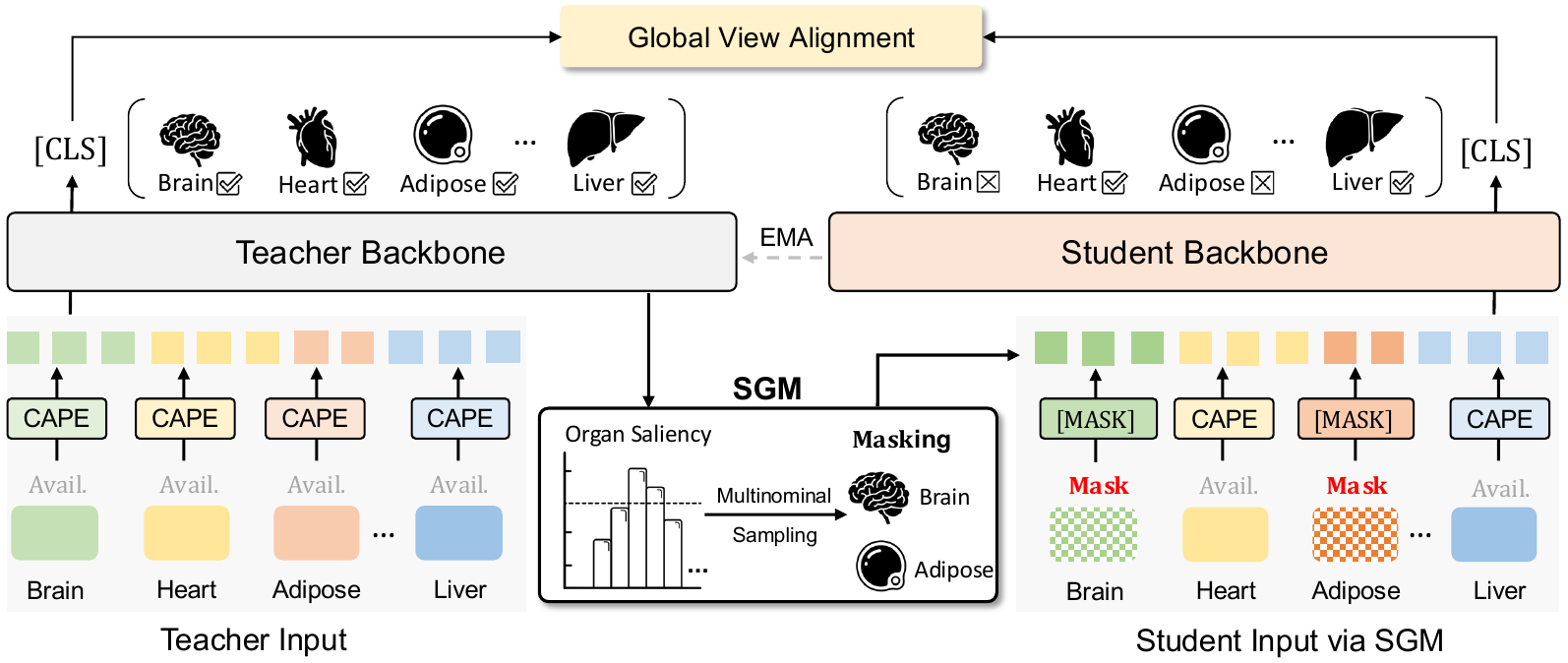} 
\end{center}
\vspace{-0.25cm}
 \caption{Overview of Pan-FM pre-training with Saliency-Guided Masking (SGM). The teacher backbone receives the full set of available organ tokens as input, while the student backbone receives a masked view constructed by SGM. Both backbones share the same architecture.}
\label{architecture}
\end{figure}

The overall architecture of Pan-FM is shown in Fig.~\ref{architecture}. Specifically, Pan-FM treats each organ's MRI features as semantic token patches, and integrates them via a Transformer backbone \cite{vit} to facilitate cross-organ representation learning. Modality availability is explicitly encoded and learnt, allowing organ missingness to serve as structured input with biological meaning rather than being treated as discarded missing-at-random noise.
 
 \noindent\textbf{Cross-Attention based Tabular Patch Encoding (CAPE).} 
 A key challenge in our multi-organ representation learning setting is the dimensional imbalance across organ systems. Hand-crafted low-dimensional MRI features vary substantially across organs: the \textit{Brain} consists of 122 volumetric regions of interest (ROIs) while the \textit{Pancreas} only has 3 features (i.e., \textit{Brain} > \textit{Heart} > \textit{Adipose} > \textit{Liver} = \textit{Kidney} = \textit{Spleen} = \textit{Pancreas}, see Appendix~\ref{appendix:feat_details}). 
To alleviate this, we employ a Cross-Attention Patch Encoder (CAPE) that converts each organ's features into a compact set of semantic tokens with a unified embedding dimension. 
Formally, for each organ $o$, organ-specific imaging features $\mathbf{x}_o \in \mathbb{R}^{D_o}$ are projected into feature embeddings $\mathbf{E}_o \in \mathbb{R}^{D_o \times d}$, and aggregated into $K_o$ compact tokens via cross-attention with learnable queries $\mathbf{Q}_o \in \mathbb{R}^{K_o \times d}$:
\begin{equation}
    \mathbf{z}_o = \text{softmax}\left(\frac{\mathbf{Q}_o \mathbf{W}_Q (\mathbf{E}_o \mathbf{W}_K)^T}{\sqrt{d_\text{head}}}\right) \mathbf{E}_o \mathbf{W}_V
\end{equation}
where $\mathbf{W}_Q, \mathbf{W}_K, \mathbf{W}_V$ are learnable projection matrices and $d_\text{head}$ is the attention head dimension. Each learnable query acts as a summary slot that attends to the most informative feature combinations within that organ. The resulting tokens $\mathbf{z}_o \in \mathbb{R}^{K_o \times d}$ share a unified dimension $d$ across all organs, enabling direct concatenation with a global \texttt{[CLS]} token for cross-organ Transformer interaction.

\textbf{Missing Organ Learner.} 
A defining characteristic of large-scale biobank data collections is systematic and biologically meaningful whole-body organ incompleteness. In practice, participants frequently have missing imaging data for one or more organ systems due to study design, contraindications, or selection bias, such that missingness itself carries biological meaning and can be explicitly modeled across organ systems. Pan-FM is designed to operate under arbitrary organ subsets during both training and inference. To achieve this, we define an organ availability vector $\mathbf{a} \in \{0,1\}$, where $a_o = 1$ indicates that organ $o$ is observed. For each organ $o$, CAPE produces a token set $\mathbf{z}_o \in \mathbb{R}^{K_o \times d}$ if observed, and substitutes a learnable mask token $[\texttt{MASK}]_o$ otherwise:

\begin{equation}
    \tilde{\mathbf{z}}_o = a_o \cdot \mathbf{z}_o + (1 - a_o) \cdot [\texttt{MASK}]_o + \mathbf{p}_o
\end{equation}
where $\mathbf{p}_o \in \mathbb{R}^{K_o \times d}$ is a learnable organ-specific embedding added to all organs regardless of availability, providing the Transformer backbone with explicit position and organ information.

\subsection{Na\"{i}ve Pre-training via Masking-based Self-Distillation}
\label{sec:naive_pretrain}
DINOv2~\cite{dinov2} has demonstrated remarkable success in learning transferable visual representations through self-distillation pre-training. Here, we adapted a DINOv2-style self-distillation for multi-organ representation learning to our multi-organ imaging features under real-world organ-missing scenarios. Specifically, we use a teacher model to receive the full set of available organs for each participant, while the student model receives an additional masked view produced by random organ masking, i.e., for each observed organ $o$, its full token set $\mathbf{z}_o$ is randomly replaced by $[\texttt{MASK}]_o$. Teacher parameters are updated via exponential moving average (EMA).


\noindent\textbf{Pre-training Objective.} The original DINOv2 pre-training objective includes three losses: 1) A global alignment loss $\mathcal{L}_\text{g}$ encourages the student's \texttt{[CLS]} token to match the teacher's full-view representation; 2) A masked token distillation loss $\mathcal{L}_\text{m}$ drives the student to reconstruct missing organ tokens from cross-organ context; 3) A KoLeo regularization term $\mathcal{L}_\text{k}$ encourages a uniform \texttt{[CLS]} embedding distribution. The overall loss is the combination of all the three losses.
\CUT{
\begin{equation}
    \mathcal{L} = \mathcal{L}_\text{global} + 
    \mathcal{L}_\text{mask} + 
    \lambda \mathcal{L}_\text{KoLeo}
\end{equation}

where $\lambda = 0.1$ by default.
}

\begin{wrapfigure}{r}{0.6\textwidth}
    \centering
    \vspace{-12pt}
    \includegraphics[width=0.6\textwidth]{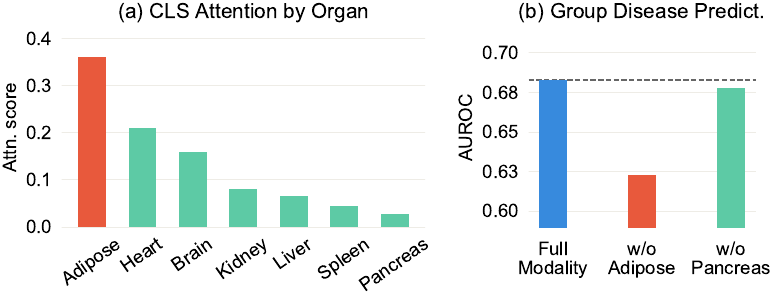}
        \vspace{-15pt}
    \caption{
    \textbf{Dominant-organ shortcut learning bias.} \textbf{(a)} Teacher CLS attention scores across organ systems. Adipose receives disproportionately high attention. \textbf{(b)} Mean group disease AUROC obtained by linear probing under organ removal. Removing adipose causes substantial degradation whereas removing pancreas has negligible effect.
    }
    \label{fig:dominant_organ}
    \vspace{-7pt}
\end{wrapfigure}

\noindent\textbf{Dominant-Organ Shortcut Learning Bias.}
We observe an intriguing phenomenon in the na\"{i}ve DINOv2-style \cite{dinov2} pre-training on our multi-organ data. That is, the model consistently concentrates its attention on a small subset of organs (see Fig. \ref{fig:dominant_organ}), which we term \textit{dominant-organ shortcut learning bias}. Notably, the most dominant organ in our setting is \textit{Adipose}, despite contributing only 16 tokens --- neither the largest token budget nor the highest feature dimensionality among the seven organ systems. This phenomenon causes the model to learn shortcut representations that over-rely on \textit{dominant organs}. As a result, as shown in Fig. \ref{fig:dominant_organ}, the model exhibits poor robustness when dominant organs are absent at inference time. 

We attribute this to two complementary factors: 1) From a biological perspective, \textit{Adipose}-derived features such as visceral and subcutaneous fat volumes are strong proxies for systemic metabolic state, exhibiting high cross-organ predictive capacity for cardiovascular, hepatic, and endocrine conditions; 2) \textit{Adipose} tokens provide sufficient cross-organ information to effectively construct a representative global \texttt{[CLS]} embedding, such that the model can minimize $\mathcal{L}_\text{global}$ by attending more to Adipose, bypassing the need to integrate the other organs.

\CUT{
\noindent\textbf{Is iBOT-Style Organ Reconstruction Necessary?}
We incorporate the iBOT loss $\mathcal{L}_\text{mask}$ in \cite{} for organ-level reconstruction. Unlike image patches, which carry rich local spatial context enabling reliable reconstruction from neighboring tokens, organ-level reconstruction lacks such structured local context. The remaining organs provide only coarse, loosely correlated signals for recovering a masked organ's representation, making cross-organ reconstruction fundamentally under-constrained.
As shown in Table.~\ref{}, $\mathcal{L}_\text{mask}$ consistently degrades downstream performance, suggesting that reconstructing organs under such uncertainty introduces noise rather than useful learning signal. We therefore remove $\mathcal{L}_\text{mask}$ from our framework, yielding a simpler and more effective pre-training objective: $\mathcal{L}_\text{All}$ = $\mathcal{L}_\text{global}$ + $\lambda \mathcal{L}_\text{KoLeo}$. \NOTE{Maybe put this in the supplementary.}
}

\subsection{Saliency-Guided Masking}
\label{sec:SGM}

\noindent\textbf{Motivation.} A primary goal of Pan-FM is to remain effective in real-world clinical settings, where patients may lack imaging data for one or more organ systems. However, as identified in Section~\ref{sec:naive_pretrain}, the na\"{i}ve pre-training leads to dominant-organ reliance, where the model disproportionately depends on a small subset of organs for feature extraction. This fundamentally limits clinical applicability, i.e., when several dominant organs are unavailable at inference time, model performance degrades significantly. In addition, this organ dominance can obscure biologically meaningful signals from less dominant organs (e.g., in the hypothesis-free setting for whole-body FMs), limiting the model’s utility not only for downstream prediction but also for biological and mechanistic discovery using computational approaches such as genome-wide associations (GWAS) and post-GWAS analyses \cite{multi2026mri}.

To mitigate dominant-organ shortcut learning and move toward a whole-body FM, we propose Saliency-Guided Masking (SGM), which dynamically identifies the most dominant organs and preferentially masks them during pre-training, forcing the student to extract discriminative features from less dominant organs and gradually yielding balanced cross-organ representations (see Appendix~Fig.~\ref{fig:saliency_compare}).

\noindent\textbf{Optimization Formulation.}
We formulate the organ masking strategy as a minimax optimization over the global self-distillation objective. Let $M \subseteq \mathcal{O}_\text{avail}$ denote a set of available organs to mask, and $\Delta(\mathcal{M})$ denote the probability simplex over the collection of all valid masking subsets $\mathcal{M} = \{M \subseteq \mathcal{O}_\text{avail} : 1 \leq |M| \leq n_\text{max}\}$, where $n_\text{max} = \min\!\left(\lfloor|\mathcal{O}_\text{avail}| \cdot r_\text{mask}\rfloor,\, |\mathcal{O}_\text{avail}| - 1\right)$ ensures at least one organ remains visible to the student, and $r_\text{mask}$ is the masking ratio. We define $\mathcal{O}_\text{miss} = \mathcal{O} \setminus \mathcal{O}_\text{avail}$ as the set of naturally 
missing organs. The student and teacher tokens for organ $o$ are constructed as:
\begin{equation}
    \label{eq:mask_generate}
    \tilde{\mathbf{z}}_o^s = 
    \begin{cases}
        [\texttt{MASK}]_o + \mathbf{p}_o, & o \in \mathcal{O}_\text{miss} \cup M, \\
        \mathbf{z}_o + \mathbf{p}_o, & \text{otherwise},
    \end{cases}
    \qquad
    \tilde{\mathbf{z}}_o^t = 
    \begin{cases}
        [\texttt{MASK}]_o + \mathbf{p}_o, & o \in \mathcal{O}_\text{miss}, \\
        \mathbf{z}_o + \mathbf{p}_o, & \text{otherwise},
    \end{cases}
\end{equation}
where $[\texttt{MASK}]_o$ is a learnable mask embedding and $\mathbf{p}_o$ is the organ-ID embedding defined in Section~\ref{sec:naive_pretrain}. We train a student backbone $f_S(\cdot;\theta)$ that minimizes the self-distillation loss under the adversarial masking distribution:
\begin{equation}
    \min_\theta \max_{P(M) \in \Delta(\mathcal{M})} \,
    \mathbb{E}_{M \sim P(M)}\!\left[
        \mathcal{L}_{g}\!\left(
            f_S\!\left(\{\tilde{\mathbf{z}}_o^s\}_{o=1}^{O};\,\theta\right),\,
            f_T\!\left(\{\tilde{\mathbf{z}}_o^t\}_{o=1}^{O}\right)
        \right)
    \right]
    \label{eq:minimax}
\end{equation}
where $\mathcal{L}_g$ is the global alignment loss defined in Section~\ref{sec:naive_pretrain}, $f_T(\cdot)$ is the fixed teacher encoder. The inner maximization seeks the adversarial masking distribution --- one that removes the organs most critical to the teacher's global representation, constructing the hard partial view for the student. The outer minimization then optimizes the student to align with the teacher's full-view representation under this adversarial mask, forcing it to leverage complementary information from other available organ systems rather than relying on some dominant organs.

\noindent\textbf{Computational Challenge.}
Directly solving the inner maximization in (\ref{eq:minimax}) requires evaluating the loss over all $|\mathcal{M}|$ masking subsets at every training step, which is computationally intractable. We therefore seek a tractable approximation to the adversarial masking distribution $P^*(M)$.

\begin{algorithm}[t]
\caption{Pan-FM PyTorch Pseudocode}
\label{alg:sgm}
\CUT{
\begin{lstlisting}[language=Python,
                   basicstyle=\small\ttfamily,
                   commentstyle=\color{sgmcolor},
                   keywordstyle=\color{black},
                   escapeinside={(*@}{@*)},
                   aboveskip=0pt,
                   belowskip=0pt,
                   lineskip=-1pt]
                   }
\definecolor{sgmcolor}{RGB}{74,124,89}   % forest green，与图片一致
\definecolor{sgmbg}{RGB}{255,255,255}    % 保持浅绿背景
\begin{lstlisting}[language=Python,
                   basicstyle=\footnotesize\ttfamily,  % small → footnotesize
                   commentstyle=\color{sgmcolor},
                   keywordstyle=\color{black},
                   escapeinside={(*@}{@*)},
                   aboveskip=0pt,
                   belowskip=0pt,
                   lineskip=-6pt,                      % 更紧凑
                   xleftmargin=0pt,                    % 去掉左边距
                   framesep=0pt]
# f_s, f_t: student/teacher backbones
# m: EMA momentum; n_mask: masking budget
# tau: softmax temperature; lambda: KoLeo loss weight 
for x, a in loader: # full-view input, availability
    with no_grad:
        z_t, attn = f_t.forward(x, a) # teacher feed-forward
        
    # SGM
    S = organ_saliency(attn, organ_spans, a) # Eq.(5)-(6)
    M = torch.Multinomial(softmax(S / tau), n_mask) # Eq.(7)
    x_s = mask_organs(x, M, a, mask_tokens) # student masking view

    z_s = f_s.forward(x_s, a) # student feed-forward w/ masked x_s
    loss = L_g(z_s, z_t) + lambda * L_k(z_s)  # overall loss
    # optimization step
    loss.backward(); optimizer.step()
    f_t.params = m * f_t.params + (1-m) * f_s.params
\end{lstlisting}
\end{algorithm}

\noindent\textbf{Attention-Based Approximation.}
We observe that the optimal inner solution $P^*(M)$ 
concentrates probability mass on masks that remove the most 
informative organs --- those that contribute most to the 
teacher's global \texttt{[CLS]} representation. We propose 
to approximate organ importance using the teacher's 
own attention maps, which are already computed in the 
 forward pass at no additional cost.

Formally, let $\mathbf{a}_l^h \in \mathbb{R}^N$ denote the 
\texttt{[CLS]}-to-patch attention weights from block $l$ and 
head $h$, where $N = \sum_{o=1}^{O} K_o$ is the total number 
of patch tokens. After the teacher's forward pass on the full 
view, we average across all $L$ blocks and $H$ heads to obtain 
a per-token attention map, and aggregate over each organ's 
token span to compute its saliency score:
\begin{equation}
\label{eq:average_atten_all_layer}
    \bar{\mathbf{a}} = \frac{1}{LH} \sum_{l=1}^{L} 
    \sum_{h=1}^{H} \mathbf{a}_l^h \in \mathbb{R}^N,
    \qquad
    S_o = \sum_{j \in \text{organ}(o)} [\bar{\mathbf{a}}]_j.
\end{equation}
The organ-specific saliency scores are then converted into a masking 
probability distribution via softmax with temperature 
$\tau$:
\begin{equation}
    P_o = \frac{\exp(S_o / \tau)}
    {\sum_{k \in \mathcal{O}_\text{avail}} \exp(S_k / \tau)}
    \label{eq:sgm_prob}
\end{equation}
where unavailable organs are excluded by setting $S_o = 
-\infty$ before the softmax. The temperature $\tau$ 
interpolates between two extremes: $\tau \rightarrow \infty$ 
recovers uniform random masking, while $\tau \rightarrow 0$ 
concentrates all probability mass on the single most 
dominant organ.

\noindent\textbf{Organ Masking.}
Given the masking probabilities in Eq.~\eqref{eq:sgm_prob}, 
the set of organs to mask is drawn without replacement via 
multinomial sampling:
\begin{equation}
    M \sim \text{Multinomial}\!\left(
        \{P_o\}_{o \in \mathcal{O}_\text{avail}},\,
        n_\text{mask},\,
        \text{replace=False}
    \right)
\end{equation}
where $n_\mathrm{mask} \sim \mathrm{Uniform}\{1,\ldots,n_\mathrm{max}\}$ follows the same sampling budget as the random masking baseline, ensuring that the only difference between SGM and the uniform random masking baseline lies in the organ masking selection distribution. 

The complete pre-training procedure is summarized in Algorithm~\ref{alg:sgm}. SGM introduces negligible computational overhead (see Appendix~\ref{sec:overhead_analysis}), as organ saliency is derived directly from the teacher's pre-computed attention maps. Moreover, we demonstrate that SGM is a plug-and-play masking strategy that can be easily incorporated into existing SSL frameworks~\cite{simclr,dinov2,vicreg,barlowtwins} for improving multi-organ learning, as validated in Table~\ref{tab:ssl_w_sgm}.

\vspace{-0.2cm}
\section{Experiments}
\label{main:experiments}
\vspace{-0.2cm}

\textbf{Dataset.} 
We conduct experiments on the UK Biobank~\cite{ukbb}, leveraging MRI-derived features from 41,969 participants across 7 organ systems (\textit{Brain}, \textit{Heart}, \textit{Adipose}, \textit{Liver}, \textit{Kidney}, \textit{Spleen}, \textit{Pancreas}). To the best of our knowledge, UK Biobank is the most comprehensive multi-organ imaging resource in the biomedical field. For conceptualizing Pan-FM, we use low-dimensional MRI features extracted by reproducible image preprocessing pipelines, such as brain volumetric ROIs and left ventricular wall thickness based on the American Heart Association (AHA) protocol \cite{american2002standardized}. Since UK Biobank is a general-population cohort with disease labels derived from clinical history, the goal of this work is not to establish state-of-the-art disease prediction, but to show the technical value of Pan-FM in handling realistic missing-organ scenarios in real-world biobank data and in improving performance over baseline approaches. More dataset details are shown in Appendix~\ref{appendix:data_details}.




\label{sec:gdp_appendix}
\label{sec:sdp_appendix}

\begin{table}[t]
\centering
\caption{\textbf{Group-disease evaluation under linear probing (LP) and full fine-tuning (FT).} 
Mean AUROC across 13 disease groups on the held-out test set. \textbf{Bold} and \underline{underlined} denote the best and second-best within each regime. \textbf{Full LP/FT BalAcc results are in Appendix Tables~\ref{tab:group_linear_prob_results} and~\ref{tab:group_ft_results}.}}
\label{tab:group_combined}
\resizebox{\textwidth}{!}{%
\begin{tabular}{l|cc|cc|cc|cc|cc|cc|cc}
\toprule
\multirow{2}{*}{\textbf{Disease Group}} 
& \multicolumn{2}{c|}{\textbf{SimCLR}~\cite{simclr}} 
& \multicolumn{2}{c|}{\textbf{BYOL}~\cite{byol}} 
& \multicolumn{2}{c|}{\textbf{MoCo-v3}~\cite{mocov3}}
& \multicolumn{2}{c|}{\textbf{VICReg}~\cite{vicreg}} 
& \multicolumn{2}{c|}{\textbf{Barlow T.}~\cite{barlowtwins}} 
& \multicolumn{2}{c|}{\textbf{DINOv2}~\cite{dinov2}}
& \multicolumn{2}{c}{\textbf{Pan-FM}} \\
& LP & FT & LP & FT & LP & FT & LP & FT & LP & FT & LP & FT & LP & FT \\
\midrule
Infectious \& Parasitic & 0.627 & 0.629 & 0.614 & 0.626 & 0.648 & 0.634 & 0.645 & 0.642 & 0.637 & 0.621 & \underline{0.657} & \underline{0.672} & \textbf{0.668} & \textbf{0.694} \\
Neoplasms               & 0.628 & 0.629 & 0.618 & 0.622 & 0.634 & 0.638 & 0.625 & 0.629 & 0.625 & 0.637 & 0.629 & \underline{0.646} & \textbf{0.671} & \textbf{0.686} \\
Blood \& Immune         & 0.659 & 0.660 & 0.635 & 0.649 & \textbf{0.672} & \textbf{0.674} & 0.635 & 0.636 & 0.646 & 0.650 & 0.657 & 0.656 & \underline{0.666} & \underline{0.669} \\
Endocrine \& Metabolic  & 0.692 & 0.703 & 0.682 & 0.689 & \underline{0.708} & \underline{0.712} & 0.680 & 0.694 & 0.692 & 0.701 & 0.700 & 0.710 & \textbf{0.713} & \textbf{0.718} \\
Mental \& Behavioural   & 0.673 & 0.657 & 0.649 & 0.645 & 0.665 & 0.657 & 0.649 & 0.644 & 0.665 & 0.656 & \underline{0.675} & \underline{0.680} & \textbf{0.690} & \textbf{0.683} \\
Nervous System          & 0.667 & 0.664 & 0.655 & 0.655 & \underline{0.668} & 0.665 & 0.645 & 0.654 & 0.647 & 0.653 & 0.655 & \underline{0.665} & \textbf{0.678} & \textbf{0.681} \\
Eye                     & \underline{0.661} & 0.653 & 0.611 & 0.613 & 0.632 & 0.631 & 0.605 & 0.622 & 0.603 & 0.622 & 0.644 & \underline{0.661} & \textbf{0.672} & \textbf{0.674} \\
Circulatory System      & 0.688 & 0.688 & 0.660 & 0.668 & \textbf{0.699} & \underline{0.702} & 0.680 & 0.691 & 0.686 & 0.692 & 0.681 & \underline{0.702} & \underline{0.693} & \textbf{0.704} \\
Respiratory System      & 0.646 & 0.649 & 0.644 & 0.651 & \underline{0.657} & 0.663 & 0.644 & 0.652 & 0.654 & 0.660 & 0.654 & \underline{0.673} & \textbf{0.667} & \textbf{0.678} \\
Digestive System        & \textbf{0.638} & 0.633 & 0.615 & 0.615 & 0.632 & 0.636 & 0.611 & 0.617 & 0.624 & 0.626 & 0.619 & \underline{0.642} & \underline{0.635} & \textbf{0.654} \\
Skin System             & 0.629 & 0.623 & 0.616 & 0.612 & \underline{0.630} & 0.623 & 0.622 & 0.621 & 0.622 & 0.620 & 0.621 & \underline{0.630} & \textbf{0.671} & \textbf{0.666} \\
Musculoskeletal         & 0.653 & 0.650 & 0.639 & 0.640 & \textbf{0.661} & 0.664 & 0.640 & 0.644 & 0.643 & 0.647 & 0.644 & \underline{0.667} & \underline{0.659} & \textbf{0.669} \\
Genitourinary           & 0.636 & 0.637 & 0.633 & 0.631 & 0.647 & 0.642 & 0.644 & 0.642 & 0.648 & 0.650 & 0.635 & \underline{0.653} & \textbf{0.659} & \textbf{0.673} \\
\midrule
\textbf{Overall Mean (AUROC)}   & 0.653 & 0.652 & 0.636 & 0.640 & \underline{0.658} & 0.657 & 0.640 & 0.645 & 0.646 & 0.649 & 0.652 & \underline{0.666} & \textbf{0.672} & \textbf{0.681} \\
\textbf{Overall Mean (BalAcc)}  & 0.622 & 0.619 & 0.606 & 0.607 & \underline{0.625} & 0.625 & 0.612 & 0.618 & 0.615 & 0.618 & 0.623 & \underline{0.632} & \textbf{0.638} & \textbf{0.641} \\
\bottomrule
\end{tabular}%
}
\end{table}

\begin{table}[t]
\centering
\caption{\textbf{Missing-organ robustness under linear probing (LP) and full fine-tuning (FT).} AUROC comparison of SSL methods across 13 disease groups on the held-out test set. \textit{Standard} uses the full test set with natural missingness; \textit{Full Organs (7)} and \textit{Drop Organ} settings are evaluated on subjects with complete 7-organ imaging data, with random organ dropout repeated over 50 (LP) / 25 (FT) runs to ensure stable and fair assessment. $\Delta$ denotes the relative improvement of Pan-FM over the DINOv2 baseline. \textbf{BalAcc results and per-regime full tables are in Appendix Tables~\ref{tab:group_linear_prob_results_miss} and~\ref{tab:group_ft_results_miss}.}}
\label{tab:group_robust_combined}
\vspace{-0.15cm}
\renewcommand{\arraystretch}{0.9}
\resizebox{\textwidth}{!}{%
\begin{tabular}{l cc cc cc cc cc}
\toprule
\multirow{2}{*}{Method} & \multicolumn{2}{c}{Standard} & \multicolumn{2}{c}{Full Organs (7)} & \multicolumn{2}{c}{Drop 1 Organ} & \multicolumn{2}{c}{Drop 2 Organs} & \multicolumn{2}{c}{Drop 3 Organs} \\
\cmidrule(lr){2-3} \cmidrule(lr){4-5} \cmidrule(lr){6-7} \cmidrule(lr){8-9} \cmidrule(lr){10-11}
 & LP & FT & LP & FT & LP & FT & LP & FT & LP & FT \\
\midrule
SimCLR~\cite{simclr}            & 0.653 & 0.652 & 0.659 & 0.664 & 0.645 & 0.652 & 0.638 & 0.645 & 0.625 & 0.636 \\
BYOL~\cite{byol}                & 0.636 & 0.640 & 0.647 & 0.653 & 0.642 & 0.645 & 0.645 & 0.648 & 0.628 & 0.634 \\
MoCo-v3~\cite{mocov3}           & 0.658 & 0.657 & 0.660 & 0.667 & 0.653 & 0.658 & 0.655 & 0.662 & 0.642 & 0.647 \\
VICReg~\cite{vicreg}            & 0.640 & 0.645 & 0.647 & 0.650 & 0.640 & 0.644 & 0.639 & 0.642 & 0.624 & 0.625 \\
Barlow Twins~\cite{barlowtwins} & 0.646 & 0.649 & 0.642 & 0.643 & 0.632 & 0.633 & 0.632 & 0.630 & 0.618 & 0.619 \\
DINOv2~\cite{dinov2}            & 0.652 & 0.666 & 0.683 & 0.691 & 0.654 & 0.681 & 0.647 & 0.668 & 0.617 & 0.645 \\
\textbf{Pan-FM (Ours)}          & \textbf{0.672} & \textbf{0.681} & \textbf{0.694} & \textbf{0.704} & \textbf{0.681} & \textbf{0.696} & \textbf{0.672} & \textbf{0.687} & \textbf{0.654} & \textbf{0.675} \\
\midrule
$\Delta$ (\%)                   & +3.1\% & +2.3\% & +1.6\% & +2.0\% & +4.1\% & +2.1\% & +3.9\% & +2.7\% & +6.0\% & +4.6\% \\
\bottomrule
\end{tabular}%
}
\end{table}

\textbf{Data split and evaluation protocol.} We hold out 500 cognitively normal (CN) participants and 780 disease-positive participants, identified using ICD codes and clinical history, for evaluation. We study two downstream protocols: {(i) Group-disease prediction}, covering 13 disease categories as binary classification against CN and evaluated with linear probing (LP) and full-backbone fine-tuning (FT); and {(ii) Single-disease prediction}, covering 14 specific diseases with smaller positive cohorts, where positives are identified from the full dataset and the same 500 held-out CN serve as negatives. We report 10-run averages for group-disease prediction, and 10-fold cross-validation results for single-disease prediction. Disease definitions, ICD mappings, and split statistics are provided in Appendices~\ref{sec:gdp_appendix} and~\ref{sec:sdp_appendix}. Performance is measured by AUROC and balanced accuracy (BalAcc).

\textbf{Implementation details.}
We pre-train Pan-FM for 200 epochs with a 12-layer ViT~\cite{vit} following DINOv2~\cite{dinov2} (batch size 64, AdamW, lr=$10^{-4}$, hidden dim 384, 8 heads). SGM introduces a single hyperparameter, i.e., the temperature $\tau=0.25$, with ablations shown in Fig.~\ref{fig:ablation}(b). For downstream evaluation, we attach a linear head to the \texttt{[CLS]} token for both linear probing (LP) and full fine-tuning (FT). Additional details are provided in Appendix~\ref{appendix:downstream_evaluation}. 

\textbf{SSL baselines.}
Since no existing SSL method is tailored to multi-organ representation learning, we compare against representative general-purpose SSL methods~\cite{simclr,dinov2,mocov3,vicreg,barlowtwins,byol}. For fairness, all baselines use the same backbone and main pre-training settings as Pan-FM, with method-specific hyperparameters tuned separately. All methods follow identical downstream protocols, with baseline details provided in Appendix~\ref{appendix:imple_details_ssl}.

\begin{figure}[t]
\begin{center}
   \includegraphics[width=1.0\linewidth]{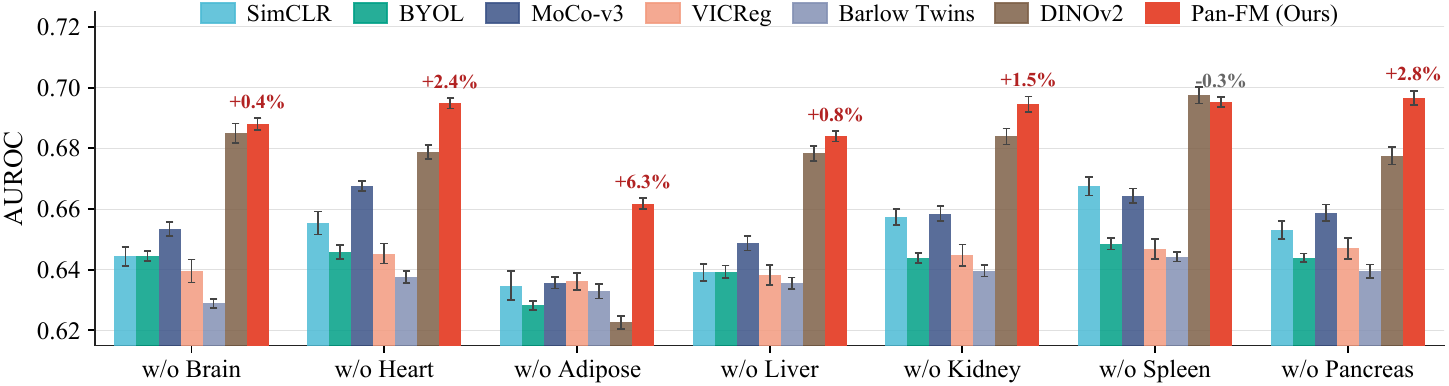}
\end{center}
\caption{\textbf{Single-organ dropout evaluation with linear probing.} Comparison of SSL methods for group-disease prediction across seven single-organ dropout (test-time organ removal) settings. Percentages denote the relative improvement of Pan-FM over the DINOv2 baseline.}
\label{fig:organ_removal}
\end{figure}

\begin{figure}[t]
\begin{center}
   \includegraphics[width=1.0\linewidth]{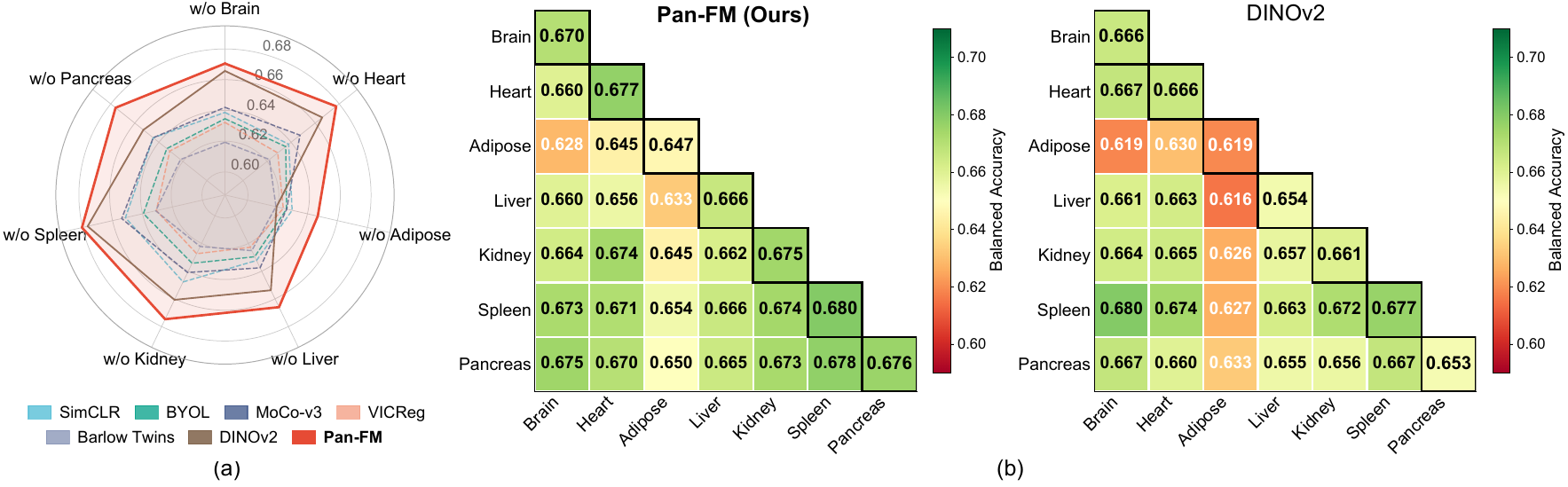}
\end{center}
\caption{\textbf{Robustness under organ dropout with full backbone fine-tuning.} (a) Radar chart of balanced accuracy across seven single-organ dropout settings. (b) Pairwise organ-dropout heatmaps for Pan-FM (left) and DINOv2 (right). Each cell shows the mean balanced accuracy across 13 disease groups when the row and column organs are simultaneously removed; diagonal entries correspond to single-organ removal. \textbf{Full radar charts and pairwise heatmaps under both LP and FT regimes (AUROC and BalAcc) are provided in Appendix Figures~\ref{fig:organ_removal_linear_balacc}--\ref{fig:organ_removal_linear_auroc_heat}.}}
\label{fig:organ_removal_ft}
\end{figure}

\subsection{Group-Disease Prediction}


\textbf{Overall performance under natural missingness.}
Table~\ref{tab:group_combined} reports group-disease prediction under natural missingness with linear probing (LP) and full-backbone fine-tuning (FT), with detailed per-disease results in Appendix Tables~\ref{tab:group_linear_prob_results} and~\ref{tab:group_ft_results}. Pan-FM achieves the best overall performance in both regimes (LP: 0.672/0.638; FT: 0.681/0.641), outperforming DINOv2 and the randomly initialized FT baseline (see Appendix~Table~\ref{tab:group_ft_results}). Contrastive SSL methods such as SimCLR and MoCo-v3 generally outperform non-contrastive alternatives, suggesting that explicit instance discrimination is beneficial across heterogeneous organ systems. Pan-FM's gains are especially pronounced for systemic conditions such as \textit{Neoplasms} (0.671 vs.\ 0.629 LP) and \textit{Skin System} (0.671 vs.\ 0.621 LP), highlighting our Pan-FM's stronger pan-organ representations.

\textbf{Missing-organ robustness.}
Table~\ref{tab:group_robust_combined} evaluates robustness under natural missingness (\emph{Standard}) and controlled organ dropout on complete-7-organ test subjects. As more organs are removed, all methods degrade, but Pan-FM remains consistently more robust: its AUROC margin over DINOv2 increases from +1.6\%/+2.0\% under \emph{Full Organs} to +6.0\%/+4.6\% under \emph{Drop-3} for LP/FT, respectively. Figs.~\ref{fig:organ_removal} and~\ref{fig:organ_removal_ft}(a) show no pronounced weak single-organ dropout direction for Pan-FM, while Fig.~\ref{fig:organ_removal_ft}(b) localizes DINOv2's failures mainly to \emph{Adipose} and \emph{Pancreas}. Although FT improves severe-missingness performance over LP, the Pan-FM--DINOv2 gap persists in both regimes, indicating that the robustness gain comes from SGM rather than the adaptation protocol.

\begin{table}[t]
\centering
\caption{\textbf{Single-disease evaluation with linear probing.} Comparison of SSL methods for single-disease prediction. Results are reported as AUROC / BalAcc.}
\label{tab:single_results}
\renewcommand{\arraystretch}{0.9}
\resizebox{\textwidth}{!}{%
\begin{tabular}{lccccccc}
\toprule
\textbf{Disease} & \textbf{SimCLR}~\cite{simclr} & \textbf{BYOL}~\cite{byol} & \textbf{MoCo-v3}~\cite{mocov3} & \textbf{VICReg}~\cite{vicreg} & \textbf{Barlow T.}~\cite{barlowtwins} & \textbf{DINOv2}~\cite{dinov2} & \textbf{Pan-FM} \\
\midrule
Alzheimer's Disease & 0.663 / 0.739 & 0.685 / 0.770 & 0.699 / 0.775 & 0.667 / 0.748 & 0.657 / 0.750 & \underline{0.719} / \textbf{0.782} & \textbf{0.726} / \underline{0.779} \\
Parkinson's Disease & 0.553 / 0.628 & 0.514 / 0.623 & 0.569 / 0.642 & 0.532 / 0.619 & 0.586 / 0.660 & \underline{0.598} / \underline{0.652} & \textbf{0.647} / \textbf{0.689} \\
Multiple Sclerosis & \textbf{0.609} / \textbf{0.644} & \underline{0.598} / \underline{0.638} & 0.591 / \underline{0.638} & 0.569 / 0.634 & 0.535 / 0.607 & 0.547 / 0.612 & 0.583 / 0.637 \\
Motor Neuron Disease & 0.547 / 0.709 & 0.576 / 0.741 & 0.535 / 0.704 & 0.487 / 0.682 & 0.500 / 0.702 & \underline{0.581} / \underline{0.731} & \textbf{0.594} / \textbf{0.742} \\
Systemic Lupus Eryth. & \textbf{0.681} / \underline{0.732} & 0.614 / 0.695 & 0.605 / 0.710 & 0.594 / 0.690 & 0.560 / 0.682 & 0.562 / 0.684 & \underline{0.675} / \textbf{0.756} \\
Sjogren Syndrome & 0.659 / 0.711 & 0.618 / 0.681 & 0.652 / 0.695 & 0.629 / 0.679 & 0.591 / 0.680 & \underline{0.665} / \underline{0.706} & \textbf{0.695} / \textbf{0.725} \\
Systemic Sclerosis & \underline{0.694} / \underline{0.794} & 0.620 / 0.735 & 0.596 / 0.719 & 0.683 / 0.777 & 0.671 / 0.758 & 0.664 / 0.762 & \textbf{0.719} / \textbf{0.804} \\
Sarcoidosis & 0.569 / 0.649 & 0.553 / 0.645 & \underline{0.589} / \underline{0.655} & 0.511 / 0.600 & 0.482 / 0.596 & 0.549 / 0.625 & \textbf{0.590} / \textbf{0.656} \\
Crohn's Disease & 0.526 / 0.595 & 0.520 / 0.601 & \textbf{0.579} / \textbf{0.631} & 0.520 / 0.586 & \underline{0.549} / 0.598 & 0.503 / 0.586 & 0.537 / \underline{0.606} \\
Ulcerative Colitis & 0.534 / 0.582 & 0.557 / 0.599 & \textbf{0.566} / \textbf{0.601} & 0.527 / 0.581 & 0.549 / \underline{0.600} & \underline{0.563} / 0.592 & 0.552 / 0.592 \\
Primary Biliary Chol. & 0.535 / 0.747 & 0.556 / 0.763 & 0.622 / 0.793 & 0.473 / 0.708 & 0.555 / 0.768 & \underline{0.636} / \underline{0.795} & \textbf{0.658} / \textbf{0.805} \\
Type 2 Diabetes & 0.777 / 0.733 & 0.762 / 0.723 & \textbf{0.808} / \textbf{0.754} & 0.771 / 0.736 & 0.758 / 0.719 & \underline{0.797} / \underline{0.752} & 0.793 / 0.750 \\
Hypertrophic Cardiom. & 0.715 / 0.763 & 0.729 / 0.793 & 0.828 / 0.855 & 0.757 / 0.806 & 0.726 / 0.776 & \underline{0.875} / \underline{0.880} & \textbf{0.893} / \textbf{0.900} \\
Pulmonary Hypertens. & \textbf{0.739} / \textbf{0.831} & 0.571 / 0.747 & 0.661 / 0.765 & 0.575 / 0.754 & 0.580 / 0.763 & 0.556 / 0.736 & \underline{0.695} / \underline{0.819} \\
\midrule
\textbf{Overall Mean} & 0.629 / 0.704 & 0.605 / 0.697 & \underline{0.636} / \underline{0.710} & 0.592 / 0.686 & 0.593 / 0.690 & 0.630 / 0.707 & \textbf{0.668} / \textbf{0.733} \\
\bottomrule
\end{tabular}%
}
\end{table}

\begin{table}[t]
\centering
\caption{Missing-organ evaluation with linear probing by applying SGM to various SSL methods.}
\label{tab:ssl_w_sgm}
\renewcommand{\arraystretch}{0.65}
\resizebox{\textwidth}{!}{%
\begin{tabular}{l cc cc cc cc cc}
\toprule
\multirow{2}{*}{Method} & \multicolumn{2}{c}{Standard} & \multicolumn{2}{c}{Full Organs (7)} & \multicolumn{2}{c}{Drop 1 Organ} & \multicolumn{2}{c}{Drop 2 Organs} & \multicolumn{2}{c}{Drop 3 Organs} \\
\cmidrule(lr){2-3} \cmidrule(lr){4-5} \cmidrule(lr){6-7} \cmidrule(lr){8-9} \cmidrule(lr){10-11}
 & AUROC & BalAcc & AUROC & BalAcc & AUROC & BalAcc & AUROC & BalAcc & AUROC & BalAcc \\
\midrule
SimCLR~\cite{simclr}              & 0.653 & 0.622 & 0.659 & 0.637 & 0.645 & 0.629 & 0.638 & 0.627 & 0.625 & 0.617 \\
SimCLR + \textbf{SGM}             & \textbf{0.671} & \textbf{0.635} & \textbf{0.682} & \textbf{0.656} & \textbf{0.663} & \textbf{0.639} & \textbf{0.656} & \textbf{0.639} & \textbf{0.648} & \textbf{0.635} \\
\midrule
VICReg~\cite{vicreg}              & 0.640 & 0.612 & 0.647 & 0.631 & 0.640 & 0.626 & 0.639 & 0.627 & 0.624 & 0.619 \\
VICReg + \textbf{SGM}             & \textbf{0.665} & \textbf{0.630} & \textbf{0.679} & \textbf{0.651} & \textbf{0.661} & \textbf{0.642} & \textbf{0.650} & \textbf{0.637} & \textbf{0.648} & \textbf{0.638} \\
\midrule
Barlow Twins~\cite{barlowtwins}   & 0.646 & 0.615 & 0.642 & 0.624 & 0.632 & 0.620 & 0.632 & 0.622 & 0.618 & 0.613 \\
Barlow Twins + \textbf{SGM}       & \textbf{0.666} & \textbf{0.633} & \textbf{0.674} & \textbf{0.653} & \textbf{0.659} & \textbf{0.640} & \textbf{0.646} & \textbf{0.632} & \textbf{0.635} & \textbf{0.626} \\
\midrule
DINOv2~\cite{dinov2}              & 0.652 & 0.623 & 0.683 & 0.654 & 0.654 & 0.642 & 0.647 & 0.639 & 0.617 & 0.616 \\
{DINOv2 + \textbf{SGM (Pan-FM)}}  & \textbf{0.672} & \textbf{0.638} & \textbf{0.694} & \textbf{0.670} & \textbf{0.681} & \textbf{0.662} & \textbf{0.672} & \textbf{0.655} & \textbf{0.654} & \textbf{0.638} \\
\bottomrule
\end{tabular}%
}
\end{table}

\textbf{Single-disease prediction.}
Table~\ref{tab:single_results} reports linear-probing results on 14 specific diseases. Pan-FM achieves the best overall performance (0.668/0.733), outperforming MoCo-v3 (0.636/0.710) and DINOv2 (0.630/0.707) across most diseases. The gains are most pronounced for multi-organ conditions, including \textit{Parkinson's Disease} (0.647 vs.\ 0.598) and \textit{Sjogren Syndrome} (0.695 vs.\ 0.665), indicating that Pan-FM learns system-level representations that improve disease-specific prediction.

\subsection{Ablation Studies}

\begin{figure}[h]
    \centering
    \includegraphics[width=1.0\linewidth]{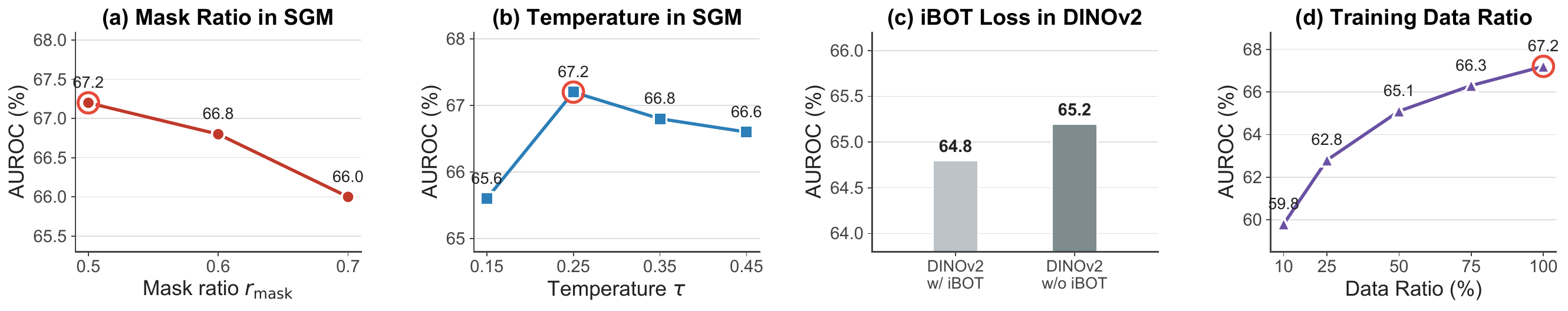}
    \caption{\textbf{Ablation studies.} 
    (a) Mask ratio $r_{\mathrm{mask}}$ in SGM. 
    (b) Temperature $\tau$ in SGM. 
    (c) Effect of the iBOT loss in DINOv2 for multi-organ representation learning. 
    (d) Effect of downstream training data ratio. 
    All results are mean AUROC (\%) across 13 disease groups under linear probing.}
    \label{fig:ablation}
\end{figure}

\textbf{Mask ratio $r_{\mathrm{mask}}$.} Since masking is applied only to participants with at least two available organs (Sec.~\ref{sec:SGM}), the budget starts from $r_{\mathrm{mask}}=0.5$. Fig.~\ref{fig:ablation}(a) shows performance degrades from 67.2 at 0.5 to 66.0 at 0.7. Unlike patch-level masking on natural images where 0.75 is tractable due to rich local context, full-organ masking removes entire anatomical units and requires a smaller ratio.

\textbf{Temperature $\tau$.} The temperature in Eq.~\ref{eq:sgm_prob} controls the sharpness of the organ-sampling distribution: small $\tau$ concentrates probability on a few dominant organs, while large $\tau$ collapses to uniform random masking. Fig.~\ref{fig:ablation}(b) shows $\tau=0.25$ achieves the best trade-off.

\textbf{iBOT loss in DINOv2.} The iBOT loss, designed for masked patch reconstruction on natural images, is ill-suited here: full-organ masking leaves insufficient context to reconstruct fine-grained organ-specific features. Fig.~\ref{fig:ablation}(c) shows iBOT slightly degrades performance (64.8 vs.\ 65.2). We therefore remove it from both Pan-FM and DINOv2, reporting the stronger iBOT-free variant throughout.

\textbf{Downstream training data ratio.} Fig.~\ref{fig:ablation}(d) shows performance scales with the amount of downstream labeled training data, improving consistently from 59.8 at 10\% to 67.2 at full scale with no saturation, indicating that Pan-FM representations continue to benefit from additional supervision.

\textbf{SGM as a plug-and-play strategy.} SGM can be integrated into other multi-view SSL frameworks. As shown in Table~\ref{tab:ssl_w_sgm}, applying SGM to SimCLR, VICReg, and Barlow Twins consistently improves both standard and missing-organ performance, demonstrating that SGM generalizes beyond the DINOv2 framework. Detailed per-disease improvements are shown in Appendix~Table~\ref{tab:group_linear_prob_results_w_sgm}.

Additional ablation studies are provided in Appendix Sections~\ref{sec:overhead_analysis}--\ref{sec:training_data_ratio}.

\section{Conclusion}
We introduced {Pan-FM}, \emph{the first} pan-organ foundation model pre-trained on 7 organs' MRI features under realistic missing-organ scenarios. Critically, we proposed {Saliency-Guided Masking (SGM)} in {Pan-FM}, which addresses the shortcut learning bias observed in naive multimodal pre-training, and can be plugged into prior existing multi-view models for superior performance. On the UK Biobank, Pan-FM consistently outperforms single-organ and missing-modality baselines across 13 disease groups and 14 individual diseases, with especially large gains under severe organ missingness. These results establish SGM as an effective remedy to modality shortcut bias and position Pan-FM as a step toward scalable, whole-body FM for systems neuroscience and clinical applications.

\newpage
\bibliographystyle{plainnat}
\bibliography{references}

\newpage
\appendix

\startcontents[appendices]
\printcontents[appendices]{}{1}{\section*{Appendix Contents}\setcounter{tocdepth}{2}}

\section{Data Details}
\label{appendix:data_split}
In this section, we detail the data splits for both pre-training and downstream tasks, including group-disease and single-disease prediction.

\subsection{Dataset}
\label{appendix:data_details}


We conduct experiments on the UK Biobank~\cite{ukbb}, a large-scale population cohort comprising approximately 500{,}000 participants recruited between 2006 and 2010. UK Biobank provides the most comprehensive population-scale resources for studying multi-organ biomedical phenotypes. In this work, we use MRI-derived, organ-specific imaging-derived phenotypes (IDPs) from 41{,}969 participants, covering seven organ systems: \textit{Brain}, \textit{Heart}, \textit{Adipose}, \textit{Liver}, \textit{Kidney}, \textit{Spleen}, and \textit{Pancreas}.

For conceptualizing our Pan-FM in this work, we use low-dimensional MRI-derived features extracted from standardized image-processing pipelines, including brain volumetric regions of interest and cardiac measurements such as left-ventricular wall thickness following the American Heart Association (AHA) segmentation protocol. This design enables us to systematically study multi-organ representation learning under heterogeneous organ availability, while leaving direct pre-training on raw multi-organ imaging data as an important direction for future work.

Of note, the UK Biobank is a general-population cohort, and disease labels are derived from inpatient records and clinical history rather than disease-specific phenotyping. As a result, absolute predictive performance is expected to be modest. Accordingly, the goal of this section is not to establish state-of-the-art disease prediction, but to demonstrate the technical value of Pan-FM in handling realistic missing-organ scenarios in real-world biobank data and in improving performance over baseline approaches. 

\subsection{Group-Disease Prediction}
\label{sec:gdp_appendix}

\begin{table}[h]
\centering
\caption{\textbf{Definition of the 13 systemic disease categories via ICD-10.} A participant is assigned a positive label for a category if any of their recorded ICD-10 diagnoses fall within the specified range.}
\label{tab:icd10_categories}
\begin{tabular}{lc}
\toprule
\textbf{Disease Category} & \textbf{ICD-10 Range} \\
\midrule
Infectious \& parasitic diseases & B \\
Neoplasms & C--D4 \\
Blood \& immune system diseases & D50--D89 \\
Endocrine, nutritional \& metabolic diseases & E \\
Mental \& behavioural disorders & F \\
Nervous system diseases & G \\
Eye diseases & H0--H5 \\
Circulatory system diseases & I \\
Respiratory system diseases & J \\
Digestive system diseases & K \\
Skin system diseases & L \\
Musculoskeletal system diseases & M \\
Genitourinary system diseases & N \\
\bottomrule
\end{tabular}
\end{table}
\textbf{Disease Category Split.} Following \cite{wen2023multiscale}, disease categories are defined following the International Classification of Diseases, 10th Revision (ICD-10). We exclude Ear diseases (H6--H9) due to its insufficient positive cases, retaining 13 categories for downstream training and evaluation. Specifically, for each UK Biobank participant, we extract hospital inpatient records (field 41270) and self-reported diagnoses and map each ICD-10 code to its corresponding chapter. A participant is assigned a positive label for a given disease category if they have \emph{any} recorded diagnosis whose ICD-10 code falls within the ranges defined in Table~\ref{tab:icd10_categories}. Table~\ref{tab:icd10_categories} lists the 13 systemic disease categories used in our group-disease prediction task. These span a broad range of organ systems and pathological mechanisms.

\begin{table}[h]
\centering
\caption{\textbf{Disease category distribution across data splits.} We report the number of positive cases per ICD-10-defined disease category in the downstream training, validation, and held-out test sets.}
\label{tab:disease_distribution}
\begin{tabular}{lccc}
\toprule
\textbf{Disease Category} & \textbf{Train} & \textbf{Val} & \textbf{Test} \\
\midrule
Infectious \& parasitic diseases     & 1{,}897  & 89  & 137 \\
Neoplasms                            & 9{,}953  & 522 & 340 \\
Blood \& immune system diseases      & 2{,}532  & 121 & 148 \\
Endocrine \& metabolic diseases      & 8{,}083  & 410 & 332 \\
Mental \& behavioural disorders      & 3{,}243  & 165 & 154 \\
Nervous system diseases              & 4{,}014  & 206 & 184 \\
Eye diseases                         & 1{,}162  & 53  & 90  \\
Circulatory system diseases          & 13{,}010 & 692 & 455 \\
Respiratory system diseases          & 6{,}295  & 336 & 281 \\
Digestive system diseases            & 17{,}393 & 910 & 509 \\
Skin system diseases                 & 4{,}896  & 249 & 213 \\
Musculoskeletal system diseases      & 11{,}316 & 607 & 383 \\
Genitourinary system diseases        & 10{,}694 & 575 & 357 \\
\midrule
\textbf{Cognitively normal (CN)}     & 5{,}927  & 312     & 500   \\
\bottomrule
\end{tabular}
\end{table}

\textbf{Disease Category Distribution.} We partition the UK Biobank dataset into four splits to support both self-supervised pre-training and downstream disease prediction. As shown in Table~\ref{tab:disease_distribution}, the \textbf{held-out test set} ($N{=}1{,}280$) is constructed first by randomly sampling 500 cognitively normal (CN) participants together with 780 disease-positive participants, stratified across the 13 ICD-10 categories (60 cases per category). Because a single participant may carry diagnoses spanning multiple categories (i.e., comorbidities), the total positive count aggregated across categories exceeds 780 in the test set, as reflected in Table~\ref{tab:disease_distribution}. This test set is strictly excluded from both pre-training and training. The \textbf{pre-training set} ($N{=}40{,}689$) comprises all remaining participants with at least one organ imaged, including CN, disease-positive, and unlabeled individuals, thereby maximizing data utilization for self-supervised learning. For downstream evaluation, we further filter the pre-training pool to retain only participants with confirmed CN or disease-positive status, yielding the \textbf{training set} ($N{=}36{,}220$), from which 5\% is held out as the \textbf{validation set} ($N{=}1{,}907$) via stratified sampling on CN status. This design ensures (i) the held-out test set provides an unbiased evaluation benchmark disjoint from all training data, (ii) pre-training leverages the largest possible pool of multi-organ imaging without requiring clean labels, and (iii) downstream supervised evaluation uses only participants with reliable diagnostic annotations.

\begin{table}[h]
\centering
\caption{\textbf{Organ imaging availability across data splits.} Number of participants with each organ imaged, along with the coverage rate relative to the split total. The majority of participants lack complete 7-organ imaging.} 
\label{tab:organ_availability}
\setlength{\tabcolsep}{5pt}
\begin{tabular}{lcccc}
\toprule
\textbf{Modality / Metric} & \textbf{Pre-train} & \textbf{Train} & \textbf{Val} & \textbf{Test} \\
\midrule
\multicolumn{5}{l}{\emph{Per-organ availability (\# participants, \% coverage)}} \\
Brain     & 31{,}181 (77\%) & 27{,}731 (77\%) & 1{,}495 (78\%) & 1{,}000 (78\%) \\
Heart     & 33{,}629 (83\%) & 29{,}924 (83\%) & 1{,}554 (81\%) & 1{,}050 (82\%) \\
Adipose   & 25{,}964 (64\%) & 23{,}062 (64\%) & 1{,}186 (62\%) & 812 (63\%)     \\
Liver     & 26{,}166 (64\%) & 23{,}206 (64\%) & 1{,}210 (63\%) & 829 (65\%)     \\
Kidney    & 34{,}119 (84\%) & 30{,}346 (84\%) & 1{,}590 (83\%) & 1{,}087 (85\%) \\
Spleen    & 26{,}857 (66\%) & 23{,}834 (66\%) & 1{,}238 (65\%) & 835 (65\%)     \\
Pancreas  & 25{,}964 (64\%) & 23{,}062 (64\%) & 1{,}186 (62\%) & 812 (63\%)     \\
\CUT{
\multicolumn{5}{l}{\emph{Organs per participant}} \\
Mean (median)         & 5.01 (6)  & 5.00 (6)  & 4.96 (6)  & 5.02 (6)   \\
1 organ               & 2{,}951   & 2{,}616   & 156       & 99         \\
2--3 organs           & 9{,}103   & 8{,}195   & 417       & 286        \\
4--6 organs           & 15{,}250  & 13{,}581  & 700       & 451        \\
7 organs (complete)   & 13{,}385  & 11{,}828  & 634       & 444        \\
}
\bottomrule
\end{tabular}
\end{table}

\textbf{Organ Availability.} A key characteristic of the UK Biobank dataset \cite{ukbb} is that \emph{organ missingness is the norm rather than the exception}: participants are scanned under different acquisition protocols and scheduling constraints, resulting in varying subsets of organs being imaged per individual. Table~\ref{tab:organ_availability} quantifies this natural missingness across our data splits. On average, each participant has imaging from only 5.0 out of the 7 target organs, and only 33\% of participants have complete 7-organ imaging (e.g., $11{,}828$ out of $36{,}220$ in the training set). The remaining 67\% exhibit varying degrees of missingness: 37\% have 4--6 organs, 23\% have 2--3 organs, and 7\% have only a single organ imaged. Organ-level coverage further varies substantially across modalities, from 84\% (kidney) down to 64\% (adipose and pancreas), reflecting protocol-specific acquisition constraints in the UK Biobank imaging study. This consistent pattern across pre-training, training, validation, and test splits establishes missing-organ scenarios as the \emph{default} clinical reality rather than a corner case—motivating foundation models that are trained and evaluated natively under organ-missingness conditions, rather than assuming complete multi-organ acquisitions.

\CUT{
\begin{table}[h]
\centering
\caption{ICD-10 code mapping for the 14 specific diseases.}
\label{tab:single_disease_icd}
\begin{tabular}{llc}
\toprule
\textbf{Category} & \textbf{Disease} & \textbf{ICD-10} \\
\midrule
\multirow{4}{*}{Neurological}
    & Alzheimer's disease               & \texttt{G30, F00} \\
    & Parkinson's disease               & \texttt{G20}      \\
    & Multiple sclerosis                & \texttt{G35}      \\
    & Motor neuron disease              & \texttt{G122}     \\
\midrule
\multirow{4}{*}{Systemic autoimmune}
    & Systemic lupus erythematosus      & \texttt{M321, M328, M329} \\
    & Sj\"ogren syndrome                & \texttt{M350}     \\
    & Systemic sclerosis                & \texttt{M34}      \\
    & Sarcoidosis                       & \texttt{D86}      \\
\midrule
\multirow{3}{*}{Digestive}
    & Crohn's disease                   & \texttt{K50}      \\
    & Ulcerative colitis                & \texttt{K51}      \\
    & Primary biliary cholangitis       & \texttt{K743}     \\
\midrule
\multirow{3}{*}{Cardiometabolic}
    & Type 2 diabetes                   & \texttt{E11}      \\
    & Hypertrophic cardiomyopathy       & \texttt{I421, I422} \\
    & Pulmonary hypertension            & \texttt{I270}     \\
\bottomrule
\end{tabular}
\end{table}
}

\subsection{Single-Disease Prediction}
\label{sec:sdp_appendix}

\textbf{Single-Disease Definitions.} In addition to the 13 ICD-10 chapter-level disease groups used for group-disease prediction, we define a complementary set of 14 specific disease entities for fine-grained single-disease prediction. These diseases are selected to span the four major organ systems including \emph{neurological}, \emph{autoimmune}, \emph{digestive}, and \emph{cardiometabolic}. 
Each disease is identified through ICD-10 codes from UK Biobank:
\begin{itemize}
    \setlength{\itemsep}{2pt}
    \setlength{\leftmargin}{1.2em}
    \item \textbf{Neurological diseases} (\textit{Alzheimer's, Parkinson's, multiple sclerosis, motor neuron disease}): central to aging research and well-established links to multi-organ manifestations (e.g., cardiovascular comorbidity in Alzheimer's, metabolic alterations in Parkinson's).
    \item \textbf{Systemic autoimmune diseases} (\textit{systemic lupus erythematosus, Sj\"ogren syndrome, systemic sclerosis, sarcoidosis}): canonical examples of pathologies intrinsically involving multiple organs (skin, kidney, lung, heart), providing a rigorous test for pan-organ representations.
    \item \textbf{Digestive diseases} (\textit{Crohn's disease, ulcerative colitis, primary biliary cholangitis}): chosen as primarily localized disorders to probe whether Pan-FM remains competitive when the diagnostic signal is concentrated in a narrow anatomical region.
    \item \textbf{Cardiometabolic diseases} (\textit{type 2 diabetes, hypertrophic cardiomyopathy, pulmonary hypertension}): high-prevalence conditions with both organ-specific anchors (heart, pancreas) and broader systemic involvement, complementing the group-level endocrine and circulatory categories.
\end{itemize}

\textbf{Evaluation protocol.} Because each specific disease has substantially fewer positive casesr, the held-out test set used for group-disease evaluation is too small to provide reliable per-disease estimates. We therefore adopt a different protocol for single-disease prediction. Positive cases for each disease are extracted from the entire dataset based on the ICD-10 codes, while the 500 cognitively normal (CN) participants from the held-out test set serve as shared negatives across all diseases. For each disease, we perform 10-fold cross-validation with linear probing: the frozen Pan-FM (or baseline) encoder produces per-participant representations, and a linear classifier is trained and evaluated fold-by-fold; we report the mean AUROC and balanced accuracy across folds. The encoder itself is never updated during single-disease evaluation, ensuring that the reported performance reflects the transferability of the pre-trained representations rather than task-specific fine-tuning. Because linear probing keeps the encoder frozen, positive cases drawn from the dataset do not constitute label leakage with respect to self-supervised pre-training, as no disease labels are seen during the pre-training stage.

\section{Multi-Organ Input Features}
\label{appendix:feat_details}

\textbf{Multi-organ Input Features.} Pan-FM operates on MRI-derived imaging phenotypes (IDPs) rather than raw voxels, ensuring that each organ is represented by a fixed-length vector of clinically validated features. Table~\ref{tab:input_features} summarizes the per-organ feature composition, totaling \textbf{228 features} across the 7 organs. Brain features comprise 119 gray matter regional volumes extracted via MUSE~\cite{doshi2016muse}, covering subcortical structures and cortical regions across both hemispheres. Heart features comprise 80 cardiac IDPs from the UK Biobank cardiac MRI pipeline (field IDs within \texttt{24100--24181}), including chamber volumes, ejection fractions, wall thickness, and myocardial strain. Abdominal features (adipose, liver, kidney, spleen, pancreas) are derived from multi-contrast abdominal MRI within UK Biobank Category 149, capturing organ volumes, fat content (PDFF), and iron concentration as appropriate.

We enforce strict \emph{intra-organ completeness}: if any feature within an organ is missing, the entire organ is marked unavailable, i.e., reflecting the clinical reality that organ-level IDPs are either fully acquired or entirely absent. All features are z-score normalized per organ modality using statistics computed on the pre-training set.

\begin{table}[h]
\centering
\caption{Per-organ input features used by Pan-FM and all SSL baselines.}
\label{tab:input_features}
\small
\setlength{\tabcolsep}{4pt}
\renewcommand{\arraystretch}{1.15}
\begin{tabular}{p{1.5cm} c p{10.5cm}}
\toprule
\textbf{Organ} & \textbf{Dim.} & \textbf{Feature Description (UKBB Field IDs)} \\
\midrule
Brain    & 119 & Gray matter regional volumes from MUSE: 21 subcortical structures (e.g., hippocampus, amygdala, thalamus, caudate, putamen, cerebellum) and 98 cortical regions across both hemispheres. \\
\midrule
Heart    & 80  & UKBB cardiac MRI IDPs: \texttt{24100--24181} (excluding \texttt{24120, 24123}). Includes LV/RV end-diastolic / end-systolic / stroke volumes, ejection fractions, myocardial mass, AHA-segmented wall thickness (16 segments), and global longitudinal/circumferential strain. \\
\midrule
Adipose  & 16  & Body composition / ectopic-fat features: \texttt{21085} (VAT), \texttt{21090} (pancreas PDFF, used as ectopic-fat proxy), \texttt{22403--22406} (anterior/posterior thigh fat-free muscle volumes, L/R), \texttt{22408} (ASAT), \texttt{22410} (total trunk fat), \texttt{22432} (total abdominal adipose tissue index), \texttt{22434} (abdominal fat ratio), \texttt{22435} (muscle fat infiltration), \texttt{23355--23356} (posterior thigh MFI L/R), \texttt{24353--24354} (anterior thigh MFI L/R), \texttt{40061} (liver fat \%, used as ectopic-fat proxy). \\
\midrule
Liver    & 4   & \texttt{21080} (volume), \texttt{21088} (PDFF), \texttt{21089} (iron), \texttt{40062} (iron-corrected T1, cT1). \\
\midrule
Kidney   & 3   & \texttt{21081} (left kidney volume), \texttt{21162} (right kidney parenchymal volume), \texttt{21163} (inter-kidney centroid distance). \\
\midrule
Spleen   & 3   & \texttt{21083} (volume), \texttt{21170} (iron, IDEAL), \texttt{21173} (iron, protocol normalised). \\
\midrule
Pancreas & 3   & \texttt{21087} (volume), \texttt{21090} (PDFF), \texttt{21091} (iron). \\
\midrule
\textbf{Total} & \textbf{228} & \\
\bottomrule
\end{tabular}
\end{table}








\section{More Implementation Details}
\label{appendix:imple_details}

\subsection{Self-Supervised Learning (SSL) Baselines}
\label{appendix:imple_details_ssl}

We benchmark five representative SSL methods under our multi-organ pre-training protocol, including contrastive-based approaches ({SimCLR}~\cite{simclr}, {BYOL}~\cite{byol}, {MoCo-v3}~\cite{mocov3}, {Barlow Twins}~\cite{barlowtwins}, and {VICReg}~\cite{vicreg}. All baselines share the same ViT backbone, CAPE tokenizer, and training schedule as the DINOv2 \cite{dinov2} baseline and our Pan-FM. The only differences are the SSL objective and the auxiliary heads required by each method. Below we describe how each method is adapted to the multi-organ setting.

\paragraph{Shared settings across SSL baselines.} To ensure fair comparison, all SSL baselines use: (i) the same 12-layer ViT backbone with hidden dimension 384 and 8 attention heads; (ii) the same CAPE tokenizer producing 128 organ-aware tokens per participant (64 brain, 32 heart, 16 adipose, 4 each for liver/kidney/spleen/pancreas); (iii) the same availability-aware input handling, where missing organs are replaced with learnable organ-aware mask tokens; (iv) the AdamW optimizer with a learning rate of $1\mathrm{e}{-4}$, cosine decay, 10 warmup epochs, batch size 64, and 200 pre-training epochs; (v) a two-view training strategy---\emph{the first view} is the full view with only inherent missingness, and \emph{the second view} is a masked view where a subset of available organs is further masked with a mask ratio of 0.5, following Eq.~(\ref{eq:mask_generate}).

\paragraph{SimCLR.} SimCLR~\cite{simclr} treats the full view and masked view as two augmentations of the same participant, pulls their projected embeddings together, and contrasts them against all other participants in the batch feature space via the NT-Xent loss~\cite{simclr}. Because multi-organ features differ in scale and dimensionality across organs, we apply NT-Xent on CLS-token embeddings rather than raw organ features, ensuring the contrastive signal operates on a unified representation. we use a 2-layer MLP projection head (Linear--BN--ReLU--Linear, hidden/output dimension 256/128) maps each CLS token to an L2-normalized embedding. NT-Xent temperature $\tau$ is tuned over $\{0.1, 0.25, 0.5, 0.75, 1.0\}$ and set to 0.5 based on the downstream group-disease prediction performance.

\paragraph{BYOL.} BYOL~\cite{byol} removes negative pairs entirely for self-supervised representation learning, i.e., an online encoder predicts the target encoder's projection of another view of the same participant, where the target encoder is an exponential moving average (EMA) of the online encoder. Collapse is prevented solely by architectural asymmetry---an additional predictor head on the online branch. In the multi-organ setting, this cross-view prediction directly encourages the online encoder to produce consistent representations regardless of which organ subset is visible. Specifically, both online and target branches use a 2-layer MLP projector (Linear--BN--ReLU--Linear--BN, hidden/output dimension 256/256); the online branch additionally has a 2-layer predictor of the same shape. The symmetric loss is $\mathcal{L} = \tfrac{1}{2}[\text{cosine}(q_1, z_2) + \text{cosine}(q_2, z_1)]$, where $q$ denotes online predictions and $z$ the stop-gradient target projections. The EMA momentum is cosine-scheduled from 0.992 to 1.0.

\paragraph{MoCo-v3.} MoCo-v3~\cite{mocov3} combines SimCLR's contrastive objective with BYOL's momentum encoder, without the memory queue used in earlier MoCo versions~\cite{moco}. In our multi-organ setting, the query branch (with predictor) encodes both the full view and the masked view of each participant, while the momentum branch (no predictor) produces stop-gradient keys from the same two views. Positives are formed \emph{cross-view} across masking conditions (query of the full view vs.\ key of the masked view, and vice versa), and negatives are the keys of all other participants in the batch. This design directly aligns with our goal of learning organ-missingness-invariant representations: the query encoder must produce embeddings that remain close to the momentum-stable target across different organ subsets, while staying discriminative against other participants. For the implementation, the query and momentum branches share a 3-layer MLP projector (hidden/output dimension 256/256), and the query branch has an additional 2-layer predictor (hidden/output dimension 256/256). The InfoNCE temperature is tuned over $\{0.1, 0.2, 0.3, 0.5\}$ and set to 0.2. Momentum is cosine-scheduled from 0.992 to 1.0, which is same with the BYOL baseline.

\paragraph{Barlow Twins.} Barlow Twins~\cite{barlowtwins} avoids both negative pairs and momentum encoders by regularizing the cross-correlation matrix between two views' projections toward the identity: diagonal entries are pushed to 1 (invariance) and off-diagonals to 0 (decorrelation). In our multi-organ setting, the invariance term drives the CLS embedding to be stable under organ-level masking. For the implementation, a 3-layer expander MLP (hidden 512, output 2048) projects CLS tokens into a high-dimensional redundancy-reduction space. Projections are batch-normalized per feature. The off-diagonal weight $\lambda$ is set to $5\mathrm{e}{-3}$.

\paragraph{VICReg.} VICReg~\cite{vicreg} replaces Barlow Twins' cross-correlation with three explicit terms: \emph{invariance} (MSE between view projections), \emph{variance} (hinge loss on per-feature standard deviation to prevent dimensional collapse), and \emph{covariance} (off-diagonal penalty on per-view covariance, which decorrelates features). In our multi-organ setting, the invariance term pulls full-view and masked-view embeddings of the same participant together; the variance term prevents the backbone from collapsing participants with heavy organ missingness to the same embedding; and the covariance term mitigates organ shortcut, analogous to Barlow Twins. Specifically, VICReg shares the same 3-layer expander (hidden 512, output 2048) as Barlow Twins. Loss weights follow the original paper: $\mu = 25.0$, $\nu = 25.0$, $\xi = 1.0$, with variance hinge threshold $\gamma = 1.0$.

Among the SSL baselines, methods with explicit inter-sample discrimination (SimCLR, MoCo-v3, DINOv2) consistently outperform those relying solely on intra-sample invariance or feature regularization (BYOL, Barlow Twins, VICReg) in the multi-organ learning setting. We attribute this gap to the weak view-level augmentation in our setting, i.e., the two views differ only by organ-level masking with no strong appearance transforms, making discrimination-based objectives more robust to shortcut learning than regularization-based ones.


\subsection{Pre-training Details}
\label{appendix:pretrain_details}
We pre-train Pan-FM for 200 epochs on a single NVIDIA V100 GPU (approximately 8 hours) using a 12-layer ViT backbone~\cite{vit} with a hidden dim of 384 and 8 attention heads. Optimization uses AdamW with batch size 64 and mixed-precision (FP16) training. The base learning rate is $10^{-4}$, linearly warmed up over the first 10 epochs and then cosine-decayed to $10^{-6}$. Weight decay is cosine-annealed from 0.04 to 0.4 over the full schedule. The teacher--student architecture uses an 8192-dim projection head with the last layer frozen for the first epoch, an EMA momentum cosine-annealed from 0.992 to 1.0, and a teacher softmax temperature linearly warmed from 0.04 to 0.07 over 10 epochs.

\noindent\textbf{Applying SGM for Various SSL Methods.} To verify that SGM is a general-purpose plug-in rather than an architecture-specific trick, we apply it to four representative SSL paradigms: contrastive (SimCLR), feature-decorrelation (VICReg, Barlow Twins), and self-distillation (DINOv2). For a fair comparison, we keep each baseline’s original hyper-parameters unchanged and only tune SGM’s temperature, setting it to 0.75 for SimCLR, and 0.5 for both VICReg and Barlow Twins. 

\subsection{Downstream Task}
\label{appendix:downstream_evaluation}

\subsubsection{Group-Disease Prediction}

\noindent\textbf{Linear Probing}. We follow a standard linear probing protocol to evaluate the quality of frozen representations. For each downstream disease, we train an independent binary linear probe (Case vs. healthy Control) consisting of a single fully-connected layer mapping the frozen CLS token to a binary logit. The backbone is entirely frozen, ensuring that the evaluation purely reflects the quality of the pre-trained representation. Input modalities are normalized per organ using pre-computed training statistics. To address class imbalance between cases and controls, we replace standard BCE with a focal loss ($\alpha=0.75$, $\gamma=2.0$) for all compared baselines. Each probe is optimized with AdamW by using a learning rate of $1\times10^{-3}$ and a weight decay of $1\times10^{-4}$) for 20 epochs with batch size 256. All baselines and our Pan-FM are evaluated under this same protocol for fair comparison. Each linear probe is trained with 10 different random seeds, and we report the average performance on the held-out test set.

\noindent\textbf{Full Backbone Fine-tuning}. We additionally evaluate each pre-trained backbone under the end-to-end fine-tuning protocol. For each downstream disease, each pre-trained SSL backbone is fine-tuned jointly with a single-layer linear classification head, keeping consistent with the linear probing evaluation. To preserve the pre-trained representations while allowing the task-specific head to adapt freely, we use a smaller learning rate for the backbone ($2\times10^{-6}$) and a larger learning rate for the head ($1\times10^{-3}$). For the Random baseline that uses a randomly initialized backbone, we use a larger learning rate of $1\times10^{-3}$ for backbone training. Training uses AdamW with a linear warmup followed by cosine decay, and the same focal loss ($\alpha=0.75$, $\gamma=2.0$) as in linear probing. Each model is fine-tuned for up to 20 epochs with batch size 32, and the best checkpoint is selected by validation loss with early stopping. We report the average over 5 runs with different random seeds on the held-out test set.

\noindent\textbf{Random organ dropout.} To assess robustness under incomplete organ availability, we design two complementary missing-organ protocols. Both are conducted on the subset of test subjects with complete 7-organ imaging data to ensure a controlled comparison. For each subject, we randomly drop $k \in \{1, 2, 3\}$ available organs at inference. To ensure stable estimates, for the linear probing setting we train 10 independent linear classifiers (different initialization seeds) and, for each classifier, repeat the random dropout 5 times using fixed seeds $\{0, 1, 2, 3, 4\}$, yielding 50 evaluation runs per drop level. For the fine-tuning setting, due to the longer training time and its empirically smaller run-to-run variance, we instead train 5 independent models and repeat the random dropout with the same 5 seeds (25 evaluation runs). \textbf{The same dropout seeds are shared across all compared methods (Pan-FM and SSL baselines), so that every method is evaluated on identical organ-dropout patterns at each run, enabling paired comparisons.} We report the mean performance across all runs. This setting simulates realistic missingness patterns where any subset of organs may be unavailable.

\noindent\textbf{Specific organ dropout.} We systematically remove one organ at a time (across all seven organs: brain, heart, adipose, liver, kidney, spleen, pancreas) to measure each model's sensitivity to the absence of a particular organ. This disentangles the contribution of each organ to the learned representation and reveals whether a model relies disproportionately on any single organ. Results are averaged over the same set of classifier seeds as above (10 for linear probing, 5 for fine-tuning).

\subsubsection{Single-Disease Prediction}
\noindent\textbf{Linear Probing}. In contrast to the 13 disease categories, individual diseases contain substantially fewer positive cases, making a single train/test split unreliable for performance evaluation. We therefore adopt 10-fold stratified cross-validation for single-disease evaluation. For each disease, we construct a case-control pool comprising all available cases across all splits and all healthy controls from the held-out test split, which are then partitioned into 10 stratified folds with balanced case/control ratios. Within each fold, a linear probe is trained on the training folds and evaluated on the held-out validation fold, following the same optimization protocol as described above (AdamW, focal loss, cosine annealing and training epochs). We report the mean AUROC and balanced accuracy across the 10 folds.

\section{More Results}
\label{appendic:more_resulrs}

\begin{table}[t]
\centering
\caption{\textbf{Group-disease evaluation with linear probing.} 
Comparison of SSL methods for group-disease prediction on the held-out test set. Values are reported as AUROC / BalAcc.}
\vspace{-0.15cm}
\label{tab:group_linear_prob_results}
\resizebox{\textwidth}{!}{%
\begin{tabular}{lccccccc}
\toprule
\textbf{Disease Group} & \textbf{SimCLR} \cite{simclr} & \textbf{BYOL} \cite{byol} & \textbf{MoCo-v3} \cite{mocov3} & \textbf{VICReg} \cite{vicreg} & \textbf{Barlow Twins} \cite{barlowtwins} & \textbf{DINOv2} \cite{dinov2} & \textbf{Pan-FM} \\
\cmidrule(lr){1-8}
Infectious \& Parasitic & 0.627 / 0.615 & 0.614 / 0.588 & 0.648 / 0.617 & 0.645 / 0.619 & 0.637 / 0.610 & \underline{0.657} / \underline{0.627} & \textbf{0.668} / \textbf{0.635} \\
Neoplasms & 0.628 / 0.598 & 0.618 / 0.586 & \underline{0.634} / \underline{0.604} & 0.625 / 0.599 & 0.625 / 0.601 & 0.629 / 0.601 & \textbf{0.671} / \textbf{0.643} \\
Blood \& Immune & 0.659 / 0.628 & 0.635 / 0.617 & \textbf{0.672} / \underline{0.632} & 0.635 / 0.618 & 0.646 / 0.624 & 0.657 / 0.628 & \underline{0.666} / \textbf{0.634} \\
Endocrine \& Metabolic & 0.692 / 0.651 & 0.682 / 0.633 & \underline{0.708} / \underline{0.662} & 0.680 / 0.638 & 0.692 / 0.652 & 0.700 / 0.657 & \textbf{0.713} / \textbf{0.666} \\
Mental \& Behavioural & 0.673 / 0.633 & 0.649 / 0.612 & 0.665 / 0.636 & 0.649 / 0.618 & 0.665 / 0.631 & \underline{0.675} / \underline{0.647} & \textbf{0.690} / \textbf{0.648} \\
Nervous System & 0.667 / 0.628 & 0.655 / 0.617 & \underline{0.668} / \underline{0.628} & 0.645 / 0.614 & 0.647 / 0.608 & 0.655 / 0.620 & \textbf{0.678} / \textbf{0.643} \\
Eye & \underline{0.661} / \underline{0.635} & 0.611 / 0.611 & 0.632 / 0.623 & 0.605 / 0.590 & 0.603 / 0.590 & 0.644 / 0.622 & \textbf{0.672} / \textbf{0.636} \\
Circulatory System & 0.688 / 0.647 & 0.660 / 0.619 & \textbf{0.699} / \textbf{0.657} & 0.680 / 0.635 & 0.686 / 0.644 & 0.681 / 0.646 & \underline{0.693} / \underline{0.648} \\
Respiratory System & 0.646 / 0.616 & 0.644 / 0.604 & \underline{0.657} / 0.620 & 0.644 / 0.612 & 0.654 / 0.618 & 0.654 / \underline{0.627} & \textbf{0.667} / \textbf{0.641} \\
Digestive System & \textbf{0.638} / \underline{0.603} & 0.615 / 0.581 & 0.632 / 0.599 & 0.611 / 0.590 & 0.624 / 0.594 & 0.619 / 0.597 & \underline{0.635} / \textbf{0.606} \\
Skin System & 0.629 / 0.608 & 0.616 / 0.594 & \underline{0.630} / \underline{0.614} & 0.622 / 0.606 & 0.622 / 0.604 & 0.621 / 0.599 & \textbf{0.671} / \textbf{0.637} \\
Musculoskeletal & 0.653 / \underline{0.619} & 0.639 / 0.599 & \textbf{0.661} / 0.619 & 0.640 / 0.605 & 0.643 / 0.603 & 0.644 / 0.615 & \underline{0.659} / \textbf{0.624} \\
Genitourinary & 0.636 / 0.607 & 0.633 / 0.610 & 0.647 / 0.617 & 0.644 / 0.611 & \underline{0.648} / \underline{0.618} & 0.635 / 0.610 & \textbf{0.659} / \textbf{0.628} \\
\midrule
\textbf{Overall Mean} & 0.653 / 0.622 & 0.636 / 0.606 & \underline{0.658} / \underline{0.625} & 0.640 / 0.612 & 0.646 / 0.615 & 0.652 / 0.623 & \textbf{0.672} / \textbf{0.638} \\
\bottomrule
\end{tabular}%
}
\end{table}

\begin{table}[t]
\vspace{-0.3cm}
\centering
\caption{\textbf{Missing-organ robustness evaluation with linear probing.} Comparison of SSL methods across 13 disease groups on the held-out test set (AUROC / BalAcc). \textit{Full Organs (7)}, \textit{Drop 1 Organ}, \textit{Drop 2 Organs}, and \textit{Drop 3 Organs} settings are evaluated on subjects with complete 7-organ imaging data in the held-out test set, with random organ dropout repeated over 50 runs to ensure stable and comparable assessment. $\Delta$ denotes the relative improvement of Pan-Organ over the DINOv2 baseline.}
\label{tab:group_linear_prob_results_miss}
\vspace{-0.15cm}
\resizebox{\textwidth}{!}{%
\begin{tabular}{l cc cc cc cc cc}
\toprule
\multirow{2}{*}{Method} & \multicolumn{2}{c}{Standard} & \multicolumn{2}{c}{Full Organs (7)} & \multicolumn{2}{c}{Drop 1 Organ} & \multicolumn{2}{c}{Drop 2 Organs} & \multicolumn{2}{c}{Drop 3 Organs} \\
\cmidrule(lr){2-3} \cmidrule(lr){4-5} \cmidrule(lr){6-7} \cmidrule(lr){8-9} \cmidrule(lr){10-11}
 & AUROC & BalAcc & AUROC & BalAcc & AUROC & BalAcc & AUROC & BalAcc & AUROC & BalAcc \\
\midrule
SimCLR~\cite{simclr}        & 0.653 & 0.622 & 0.659 & 0.637 & 0.645 & 0.629 & 0.638 & 0.627 & 0.625 & 0.617 \\
BYOL~\cite{byol}          & 0.636 & 0.606 & 0.647 & 0.634 & 0.642 & 0.630 & 0.645 & 0.632 & 0.628 & 0.621 \\
MoCo-v3~\cite{mocov3}       & 0.658 & 0.625 & 0.660 & 0.642 & 0.653 & 0.636 & 0.655 & 0.641 & 0.642 & 0.630 \\
VICReg~\cite{vicreg}      & 0.640 & 0.612 & 0.647 & 0.631 & 0.640 & 0.626 & 0.639 & 0.627 & 0.624 & 0.619 \\
Barlow Twins~\cite{barlowtwins}  & 0.646 & 0.615 & 0.642 & 0.624 & 0.632 & 0.620 & 0.632 & 0.622 & 0.618 & 0.613 \\
DINOv2~\cite{dinov2}        & 0.652 & 0.623 & 0.683 & 0.654 & 0.654 & 0.642 & 0.647 & 0.639 & 0.617 & 0.616 \\
\textbf{Pan-FM (Ours)}  & \textbf{0.672} & \textbf{0.638} & \textbf{0.694} & \textbf{0.670} & \textbf{0.681} & \textbf{0.662} & \textbf{0.672} & \textbf{0.655} & \textbf{0.654} & \textbf{0.638} \\
\midrule
$\Delta$ (\%) & +3.1\% & +2.4\% & +1.6\% & +2.4\% & +4.1\% & +3.1\% & +3.9\% & +2.5\% & +6.0\% & +3.6\% \\
\bottomrule
\end{tabular}%
}
\end{table}

\begin{table}[t]
\centering
\caption{\textbf{Group-disease evaluation with full backbone fine-tuning.} Comparison of SSL methods for group-disease prediction on the held-out test set. Values are reported as AUROC / BalAcc.}
\vspace{-0.15cm}
\label{tab:group_ft_results}
\resizebox{\textwidth}{!}{
\begin{tabular}{lccccccc}
\toprule
\textbf{Disease Group} & \textbf{Random} & \textbf{SimCLR} \cite{simclr} & \textbf{MoCo-v3} \cite{mocov3} & \textbf{VICReg} \cite{vicreg} & \textbf{Barlow Twins} \cite{barlowtwins} & \textbf{DINOv2} \cite{dinov2} & \textbf{Pan-FM} \\
\cmidrule(lr){1-8}
Infectious \& Parasitic & 0.638 / \underline{0.635} & 0.629 / 0.614 & 0.634 / 0.610 & 0.642 / 0.612 & 0.621 / 0.604 & \underline{0.672} / 0.630 & \textbf{0.694} / \textbf{0.656} \\
Neoplasms & 0.605 / 0.589 & 0.629 / 0.601 & 0.638 / 0.612 & 0.629 / 0.596 & 0.637 / 0.608 & \underline{0.646} / \underline{0.612} & \textbf{0.686} / \textbf{0.646} \\
Blood \& Immune & 0.628 / 0.608 & 0.660 / 0.628 & \textbf{0.674} / \textbf{0.636} & 0.636 / 0.609 & 0.650 / 0.621 & 0.656 / 0.630 & \underline{0.669} / \underline{0.632} \\
Endocrine \& Metabolic & 0.674 / 0.633 & 0.703 / 0.655 & \underline{0.712} / \underline{0.669} & 0.694 / 0.655 & 0.701 / 0.659 & 0.710 / 0.661 & \textbf{0.718} / \textbf{0.673} \\
Mental \& Behavioural & 0.649 / 0.629 & 0.657 / 0.613 & 0.657 / 0.627 & 0.644 / 0.616 & 0.656 / 0.633 & \underline{0.680} / \textbf{0.654} & \textbf{0.683} / \underline{0.638} \\
Nervous System & 0.636 / 0.619 & 0.664 / 0.624 & 0.665 / 0.623 & 0.654 / \underline{0.640} & 0.653 / 0.618 & \underline{0.665} / 0.628 & \textbf{0.681} / \textbf{0.652} \\
Eye & 0.615 / 0.610 & 0.653 / 0.630 & 0.631 / 0.616 & 0.622 / 0.610 & 0.622 / 0.596 & \underline{0.661} / \underline{0.631} & \textbf{0.674} / \textbf{0.635} \\
Circulatory System & 0.654 / 0.611 & 0.688 / 0.637 & \underline{0.702} / \textbf{0.663} & 0.691 / 0.653 & 0.692 / 0.647 & 0.702 / \underline{0.661} & \textbf{0.704} / 0.656 \\
Respiratory System & 0.607 / 0.593 & 0.649 / 0.612 & 0.663 / 0.625 & 0.652 / 0.624 & 0.660 / 0.624 & \underline{0.673} / \underline{0.643} & \textbf{0.678} / \textbf{0.645} \\
Digestive System & 0.611 / 0.592 & 0.633 / 0.601 & 0.636 / 0.606 & 0.617 / 0.591 & 0.626 / 0.595 & \underline{0.642} / \underline{0.609} & \textbf{0.654} / \textbf{0.610} \\
Skin System & 0.623 / 0.605 & 0.623 / \underline{0.609} & 0.623 / 0.605 & 0.621 / 0.606 & 0.620 / 0.596 & \underline{0.630} / 0.602 & \textbf{0.666} / \textbf{0.629} \\
Musculoskeletal & 0.641 / 0.607 & 0.650 / 0.615 & 0.664 / \underline{0.624} & 0.644 / 0.611 & 0.647 / 0.610 & \underline{0.667} / 0.622 & \textbf{0.669} / \textbf{0.627} \\
Genitourinary & 0.623 / 0.602 & 0.637 / 0.605 & 0.642 / 0.610 & 0.642 / 0.606 & 0.650 / 0.623 & \underline{0.653} / \underline{0.627} & \textbf{0.673} / \textbf{0.634} \\
\midrule
\textbf{Overall Mean}  & 0.631 / 0.610 & 0.652 / 0.619 & 0.657 / 0.625 & 0.645 / 0.618 & 0.649 / 0.618 & \underline{0.666} / \underline{0.632} & \textbf{0.681} / \textbf{0.641} \\
\bottomrule
\end{tabular}
}
\end{table}

\begin{table}[t]
\centering
\caption{\textbf{Missing-organ robustness evaluation with full backbone fine-tuning.} Comparison of SSL methods  across 13 disease groups on the held-out test set (AUROC / BalAcc). \emph{Full Organs (7)}, \emph{Drop 1 Organ}, and \emph{Drop 3 Organs} settings are evaluated on subjects with complete 7-organ imaging data in the held-out test set, with random organ dropout repeated over 25 runs to ensure stable and comparable assessment. $\Delta$ denotes the relative improvement of Pan-FM over the DINOv2 baseline.}
\label{tab:group_ft_results_miss}
\vspace{-0.15cm}
\resizebox{\columnwidth}{!}{
\begin{tabular}{l cc cc @{\hskip 1.2em} cc cc cc}
\toprule
\multirow{2}{*}{Method}
 & \multicolumn{2}{c}{Standard} & \multicolumn{2}{c}{Full Organs (7)}
 & \multicolumn{2}{c}{Drop 1 Organ} & \multicolumn{2}{c}{Drop 2 Organs} & \multicolumn{2}{c}{Drop 3 Organs} \\
\cmidrule(lr){2-3}\cmidrule(lr){4-5}\cmidrule(lr){6-7}\cmidrule(lr){8-9}\cmidrule(lr){10-11}
 & AUROC & BalAcc & AUROC & BalAcc & AUROC & BalAcc & AUROC & BalAcc & AUROC & BalAcc \\
\midrule
SimCLR~\cite{simclr}
  & 0.652 & 0.619 & 0.664 & 0.643
  & 0.652 & 0.635 & 0.645 & 0.630 & 0.636 & 0.624 \\
BYOL~\cite{byol}
  & 0.640 & 0.607 & 0.653 & 0.638
  & 0.645 & 0.632 & 0.648 & 0.633 & 0.634 & 0.624 \\
MoCo-v3~\cite{mocov3}
  & 0.657 & 0.625 & 0.667 & 0.649
  & 0.658 & 0.641 & 0.662 & 0.645 & \underline{0.647} & \underline{0.634} \\
VICReg~\cite{vicreg}
  & 0.645 & 0.618 & 0.650 & 0.628
  & 0.644 & 0.627 & 0.642 & 0.629 & 0.625 & 0.619 \\
Barlow Twins~\cite{barlowtwins}
  & 0.649 & 0.618 & 0.643 & 0.626
  & 0.633 & 0.621 & 0.630 & 0.622 & 0.619 & 0.616 \\
DINOv2~\cite{dinov2}
  & \underline{0.666} & \underline{0.632} & \underline{0.691} & \underline{0.662}
  & \underline{0.681} & \underline{0.660} & \underline{0.668} & \underline{0.650} & 0.645 & 0.632 \\
\textbf{Pan-FM (Ours)}
  & \textbf{0.681} & \textbf{0.641} & \textbf{0.704} & \textbf{0.675}
  & \textbf{0.696} & \textbf{0.672} & \textbf{0.687} & \textbf{0.662} & \textbf{0.675} & \textbf{0.650} \\
\midrule
$\Delta$ (\%)
  & $+2.3\%$ & $+1.5\%$ & $+2.0\%$ & $+1.9\%$
  & $+2.1\%$ & $+1.8\%$ & $+2.7\%$ & $+1.8\%$ & $+4.6\%$ & $+2.9\%$ \\
\bottomrule
\end{tabular}
}
\end{table}

\subsection{Group-Disease Prediction}
\textbf{Linear Probing.} As shown in Table~\ref{tab:group_linear_prob_results}, Pan-FM achieves the best overall performance, with a mean AUROC of 0.672 and BalAcc of 0.638. The gains are especially pronounced on systemic conditions such as \textit{Neoplasms}, \textit{Skin System}, and \textit{Metabolic}, where Pan-FM substantially outperforms the DINOv2 baseline, e.g., 0.671 vs.\ 0.629 on Neoplasms, and 0.671 vs.\ 0.621 on Skin System. Since Pan-FM and DINOv2 share identical backbones and training configurations, this improvement directly demonstrates the effectiveness of our proposed SGM in capturing systemic pan-organ representations.

\textbf{Missing Robustness w/  Linear-Probing.} Table~\ref{tab:group_linear_prob_results_miss} reports missing-organ robustness under linear probing. \emph{Standard} uses the full test set with natural missingness; the other four settings apply controlled dropout on the complete-7-organ subset. As more organs are dropped, all baselines degrade substantially, while Pan-FM degrades gracefully and consistently leads, with its margin over DINOv2 growing from +1.6\% (Full) to +6.0\% (Drop-3). Several baselines (BYOL, MoCo-v3) even degrade non-monotonically from Drop-1 to Drop-2, as they rely on a few dominant organs whose redundant counterparts can be dropped without loss (Fig.~\ref{fig:organ_removal}). Pan-FM instead leads in 6/7 single-organ dropout settings, peaking at +6.3\% over DINOv2 on \emph{w/o Adipose}, confirming that SGM yields balanced pan-organ representations robust to organ absence.

\textbf{Full-Backbone Fine-tuning.} Table~\ref{tab:group_ft_results} reports group-disease prediction results when the entire backbone is fine-tuned together with the linear head. Pan-FM still achieves the best overall performance (0.681/0.641), outperforming the strongest SSL baseline DINOv2 (0.666/0.632) and substantially surpassing the randomly initialized backbone (0.631/0.610), which demonstrates the effectiveness of multi-organ SSL. In addition, DINOv2 shows a larger relative gain from linear probing to full fine-tuning than other SSL baselines, suggesting that its representations encode transferable features that particularly benefit for fine-tuning. Together with the linear-probing results in Table~\ref{tab:group_linear_prob_results}, these results demonstrate that the representations learned by SGM are beneficial both as frozen features and as initialization for end-to-end adaptation.

\textbf{Missing Robustness w/ Backbone Fine-tuning.} 
Table~\ref{tab:group_ft_results_miss} and Figure~\ref{fig:organ_removal_ft} assess test-time organ removal after backbone fine-tuning. Pan-FM leads at every missingness level, and its margin over DINOv2 scales with the number of dropped organs (+2.0\% at Full Organs, +4.6\% at Drop 3). The radar chart in Figure~\ref{fig:organ_removal_ft}(a) indicates no weak direction, and the pairwise heatmap in (b) localizes DINOv2's failure modes to \emph{Adipose} and \emph{Pancreas}, where Pan-FM remains stable. Compared with linear probing, full backbone fine-tuning recovers additional accuracy under severe missingness (0.675 vs.\ 0.654 AUROC at Drop 3), as trainable parameters can reallocate capacity to observed organs; however, the DINOv2–Pan-FM gap persists in both regimes, isolating SGM as the source of robustness rather than the adaptation protocol.

\begin{table}[t]
\centering
\caption{\textbf{Group-disease evaluation with linear probing.} Adding \textbf{our proposed SGM} as a plug-in consistently improves four representative SSL methods on the held-out test set across 13 disease groups. Values are AUROC / BalAcc; bold marks the better result within each (base, +SGM) pair.}
\vspace{-0.15cm}
\label{tab:group_linear_prob_results_w_sgm}
\setlength{\tabcolsep}{4pt}
\renewcommand{\arraystretch}{0.95}
\resizebox{\textwidth}{!}{%
\begin{tabular}{l cc cc cc cc}
\toprule
\multirow{2}{*}{\textbf{Disease Group}}
 & \multicolumn{2}{c}{\textbf{SimCLR}~\cite{simclr}}
 & \multicolumn{2}{c}{\textbf{VICReg}~\cite{vicreg}}
 & \multicolumn{2}{c}{\textbf{Barlow Twins}~\cite{barlowtwins}}
 & \multicolumn{2}{c}{\textbf{DINOv2}~\cite{dinov2}} \\
\cmidrule(lr){2-3} \cmidrule(lr){4-5} \cmidrule(lr){6-7} \cmidrule(lr){8-9}
 & Base & +SGM & Base & +SGM & Base & +SGM & Base & +SGM (Pan-FM) \\
\midrule
Infectious \& Parasitic   & 0.627/0.615 & \textbf{0.638}/\textbf{0.616} & 0.645/\textbf{0.619} & \textbf{0.651}/0.618 & 0.637/0.610 & \textbf{0.658}/\textbf{0.648} & 0.657/0.627 & \textbf{0.668}/\textbf{0.635} \\
Neoplasms                 & 0.628/0.598 & \textbf{0.647}/\textbf{0.616} & 0.625/\textbf{0.599} & \textbf{0.637}/0.597 & 0.625/0.601 & \textbf{0.651}/\textbf{0.612} & 0.629/0.601 & \textbf{0.671}/\textbf{0.643} \\
Blood \& Immune           & 0.659/0.628 & \textbf{0.687}/\textbf{0.665} & 0.635/0.618 & \textbf{0.665}/\textbf{0.627} & 0.646/0.624 & \textbf{0.653}/\textbf{0.628} & 0.657/0.628 & \textbf{0.666}/\textbf{0.634} \\
Endocrine \& Metabolic    & 0.692/0.651 & \textbf{0.714}/\textbf{0.666} & 0.680/0.638 & \textbf{0.703}/\textbf{0.655} & 0.692/0.652 & \textbf{0.715}/\textbf{0.672} & 0.700/0.657 & \textbf{0.713}/\textbf{0.666} \\
Mental \& Behavioural     & \textbf{0.673}/\textbf{0.633} & 0.672/0.628 & 0.649/0.618 & \textbf{0.692}/\textbf{0.645} & 0.665/0.631 & \textbf{0.671}/\textbf{0.640} & 0.675/0.647 & \textbf{0.690}/\textbf{0.648} \\
Nervous System            & 0.667/0.628 & \textbf{0.687}/\textbf{0.650} & 0.645/0.614 & \textbf{0.674}/\textbf{0.642} & 0.647/0.608 & \textbf{0.677}/\textbf{0.641} & 0.655/0.620 & \textbf{0.678}/\textbf{0.643} \\
Eye                       & 0.661/0.635 & \textbf{0.664}/\textbf{0.637} & 0.605/0.590 & \textbf{0.658}/\textbf{0.641} & 0.603/0.590 & \textbf{0.672}/\textbf{0.653} & 0.644/0.622 & \textbf{0.672}/\textbf{0.636} \\
Circulatory System        & 0.688/0.647 & \textbf{0.707}/\textbf{0.648} & 0.680/0.635 & \textbf{0.702}/\textbf{0.650} & 0.686/0.644 & \textbf{0.705}/\textbf{0.655} & 0.681/0.646 & \textbf{0.693}/\textbf{0.648} \\
Respiratory System        & 0.646/0.616 & \textbf{0.670}/\textbf{0.634} & 0.644/0.612 & \textbf{0.659}/\textbf{0.628} & 0.654/0.618 & \textbf{0.662}/\textbf{0.621} & 0.654/0.627 & \textbf{0.667}/\textbf{0.641} \\
Digestive System          & 0.638/0.603 & \textbf{0.655}/\textbf{0.617} & 0.611/0.590 & \textbf{0.643}/\textbf{0.605} & 0.624/0.594 & \textbf{0.646}/\textbf{0.612} & 0.619/0.597 & \textbf{0.635}/\textbf{0.606} \\
Skin System               & 0.629/0.608 & \textbf{0.650}/\textbf{0.623} & 0.622/0.606 & \textbf{0.640}/\textbf{0.618} & 0.622/0.604 & \textbf{0.626}/\textbf{0.611} & 0.621/0.599 & \textbf{0.671}/\textbf{0.637} \\
Musculoskeletal           & 0.653/0.619 & \textbf{0.675}/\textbf{0.629} & 0.640/0.605 & \textbf{0.672}/\textbf{0.640} & 0.643/0.603 & \textbf{0.678}/\textbf{0.625} & 0.644/0.615 & \textbf{0.659}/\textbf{0.624} \\
Genitourinary             & 0.636/0.607 & \textbf{0.654}/\textbf{0.623} & 0.644/0.611 & \textbf{0.646}/\textbf{0.618} & 0.648/\textbf{0.618} & \textbf{0.649}/0.615 & 0.635/0.610 & \textbf{0.659}/\textbf{0.628} \\
\midrule
\textbf{Overall Mean}     & 0.653/0.622 & \textbf{0.671}/\textbf{0.635} & 0.640/0.612 & \textbf{0.665}/\textbf{0.630} & 0.646/0.615 & \textbf{0.666}/\textbf{0.633} & 0.652/0.623 & \textbf{0.672}/\textbf{0.638} \\
\bottomrule
\end{tabular}%
}
\end{table}

\subsection{Applying SGM for More SSL Methods}
To verify that SGM is a general-purpose plug-in rather than an architecture-specific trick, we apply it to four representative SSL paradigms: contrastive (SimCLR), feature-decorrelation (VICReg, Barlow Twins), and self-distillation (DINOv2). For a fair comparison, we keep each baseline's original hyper-parameters unchanged and only tune SGM's temperature, setting it to 0.75 for SimCLR, and 0.5 for both VICReg and Barlow Twins. As shown in Table~\ref{tab:group_linear_prob_results_w_sgm}, SGM consistently improves all four baselines, lifting mean AUROC by +1.8 to +2.5 absolute points and winning 96\% (100/104) of head-to-head per-disease comparisons. Notably, SGM benefits are broadly distributed across disease groups rather than concentrated on a few, i.e., every disease except SimCLR-Mental sees consistent gains. Our final model, Pan-FM (DINOv2 + SGM), attains the best overall performance (0.672 AUROC / 0.638 BalAcc), confirming that SGM compounds with the strongest available backbone. These results demonstrate that SGM is a method-agnostic plug-in that reliably enhances representation quality across both contrastive (SimCLR), feature-decorrelation (VICReg, Barlow Twins), and self-distillation (DINOv2) paradigms.

\subsection{Single Organ Dropout}
\begin{figure}[t]
\begin{center}
   \includegraphics[width=1.0\linewidth]{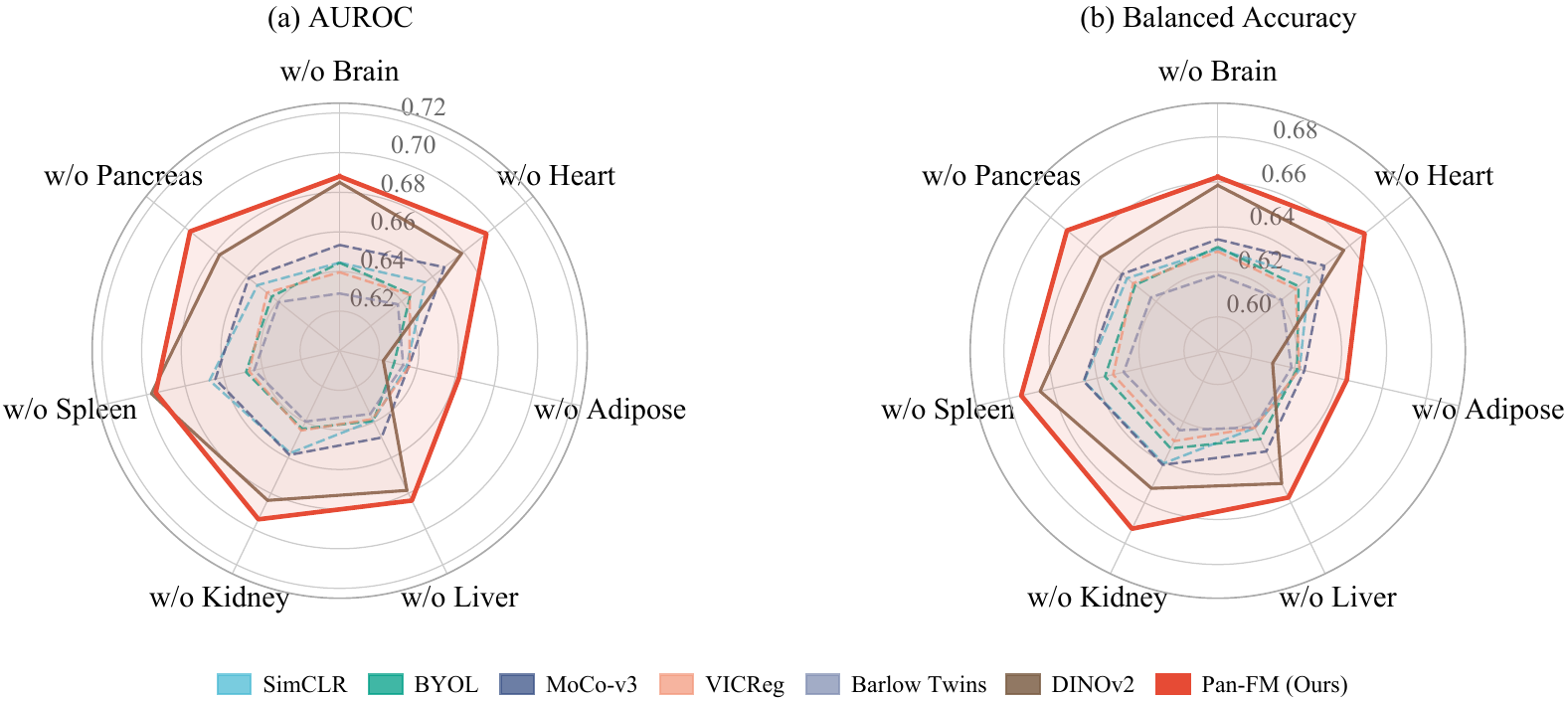} 
\end{center}
\vspace{-0.3cm}
 \caption{Robustness evaluation under specific organ dropout (test-time organ removal) with \textbf{linear probing}.}
\label{fig:organ_removal_linear_balacc}
\vspace{-0.25cm}
\end{figure}

\textbf{Missing Robustness Evaluation under linear probing}. Figure~\ref{fig:organ_removal_linear_balacc} reports AUROC and Balanced Accuracy when each of the seven organs is removed individually at test time, evaluated on subjects with complete 7-organ data. Across all seven removal scenarios, Pan-FM (red curve) consistently dominates the radar plot, forming the outermost envelope on both metrics. The gap between Pan-FM and the strongest competing baseline is most pronounced when removing organs that carry rich diagnostic signal (e.g., \emph{Heart}, \emph{Spleen}, \emph{Kidney}, and \emph{Adipose}), where existing SSL methods drop sharply, while Pan-FM remains stable. This indicates that the cross-organ token interactions learned through SGM enable Pan-FM to recover predictive information from the remaining organs rather than relying on any single modality. In contrast, baselines such as VICReg and Barlow Twins exhibit visibly contracted polygons, reflecting their fragility to organ unavailability, which is a realistic bottleneck in clinical deployment.

\begin{figure}[t]
\begin{center}
   \includegraphics[width=1.0\linewidth]{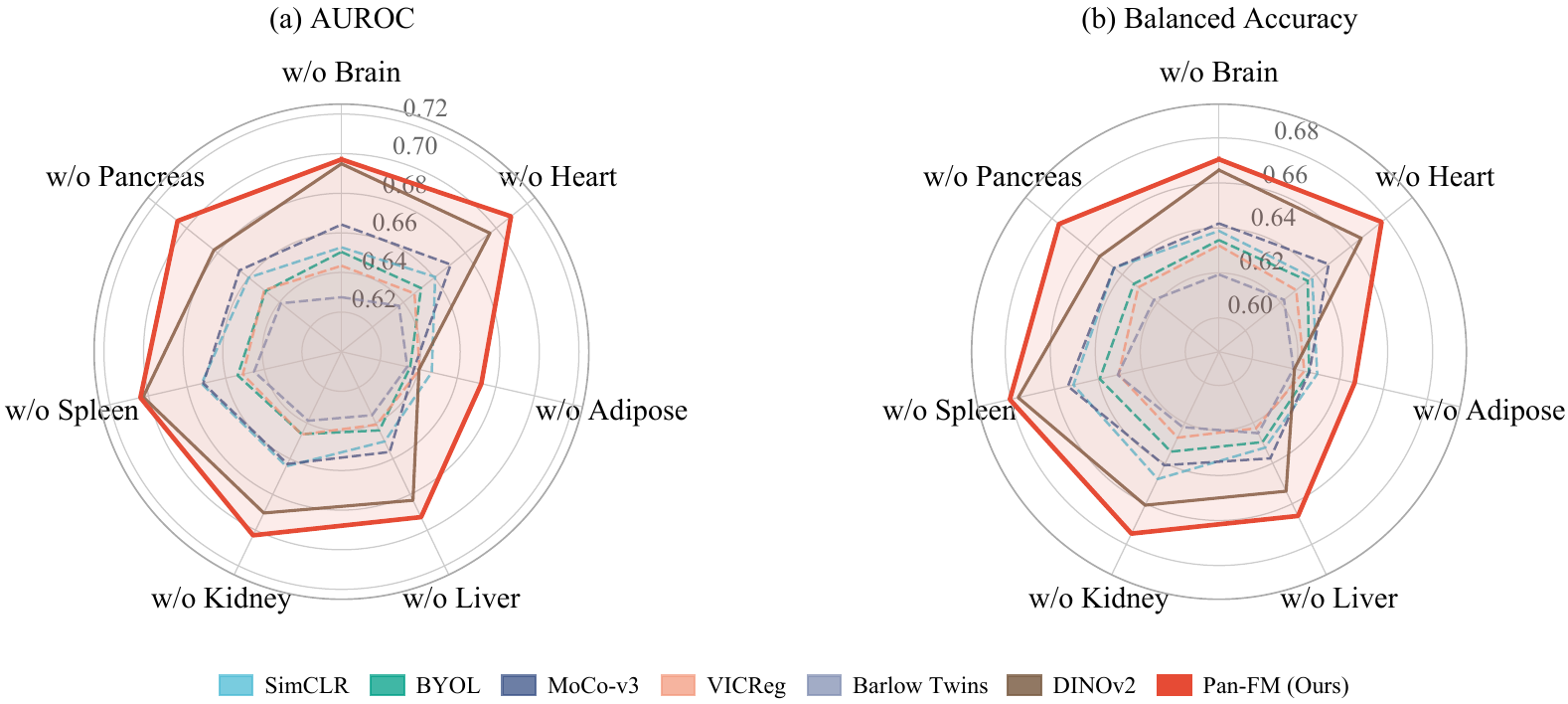} 
\end{center}
\vspace{-0.35cm}
 \caption{Robustness evaluation under specific organ dropout (test-time organ removal) with \textbf{full backbone fine-tuning}.}
\label{fig:organ_removal_ft_balacc}
\end{figure}

\textbf{Missing Robustness under full backbone fine-tuning}. Figure~\ref{fig:organ_removal_ft_balacc} shows that Pan-FM's robustness advantage persists, and is in fact \emph{amplified}, after full backbone fine-tuning. Compared to linear probing, all methods improve in absolute terms once the backbone is fine-tuned, yet the relative ranking remains unchanged: Pan-FM forms the outermost envelope across all seven organ-removal conditions on both AUROC and Balanced Accuracy. The persistent gap suggests that Pan-FM's robustness stems from its pre-trained representations rather than from supervised adaptation, i.e., fine-tuning cannot close the gap for baselines, indicating that SGM injects an inductive bias toward missing-organ generalization that is hard to acquire from labels alone.

Overall, Figs~\ref{fig:organ_removal_linear_balacc} and~\ref{fig:organ_removal_ft_balacc} demonstrate that Pan-FM delivers reliable performance under realistic missing-organ conditions.

\subsection{Pairwise Organ Dropout}
\label{sec:pairwise_organ_dropout_complete}
\begin{figure}[t]
\begin{center}
   \includegraphics[width=1.0\linewidth]{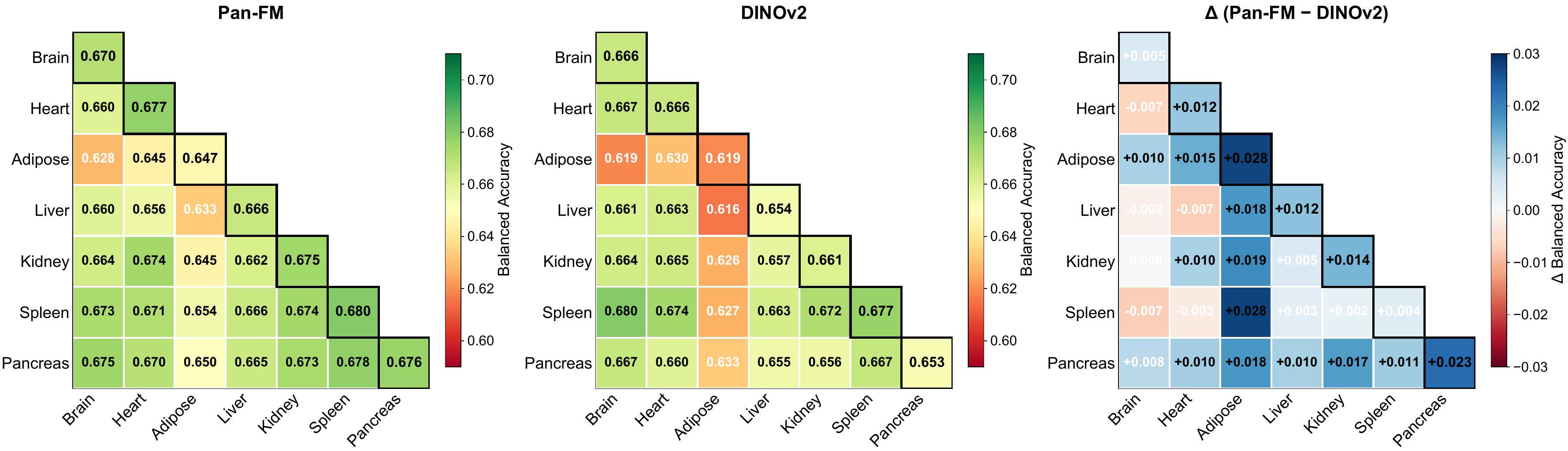} 
\end{center}
 \caption{Pairwise organ dropout heatmaps under \textbf{full backbone fine-tuning} for our Pan-FM and DINOv2. Each cell shows the mean \textbf{balanced accuracy}  across 13 group diseases when the corresponding row and column organs are simultaneously removed. Diagonal entries correspond to single-organ removal.}
\label{fig:organ_removal_ft_balacc_heat}
\end{figure}

\begin{figure}[t]
\begin{center}
   \includegraphics[width=1.0\linewidth]{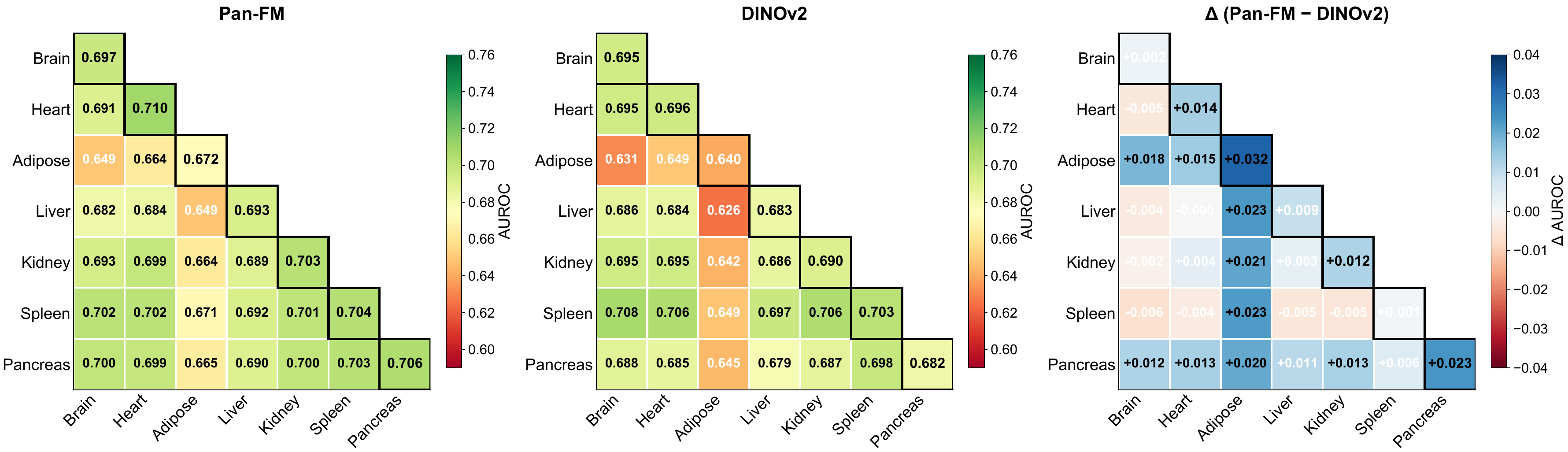} 
\end{center}
 \caption{Pairwise organ dropout heatmaps under \textbf{full backbone fine-tuning} for our Pan-FM and DINOv2. Each cell shows the mean \textbf{AUROC}  across 13 group diseases when the corresponding row and column organs are simultaneously removed. Diagonal entries correspond to single-organ removal.}
\label{fig:organ_removal_ft_auroc_heat}
\end{figure}

\begin{figure}[t]
\begin{center}
   \includegraphics[width=1.0\linewidth]{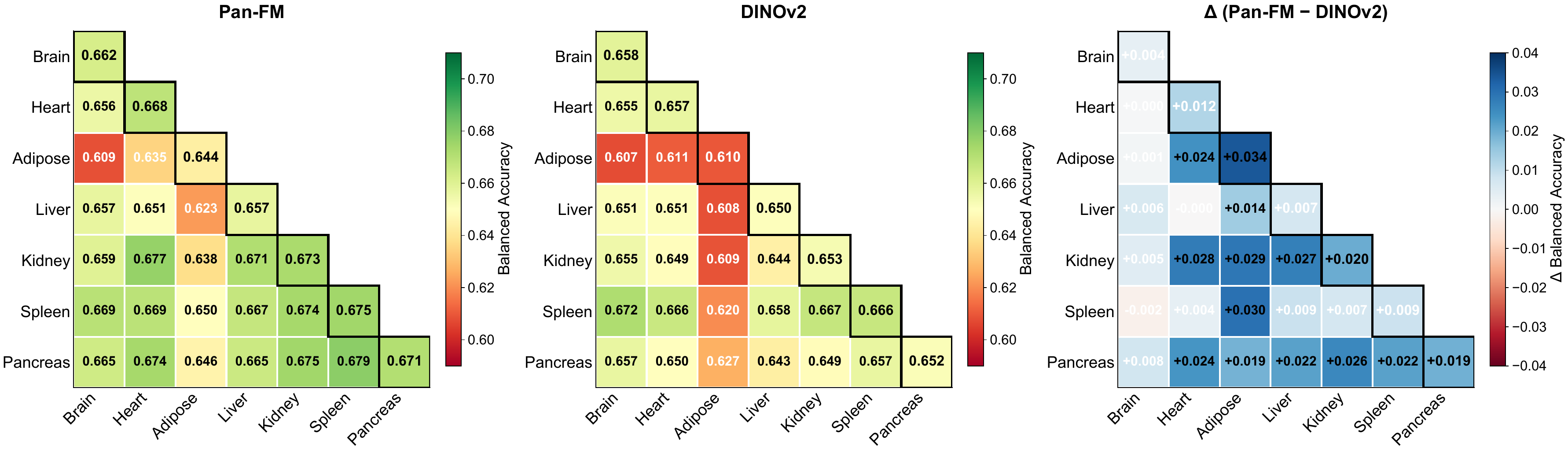} 
\end{center}
 \caption{Pairwise organ dropout heatmaps under \textbf{linear probing} for our Pan-FM and DINOv2. Each cell shows the mean \textbf{balanced accuracy}  across 13 group diseases when the corresponding row and column organs are simultaneously removed. Diagonal entries correspond to single-organ removal.}
\label{fig:organ_removal_linear_balacc_heat}
\end{figure}

\begin{figure}[t]
\begin{center}
   \includegraphics[width=1.0\linewidth]{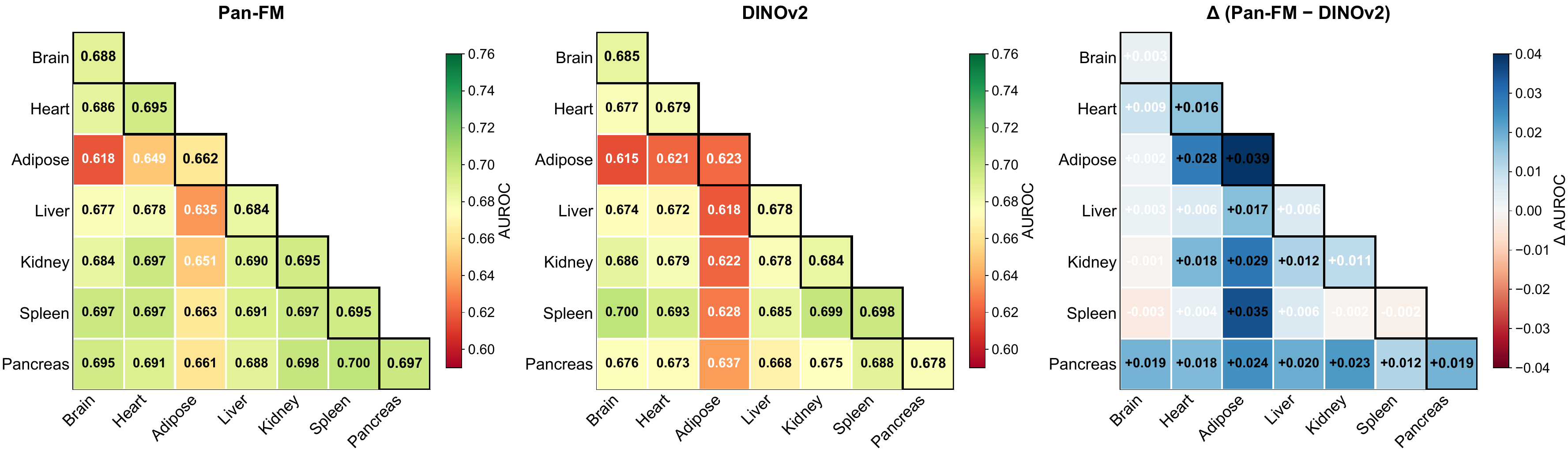} 
\end{center}
 \caption{Pairwise organ dropout heatmaps under \textbf{linear probing} for our Pan-FM and DINOv2. Each cell shows the mean \textbf{AUROC}  across 13 group diseases when the corresponding row and column organs are simultaneously removed. Diagonal entries correspond to single-organ removal.}
\label{fig:organ_removal_linear_auroc_heat}
\end{figure}

To see how the models behave when more than one organ is missing at test time, we remove organs two at a time and evaluate the probes on patients who originally have all seven organs available. There are $\binom{7}{2}=21$ pairs, and together with the 7 single-organ removals they form the lower-triangular heatmaps in Figs~\ref{fig:organ_removal_ft_balacc_heat}--\ref{fig:organ_removal_linear_auroc_heat}. Specifically, off-diagonal cells correspond to the masked organ pairs, diagonal cells indicate the single-organ dropout, and each cell reports the mean across the 13 diseases. Pan-FM's heatmaps look nearly uniform under both linear probing and full backbone fine-tuning, demonstrating that no pair organs is a critical bottleneck. For DINOv2, it has clear weak spots, i.e., organ drop pairs involving adipose degrade the performance significantly, suggesting that it relies heavily on a few organs and suffers disproportionately when they go missing.

\CUT{
\subsection{Comparison with Missing-Modality Baselines}
}

\begin{table}[t]
\centering
\caption{{Group-disease comparison under the Drop-3 Organs setting (linear probing).} Results are aggregated over 50 runs. $\Delta$ denotes the relative improvement of Pan-FM over the DINOv2 baseline.}
\label{tab:drop3_per_disease}
\resizebox{\textwidth}{!}{%
\begin{tabular}{l cc c cc c}
\toprule
\multirow{2}{*}{Disease} & \multicolumn{3}{c}{AUROC} & \multicolumn{3}{c}{Balanced Accuracy} \\
\cmidrule(lr){2-4} \cmidrule(lr){5-7}
 & DINOv2 & Pan-FM (Ours) & $\Delta$ (\%) & DINOv2 & Pan-FM (Ours) & $\Delta$ (\%) \\
\midrule
Infectious \& Parasitic     & \textbf{0.648\,\scriptsize$\pm$0.006} & 0.618\,\scriptsize$\pm$0.014 & $-4.57$ & \textbf{0.633\,\scriptsize$\pm$0.009} & 0.616\,\scriptsize$\pm$0.006 & $-2.66$ \\
Neoplasms                   & 0.584\,\scriptsize$\pm$0.006 & \textbf{0.629\,\scriptsize$\pm$0.006} & $+7.78$ & 0.592\,\scriptsize$\pm$0.004 & \textbf{0.618\,\scriptsize$\pm$0.006} & $+4.39$ \\
Blood \& Immune             & 0.601\,\scriptsize$\pm$0.013 & \textbf{0.659\,\scriptsize$\pm$0.010} & $+9.73$ & 0.609\,\scriptsize$\pm$0.012 & \textbf{0.644\,\scriptsize$\pm$0.012} & $+5.76$ \\
Endocrine \& Metabolic      & 0.659\,\scriptsize$\pm$0.003 & \textbf{0.725\,\scriptsize$\pm$0.004} & $+9.99$ & 0.649\,\scriptsize$\pm$0.004 & \textbf{0.677\,\scriptsize$\pm$0.003} & $+4.32$ \\
Mental \& Behavioural       & 0.634\,\scriptsize$\pm$0.010 & \textbf{0.686\,\scriptsize$\pm$0.012} & $+8.28$ & 0.634\,\scriptsize$\pm$0.008 & \textbf{0.667\,\scriptsize$\pm$0.011} & $+5.18$ \\
Nervous System              & 0.606\,\scriptsize$\pm$0.006 & \textbf{0.668\,\scriptsize$\pm$0.011} & $+10.25$ & 0.609\,\scriptsize$\pm$0.005 & \textbf{0.653\,\scriptsize$\pm$0.009} & $+7.20$ \\
Eye                         & \textbf{0.622\,\scriptsize$\pm$0.009} & 0.611\,\scriptsize$\pm$0.016 & $-1.79$ & 0.628\,\scriptsize$\pm$0.008 & \textbf{0.632\,\scriptsize$\pm$0.012} & $+0.60$ \\
Circulatory System          & 0.630\,\scriptsize$\pm$0.003 & \textbf{0.677\,\scriptsize$\pm$0.005} & $+7.47$ & 0.623\,\scriptsize$\pm$0.003 & \textbf{0.643\,\scriptsize$\pm$0.005} & $+3.16$ \\
Respiratory System          & 0.627\,\scriptsize$\pm$0.006 & \textbf{0.657\,\scriptsize$\pm$0.006} & $+4.84$ & 0.613\,\scriptsize$\pm$0.004 & \textbf{0.638\,\scriptsize$\pm$0.006} & $+4.05$ \\
Digestive System            & 0.592\,\scriptsize$\pm$0.003 & \textbf{0.624\,\scriptsize$\pm$0.004} & $+5.55$ & 0.599\,\scriptsize$\pm$0.004 & \textbf{0.613\,\scriptsize$\pm$0.004} & $+2.36$ \\
Skin                        & 0.612\,\scriptsize$\pm$0.006 & \textbf{0.659\,\scriptsize$\pm$0.005} & $+7.65$ & 0.606\,\scriptsize$\pm$0.003 & \textbf{0.643\,\scriptsize$\pm$0.006} & $+6.14$ \\
Musculoskeletal System      & 0.602\,\scriptsize$\pm$0.003 & \textbf{0.645\,\scriptsize$\pm$0.006} & $+7.07$ & 0.604\,\scriptsize$\pm$0.003 & \textbf{0.625\,\scriptsize$\pm$0.006} & $+3.55$ \\
Genitourinary System        & 0.605\,\scriptsize$\pm$0.004 & \textbf{0.641\,\scriptsize$\pm$0.004} & $+5.92$ & 0.604\,\scriptsize$\pm$0.004 & \textbf{0.623\,\scriptsize$\pm$0.004} & $+3.22$ \\
\midrule
\textbf{Mean}               & 0.617 & \textbf{0.654} & $+6.00$ & 0.616 & \textbf{0.638} & $+3.62$ \\
\bottomrule
\end{tabular}%
}
\end{table}

\begin{table}[t]
\centering
\caption{{Group-disease comparison under the Drop-3 Organs setting (full backbone fine-tuning).} Results are aggregated over 25 runs. $\Delta$ denotes the relative improvement of Pan-FM over the DINOv2 baseline.}
\label{tab:drop3_per_disease_ft}
\resizebox{\textwidth}{!}{%
\begin{tabular}{l cc c cc c}
\toprule
\multirow{2}{*}{Disease} & \multicolumn{3}{c}{AUROC} & \multicolumn{3}{c}{Balanced Accuracy} \\
\cmidrule(lr){2-4} \cmidrule(lr){5-7}
 & DINOv2 & Pan-FM (Ours) & $\Delta$ (\%) & DINOv2 & Pan-FM (Ours) & $\Delta$ (\%) \\
\midrule
Infectious \& Parasitic  & 0.642\,\scriptsize$\pm$0.034 & \textbf{0.643\,\scriptsize$\pm$0.046} & $+0.23$ & 0.632\,\scriptsize$\pm$0.025 & \textbf{0.632\,\scriptsize$\pm$0.033} & $+0.03$ \\
Neoplasms                & 0.619\,\scriptsize$\pm$0.019 & \textbf{0.666\,\scriptsize$\pm$0.014} & $+7.61$ & 0.605\,\scriptsize$\pm$0.009 & \textbf{0.644\,\scriptsize$\pm$0.010} & $+6.35$ \\
Blood \& Immune          & 0.616\,\scriptsize$\pm$0.035 & \textbf{0.673\,\scriptsize$\pm$0.028} & $+9.33$ & 0.610\,\scriptsize$\pm$0.024 & \textbf{0.656\,\scriptsize$\pm$0.027} & $+7.52$ \\
Endocrine \& Metabolic   & 0.695\,\scriptsize$\pm$0.019 & \textbf{0.744\,\scriptsize$\pm$0.011} & $+7.11$ & 0.663\,\scriptsize$\pm$0.016 & \textbf{0.693\,\scriptsize$\pm$0.017} & $+4.51$ \\
Mental \& Behavioural    & 0.654\,\scriptsize$\pm$0.030 & \textbf{0.695\,\scriptsize$\pm$0.016} & $+6.26$ & 0.639\,\scriptsize$\pm$0.011 & \textbf{0.666\,\scriptsize$\pm$0.016} & $+4.18$ \\
Nervous System           & 0.629\,\scriptsize$\pm$0.019 & \textbf{0.673\,\scriptsize$\pm$0.016} & $+6.86$ & 0.618\,\scriptsize$\pm$0.017 & \textbf{0.654\,\scriptsize$\pm$0.010} & $+5.86$ \\
Eye                      & \textbf{0.636\,\scriptsize$\pm$0.014} & 0.622\,\scriptsize$\pm$0.019 & $-2.22$ & 0.630\,\scriptsize$\pm$0.024 & \textbf{0.635\,\scriptsize$\pm$0.019} & $+0.75$ \\
Circulatory System       & 0.675\,\scriptsize$\pm$0.019 & \textbf{0.706\,\scriptsize$\pm$0.005} & $+4.59$ & 0.652\,\scriptsize$\pm$0.021 & \textbf{0.659\,\scriptsize$\pm$0.005} & $+1.07$ \\
Respiratory System       & \textbf{0.674\,\scriptsize$\pm$0.023} & 0.668\,\scriptsize$\pm$0.013 & $-0.79$ & \textbf{0.654\,\scriptsize$\pm$0.028} & 0.637\,\scriptsize$\pm$0.009 & $-2.54$ \\
Digestive System         & 0.628\,\scriptsize$\pm$0.021 & \textbf{0.666\,\scriptsize$\pm$0.020} & $+6.03$ & 0.613\,\scriptsize$\pm$0.018 & \textbf{0.633\,\scriptsize$\pm$0.017} & $+3.36$ \\
Skin                     & 0.641\,\scriptsize$\pm$0.027 & \textbf{0.660\,\scriptsize$\pm$0.015} & $+3.00$ & 0.635\,\scriptsize$\pm$0.017 & \textbf{0.637\,\scriptsize$\pm$0.014} & $+0.32$ \\
Musculoskeletal System   & 0.642\,\scriptsize$\pm$0.021 & \textbf{0.687\,\scriptsize$\pm$0.013} & $+6.93$ & 0.636\,\scriptsize$\pm$0.021 & \textbf{0.659\,\scriptsize$\pm$0.015} & $+3.58$ \\
Genitourinary System     & 0.637\,\scriptsize$\pm$0.016 & \textbf{0.671\,\scriptsize$\pm$0.014} & $+5.37$ & 0.625\,\scriptsize$\pm$0.008 & \textbf{0.647\,\scriptsize$\pm$0.017} & $+3.50$ \\
\midrule
\textbf{Mean}            & 0.645 & \textbf{0.675} & $+4.61$ & 0.632 & \textbf{0.650} & $+2.92$ \\
\bottomrule
\end{tabular}%
}
\end{table}

\subsection{Group-Disease Comparison with More Missing Organs}
\label{sec:per_disease_drop3}

To evaluate performance under a more challenging missing-modality regime, we randomly drop three organs at test time and report per-disease AUROC and BalAcc in Tables~\ref{tab:drop3_per_disease}  and~\ref{tab:drop3_per_disease_ft}. Under linear probing, Pan-FM outperforms DINOv2 on 12 of 13 diseases in AUROC, with a mean gain of $+6.00\%$ AUROC and $+3.62\%$ balanced accuracy; the largest improvements appear on Nervous System ($+10.25\%$), Endocrine \& Metabolic ($+9.99\%$), and Blood \& Immune ($+9.73\%$) diseases, which typically benefit from evidence integrated across multiple organs. The same trend holds under full backbone fine-tuning (mean gains of $+4.61\%$ AUROC and $+2.92\%$ balanced accuracy), showing that the  whole-body representations learned by Pan-FM is also beneficial for backbone fine-tuning based downstream adaptation.

\begin{table}[t]
\centering
\caption{Comparison of AUROC on the held-out test set between more \textbf{single-organ baselines} trained with organ-specific data and our proposed pan-organ model across 13 group diseases, evaluated on subjects with complete 7-organ imaging data.}
\label{tab:group_ft_single_organ_auroc}
\resizebox{\textwidth}{!}{%
\begin{tabular}{llccccccccccccc|c}
\toprule
\multirow{1}{*}{\textbf{Organ}} & \textbf{Method}
  & \shortstack{Infect.\\Paras.}
  & \shortstack{Neo-\\plasms}
  & \shortstack{Blood\\Immune}
  & \shortstack{Endoc.\\Metabol.}
  & \shortstack{Mental\\Behav.}
  & \shortstack{Nervous\\System}
  & \shortstack{Eye\\Disease}
  & \shortstack{Circulat.\\System}
  & \shortstack{Resp.\\System}
  & \shortstack{Digest.\\System}
  & \shortstack{Skin\\System}
  & \shortstack{Musculo-\\skeletal}
  & \shortstack{Genito-\\urinary}
  & \textbf{Mean} \\
\midrule

\multirow{5}{*}{Brain}  & Lasso  & 0.534  & 0.560  & 0.594  & 0.650  & 0.663  & 0.565  & 0.647  & 0.647  & 0.599  & 0.581  & 0.566  & 0.567  & 0.561  & 0.595 \\
  & ElasticNet  & 0.533  & 0.572  & 0.602  & 0.650  & 0.666  & 0.565  & 0.626  & 0.647  & 0.603  & 0.586  & 0.568  & 0.567  & 0.559  & 0.596 \\
  & Linear SVM  & 0.545  & 0.565  & 0.572  & 0.662  & 0.646  & 0.574  & 0.584  & 0.576  & 0.607  & 0.568  & 0.576  & 0.526  & 0.583  & 0.583 \\
  & RF  & 0.546  & 0.592  & 0.571  & 0.655  & \underline{0.682}  & 0.543  & 0.571  & 0.669  & 0.623  & 0.624  & 0.603  & 0.620  & 0.604  & 0.608 \\
  & MLP  & 0.515  & 0.538  & 0.638  & 0.657  & 0.644  & 0.563  & 0.629  & 0.655  & 0.586  & 0.605  & 0.561  & 0.596  & 0.564  & 0.596 \\
\midrule

\multirow{5}{*}{Heart}  & Lasso  & \underline{0.688}  & 0.630  & 0.633  & 0.707  & 0.656  & 0.655  & 0.668  & \underline{0.684}  & 0.660  & 0.648  & 0.664  & 0.666  & 0.645  & 0.662 \\
  & ElasticNet  & \underline{0.688}  & 0.628  & 0.643  & 0.706  & 0.663  & 0.655  & 0.651  & \underline{0.684}  & 0.652  & 0.648  & 0.664  & 0.666  & 0.642  & 0.661 \\
  & Linear SVM  & 0.580  & 0.568  & 0.601  & 0.685  & 0.616  & 0.663  & 0.494  & 0.617  & 0.678  & 0.528  & 0.666  & 0.544  & 0.579  & 0.601 \\
  & RF  & 0.682  & 0.601  & 0.602  & 0.659  & 0.662  & 0.627  & 0.550  & 0.654  & 0.629  & 0.623  & 0.605  & 0.634  & 0.626  & 0.627 \\
  & MLP  & 0.624  & 0.609  & 0.626  & 0.703  & 0.644  & 0.661  & 0.596  & 0.670  & 0.665  & 0.602  & 0.611  & 0.642  & 0.629  & 0.637 \\
\midrule

\multirow{5}{*}{Adipose}  & Lasso  & 0.640  & 0.665  & 0.592  & 0.719  & 0.644  & 0.645  & \underline{0.716}  & 0.679  & 0.665  & 0.663  & 0.660  & 0.678  & 0.667  & 0.664 \\
  & ElasticNet  & 0.653  & 0.665  & 0.592  & \underline{0.723}  & 0.647  & 0.647  & \textbf{0.717}  & 0.680  & 0.667  & 0.663  & 0.659  & 0.683  & 0.679  & \underline{0.667} \\
  & Linear SVM  & 0.587  & 0.626  & 0.517  & 0.722  & 0.640  & 0.632  & 0.430  & 0.644  & 0.637  & 0.645  & 0.650  & 0.639  & 0.580  & 0.611 \\
  & RF  & 0.663  & 0.640  & 0.611  & 0.701  & 0.645  & 0.664  & 0.638  & 0.656  & \underline{0.683}  & 0.657  & \underline{0.672}  & \underline{0.706}  & 0.670  & 0.662 \\
  & MLP  & 0.648  & \underline{0.673}  & 0.581  & 0.707  & 0.647  & \underline{0.700}  & 0.501  & 0.657  & \underline{0.683}  & \underline{0.673}  & \textbf{0.692}  & \underline{0.706}  & \underline{0.688}  & 0.658 \\
\midrule

\multirow{5}{*}{Liver}  & Lasso  & 0.500  & 0.555  & 0.572  & 0.573  & 0.550  & 0.600  & 0.479  & 0.576  & 0.568  & 0.562  & 0.577  & 0.586  & 0.589  & 0.561 \\
  & ElasticNet  & 0.500  & 0.558  & 0.572  & 0.573  & 0.548  & 0.600  & 0.446  & 0.576  & 0.567  & 0.558  & 0.582  & 0.591  & 0.590  & 0.559 \\
  & Linear SVM  & 0.471  & 0.510  & 0.485  & 0.566  & 0.522  & 0.521  & 0.461  & 0.513  & 0.561  & 0.508  & 0.511  & 0.529  & 0.528  & 0.514 \\
  & RF  & 0.665  & 0.587  & 0.593  & 0.663  & 0.639  & 0.594  & 0.593  & 0.592  & 0.627  & 0.596  & 0.605  & 0.598  & 0.605  & 0.612 \\
  & MLP  & 0.359  & 0.612  & 0.623  & 0.688  & 0.647  & 0.634  & 0.455  & 0.658  & 0.653  & 0.602  & 0.609  & 0.611  & 0.610  & 0.597 \\
\midrule

\multirow{5}{*}{Kidney}  & Lasso  & 0.476  & 0.505  & 0.493  & 0.544  & 0.500  & 0.433  & 0.502  & 0.531  & 0.540  & 0.497  & 0.521  & 0.500  & 0.530  & 0.506 \\
  & ElasticNet  & 0.476  & 0.505  & 0.493  & 0.546  & 0.500  & 0.445  & 0.452  & 0.532  & 0.534  & 0.503  & 0.519  & 0.500  & 0.530  & 0.503 \\
  & Linear SVM  & 0.491  & 0.529  & 0.513  & 0.587  & 0.508  & 0.538  & 0.435  & 0.539  & 0.561  & 0.521  & 0.524  & 0.511  & 0.541  & 0.523 \\
  & RF  & 0.469  & 0.542  & 0.490  & 0.602  & 0.538  & 0.580  & 0.481  & 0.576  & 0.568  & 0.528  & 0.503  & 0.521  & 0.491  & 0.530 \\
  & MLP  & 0.452  & 0.587  & 0.514  & 0.587  & 0.520  & 0.582  & 0.407  & 0.593  & 0.599  & 0.573  & 0.588  & 0.568  & 0.534  & 0.546 \\
\midrule

\multirow{5}{*}{Spleen}  & Lasso  & 0.576  & 0.521  & 0.526  & 0.525  & 0.533  & 0.531  & 0.475  & 0.515  & 0.499  & 0.500  & 0.528  & 0.487  & 0.490  & 0.516 \\
  & ElasticNet  & 0.576  & 0.521  & 0.526  & 0.525  & 0.533  & 0.531  & 0.443  & 0.511  & 0.499  & 0.500  & 0.527  & 0.487  & 0.490  & 0.513 \\
  & Linear SVM  & 0.581  & 0.482  & 0.525  & 0.515  & 0.475  & 0.456  & 0.517  & 0.502  & 0.483  & 0.513  & 0.455  & 0.466  & 0.483  & 0.496 \\
  & RF  & 0.482  & 0.548  & 0.548  & 0.585  & 0.540  & 0.542  & 0.443  & 0.552  & 0.520  & 0.534  & 0.569  & 0.507  & 0.561  & 0.533 \\
  & MLP  & 0.544  & 0.526  & 0.550  & 0.576  & 0.533  & 0.520  & 0.399  & 0.527  & 0.530  & 0.506  & 0.555  & 0.501  & 0.513  & 0.522 \\
\midrule

\multirow{5}{*}{Pancreas}  & Lasso  & 0.578  & 0.593  & 0.619  & 0.618  & 0.580  & 0.607  & 0.637  & 0.618  & 0.618  & 0.628  & 0.597  & 0.585  & 0.608  & 0.607 \\
  & ElasticNet  & 0.578  & 0.593  & 0.619  & 0.617  & 0.579  & 0.617  & 0.636  & 0.619  & 0.621  & 0.629  & 0.598  & 0.581  & 0.611  & 0.608 \\
  & Linear SVM  & 0.593  & 0.599  & \underline{0.654}  & 0.618  & 0.533  & 0.630  & 0.643  & 0.613  & 0.608  & 0.633  & 0.606  & 0.587  & 0.622  & 0.611 \\
  & RF  & 0.582  & 0.562  & 0.610  & 0.626  & 0.592  & 0.626  & 0.541  & 0.597  & 0.609  & 0.607  & 0.578  & 0.546  & 0.589  & 0.590 \\
  & MLP  & 0.550  & 0.584  & 0.625  & 0.616  & 0.579  & 0.640  & 0.603  & 0.613  & 0.612  & 0.591  & 0.602  & 0.594  & 0.621  & 0.602 \\
\midrule

\multicolumn{2}{l}{\textbf{Pan-FM (Ours)}}
  & \textbf{0.691}
  & \textbf{0.702}
  & \textbf{0.681}
  & \textbf{0.772}
  & \textbf{0.735}
  & \textbf{0.737}
  & 0.683
  & \textbf{0.722}
  & \textbf{0.685}
  & \textbf{0.678}
  & 0.671
  & \textbf{0.708}
  & \textbf{0.692}
  & \textbf{0.704} \\
\bottomrule
\end{tabular}%
}
\end{table}

\begin{table}[t]
\centering
\caption{Comparison of balanced accuracy on the held-out test set between more \textbf{single-organ baselines} trained with organ-specific data and our proposed pan-organ model across 13 group diseases, evaluated on subjects with complete 7-organ imaging data.}
\label{tab:group_ft_single_organ_balacc}
\resizebox{\textwidth}{!}{%
\begin{tabular}{llccccccccccccc|c}
\toprule
\multirow{1}{*}{\textbf{Organ}} & \textbf{Method}
& \shortstack{Infect.\\Paras.}
& \shortstack{Neo-\\plasms}
& \shortstack{Blood\\Immune}
& \shortstack{Endoc.\\Metabol.}
& \shortstack{Mental\\Behav.}
& \shortstack{Nervous\\System}
& \shortstack{Eye\\Disease}
& \shortstack{Circulat.\\System}
& \shortstack{Resp.\\System}
& \shortstack{Digest.\\System}
& \shortstack{Skin\\System}
& \shortstack{Musculo-\\skeletal}
& \shortstack{Genito-\\urinary}
& \textbf{Mean} \\
\midrule

\multirow{5}{*}{Brain}
& Lasso
& 0.557 & 0.564 & 0.603 & 0.620 & 0.657 & 0.572 & 0.625 & 0.629 & 0.606 & 0.577 & 0.584 & 0.588 & 0.561 & 0.596 \\
& ElasticNet
& 0.557 & 0.573 & 0.619 & 0.617 & 0.656 & 0.574 & 0.607 & 0.626 & 0.609 & 0.574 & 0.579 & 0.587 & 0.563 & 0.595 \\
& Linear SVM
& 0.553 & 0.571 & 0.579 & 0.628 & 0.628 & 0.596 & 0.610 & 0.584 & 0.612 & 0.584 & 0.585 & 0.556 & 0.592 & 0.591 \\
& RF
& 0.566 & 0.599 & 0.579 & 0.644 & \underline{0.660} & 0.555 & 0.586 & 0.640 & 0.615 & 0.614 & 0.627 & 0.626 & 0.629 & 0.611 \\
& MLP
& 0.556 & 0.553 & 0.629 & 0.623 & 0.634 & 0.582 & 0.633 & 0.628 & 0.601 & 0.590 & 0.574 & 0.600 & 0.570 & 0.598 \\
\midrule

\multirow{5}{*}{Heart}
& Lasso
& \textbf{0.666} & 0.623 & 0.640 & 0.669 & 0.644 & 0.646 & 0.682 & \underline{0.666} & 0.621 & 0.629 & 0.647 & 0.641 & 0.621 & 0.646 \\
& ElasticNet
& \textbf{0.666} & 0.623 & 0.619 & 0.665 & 0.650 & 0.645 & 0.670 & \underline{0.666} & 0.632 & 0.629 & 0.647 & 0.641 & 0.619 & 0.644 \\
& Linear SVM
& 0.603 & 0.559 & 0.607 & 0.649 & 0.603 & 0.660 & 0.552 & 0.594 & 0.636 & 0.548 & 0.652 & 0.572 & 0.593 & 0.602 \\
& RF
& \underline{0.664} & 0.590 & 0.595 & 0.654 & 0.646 & 0.623 & 0.573 & 0.643 & 0.644 & 0.614 & 0.595 & 0.628 & 0.602 & 0.620 \\
& MLP
& 0.633 & 0.616 & \textbf{0.646} & 0.662 & 0.654 & 0.638 & 0.611 & 0.656 & \textbf{0.656} & 0.596 & 0.612 & 0.611 & 0.605 & 0.630 \\
\midrule

\multirow{5}{*}{Adipose}
& Lasso
& 0.642 & 0.644 & 0.589 & \underline{0.679} & 0.646 & 0.636 & \textbf{0.722} & 0.646 & 0.636 & \underline{0.637} & 0.648 & 0.668 & 0.639 & 0.649 \\
& ElasticNet
& 0.658 & 0.644 & 0.587 & 0.673 & 0.657 & 0.636 & \underline{0.712} & 0.647 & 0.645 & \underline{0.637} & \underline{0.661} & \underline{0.676} & 0.653 & \underline{0.653} \\
& Linear SVM
& 0.591 & \underline{0.648} & 0.561 & 0.665 & 0.648 & 0.620 & 0.529 & 0.621 & 0.632 & 0.630 & 0.620 & 0.605 & 0.581 & 0.609 \\
& RF
& 0.659 & 0.612 & 0.604 & 0.662 & 0.619 & \underline{0.675} & 0.639 & 0.630 & 0.630 & 0.630 & 0.655 & 0.663 & \underline{0.672} & 0.642 \\
& MLP
& 0.639 & \underline{0.648} & 0.588 & 0.672 & 0.642 & 0.670 & 0.594 & 0.655 & 0.648 & 0.631 & \textbf{0.673} & 0.674 & 0.662 & 0.646 \\
\midrule

\multirow{5}{*}{Liver}
& Lasso
& 0.500 & 0.578 & 0.593 & 0.590 & 0.584 & 0.622 & 0.545 & 0.597 & 0.589 & 0.569 & 0.586 & 0.590 & 0.583 & 0.579 \\
& ElasticNet
& 0.500 & 0.569 & 0.593 & 0.590 & 0.581 & 0.622 & 0.513 & 0.597 & 0.581 & 0.566 & 0.582 & 0.599 & 0.580 & 0.575 \\
& Linear SVM
& 0.524 & 0.534 & 0.518 & 0.599 & 0.588 & 0.540 & 0.559 & 0.552 & 0.583 & 0.533 & 0.557 & 0.544 & 0.545 & 0.552 \\
& RF
& \underline{0.664} & 0.580 & 0.603 & 0.657 & 0.630 & 0.619 & 0.578 & 0.579 & 0.635 & 0.595 & 0.586 & 0.602 & 0.590 & 0.609 \\
& MLP
& 0.509 & 0.609 & 0.632 & 0.654 & 0.637 & 0.645 & 0.515 & 0.629 & 0.632 & 0.592 & 0.603 & 0.607 & 0.598 & 0.605 \\
\midrule

\multirow{5}{*}{Kidney}
& Lasso
& 0.524 & 0.536 & 0.562 & 0.564 & 0.500 & 0.504 & 0.550 & 0.549 & 0.555 & 0.517 & 0.552 & 0.500 & 0.544 & 0.535 \\
& ElasticNet
& 0.524 & 0.536 & 0.552 & 0.561 & 0.500 & 0.506 & 0.547 & 0.549 & 0.556 & 0.525 & 0.548 & 0.500 & 0.544 & 0.534 \\
& Linear SVM
& 0.556 & 0.552 & 0.552 & 0.596 & 0.538 & 0.597 & 0.556 & 0.565 & 0.563 & 0.558 & 0.563 & 0.534 & 0.550 & 0.560 \\
& RF
& 0.538 & 0.557 & 0.543 & 0.599 & 0.555 & 0.573 & 0.528 & 0.567 & 0.586 & 0.535 & 0.538 & 0.541 & 0.539 & 0.554 \\
& MLP
& 0.527 & 0.592 & 0.559 & 0.588 & 0.552 & 0.585 & 0.547 & 0.595 & 0.598 & 0.576 & 0.589 & 0.572 & 0.537 & 0.571 \\
\midrule

\multirow{5}{*}{Spleen}
& Lasso
& 0.595 & 0.549 & 0.568 & 0.574 & 0.564 & 0.553 & 0.542 & 0.551 & 0.544 & 0.500 & 0.556 & 0.535 & 0.526 & 0.551 \\
& ElasticNet
& 0.595 & 0.549 & 0.568 & 0.573 & 0.565 & 0.553 & 0.515 & 0.543 & 0.546 & 0.500 & 0.560 & 0.535 & 0.526 & 0.548 \\
& Linear SVM
& 0.607 & 0.514 & 0.561 & 0.550 & 0.535 & 0.519 & 0.591 & 0.519 & 0.529 & 0.524 & 0.515 & 0.516 & 0.520 & 0.538 \\
& RF
& 0.554 & 0.557 & 0.574 & 0.600 & 0.567 & 0.560 & 0.544 & 0.561 & 0.544 & 0.548 & 0.594 & 0.529 & 0.601 & 0.564 \\
& MLP
& 0.556 & 0.553 & 0.560 & 0.594 & 0.567 & 0.556 & 0.520 & 0.547 & 0.534 & 0.541 & 0.565 & 0.527 & 0.528 & 0.550 \\
\midrule

\multirow{5}{*}{Pancreas}
& Lasso
& 0.580 & 0.586 & \underline{0.645} & 0.596 & 0.604 & 0.613 & 0.639 & 0.598 & 0.605 & 0.614 & 0.594 & 0.579 & 0.608 & 0.605 \\
& ElasticNet
& 0.583 & 0.586 & \underline{0.645} & 0.596 & 0.599 & 0.616 & 0.631 & 0.604 & 0.599 & 0.615 & 0.594 & 0.579 & 0.605 & 0.604 \\
& Linear SVM
& 0.600 & 0.595 & 0.632 & 0.596 & 0.548 & 0.633 & 0.650 & 0.602 & 0.596 & 0.603 & 0.606 & 0.582 & 0.596 & 0.603 \\
& RF
& 0.592 & 0.581 & 0.625 & 0.599 & 0.599 & 0.637 & 0.567 & 0.576 & 0.604 & 0.601 & 0.572 & 0.564 & 0.593 & 0.593 \\
& MLP
& 0.562 & 0.578 & 0.623 & 0.595 & 0.613 & 0.635 & 0.617 & 0.596 & 0.601 & 0.588 & 0.585 & 0.589 & 0.618 & 0.600 \\
\midrule

\multicolumn{2}{l}{\textbf{Pan-FM (Ours)}}
& 0.640
& \textbf{0.687}
& \underline{0.645}
& \textbf{0.711}
& \textbf{0.718}
& \textbf{0.711}
& 0.688
& \textbf{0.685}
& \underline{0.652}
& \textbf{0.648}
& 0.628
& \textbf{0.680}
& \textbf{0.675}
& \textbf{0.674} \\
\bottomrule
\end{tabular}%
}
\end{table}

\begin{figure}[t]
\begin{center}
   \includegraphics[width=1.0\linewidth]{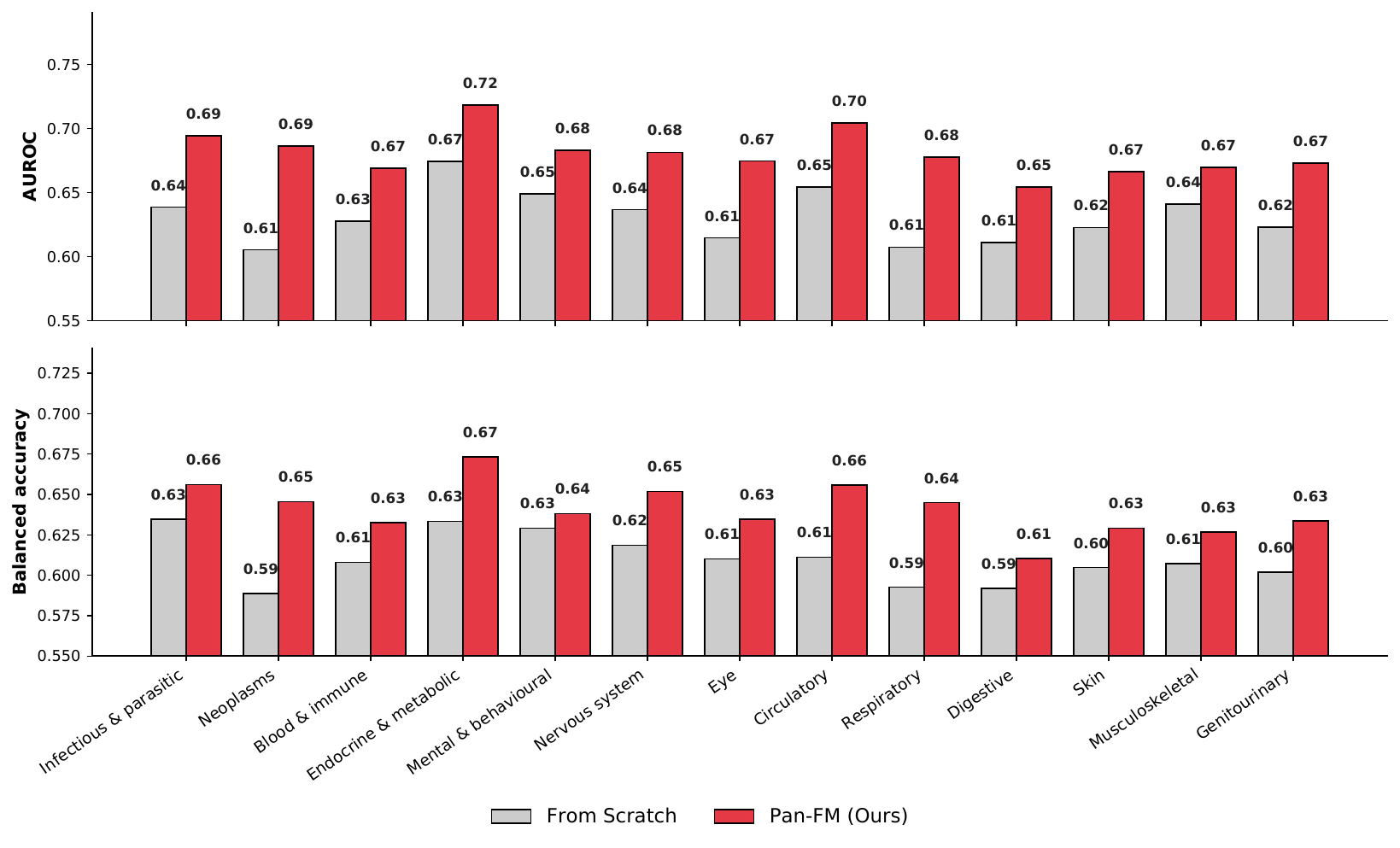} 
\end{center}
 \caption{\textbf{Per-disease comparison: Pan-FM vs. from-scratch.} Both models use the same full ViT multi-organ backbone and are fine-tuned end-to-end with identical schedules. The from-scratch baseline initialises the backbone with the standard ViT scheme (truncated-normal) while Pan-FM initialises it with our pretrained weights.}
 \label{fig: fig_comparison_random}
\end{figure}

\subsection{Comparison with Single-Organ Baselines}
\label{sec:more_single_organ_baselines}

We further compare Pan-FM against more classical single-organ baselines, where each organ is trained independently with five predictors (i.e., Lasso, ElasticNet, Linear SVM, Random Forest, and MLP) in a 10-fold cross-validation way. As shown in Tables~\ref{tab:group_ft_single_organ_auroc} and~\ref{tab:group_ft_single_organ_balacc}, Pan-FM achieves the best mean AUROC ($0.704$) and balanced accuracy ($0.674$), outperforming all the single-organ baselines. Gains are largest on diseases with multi-system aetiology (Endocrine \& Metabolic, Mental \& Behavioural, Nervous System, Circulatory System), where no single organ can provide sufficient evidence. 
These results demonstrate that Pan-FM's advantage arises not from any single organ modality, but from jointly reasoning across all the organs. This yields effective \emph{whole-body representations} and marks a step toward more generalizable whole-body foundation models.

\subsection{Comparison with Random (From-Scratch) Baseline}
\label{sec:finetune_necessity}

To verify the effectiveness of the proposed Pan-FM pre-training, we compare two identical full ViT multi-organ backbones, fine-tuned end-to-end on the 13-way disease tasks. The main difference is the backbone initialisation: weights from our Pan-FM pre-training, or the standard ViT scheme (truncated-normal). As shown in Fig.~\ref{fig: fig_comparison_random}, Pan-FM outperforms the from-scratch baseline on all 13 disease categories, with a mean gain of $+5.0$ AUROC points. 
This gap reflects a well-known property of high-capacity Transformers, i.e., end-to-end training from scratch requires data at a scale that downstream clinical cohorts rarely provide. Large-scale multi-organ self-supervision is therefore not merely a favourable initialisation but a practical prerequisite for adapting full ViT backbones in the low-label regime typical of medical applications.

\section{More Ablation Studies}
\label{sec:more_ablation_studies}
In this section, we conduct more ablation studies to demonstrate the effectiveness of our proposed Pan-FM.

\begin{table}[t]
    \centering
    \caption{\textbf{Computational overhead of saliency-guided masking (SGM).} Measured over 1{,}000 iterations after 50 warmup iterations on
    a single V100 GPU with FP16 mixed precision. Both runs use identical model weights and batch sequences. The only difference
    is the masking strategy.}
    \label{tab:sgm_overhead}
    \small
    \begin{tabular}{lccc}
        \toprule
        Metric & DINOv2 & Pan-FM & $\Delta$ \\
        \midrule
        Time / iter (ms)            & $79.3 \pm 3.3$ & $82.1 \pm 0.6$ & $+3.5\%$ \\
        FLOPs (G)   & $96.4$         & $96.4$         & $\phantom{+}0.0\%$ \\
        \bottomrule
    \end{tabular}
\end{table}

\subsection{Computational Overhead}
\label{sec:overhead_analysis}
We compare the per-iteration training cost of our Pan-FM with SGM
against the DINOv2 baseline in Table~\ref{tab:sgm_overhead}. Pan-FM
introduces a marginal $+3.5\%$ wall-clock overhead per iteration
($82.1$ vs $79.3$ ms). Notably, the backbone forward FLOPs are
identical between the two models ($96.4$ G), as Pan-FM does not modify
the DINOv2 architecture. The entire overhead arises from teacher-side
attention extraction and multinomial sampling. Extrapolated to a
$200$-epoch pretraining run, this leads to only $\sim\!11$
additional minutes of total wall-clock time, which is an essentially negligible cost. Crucially, this overhead
is moreover confined to the self-supervised pre-training phase. At
inference and during downstream fine-tuning, our model is
architecturally identical to the baseline and incurs zero additional
cost.

\begin{figure}[t]
\begin{center}
   \includegraphics[width=1.0\linewidth]{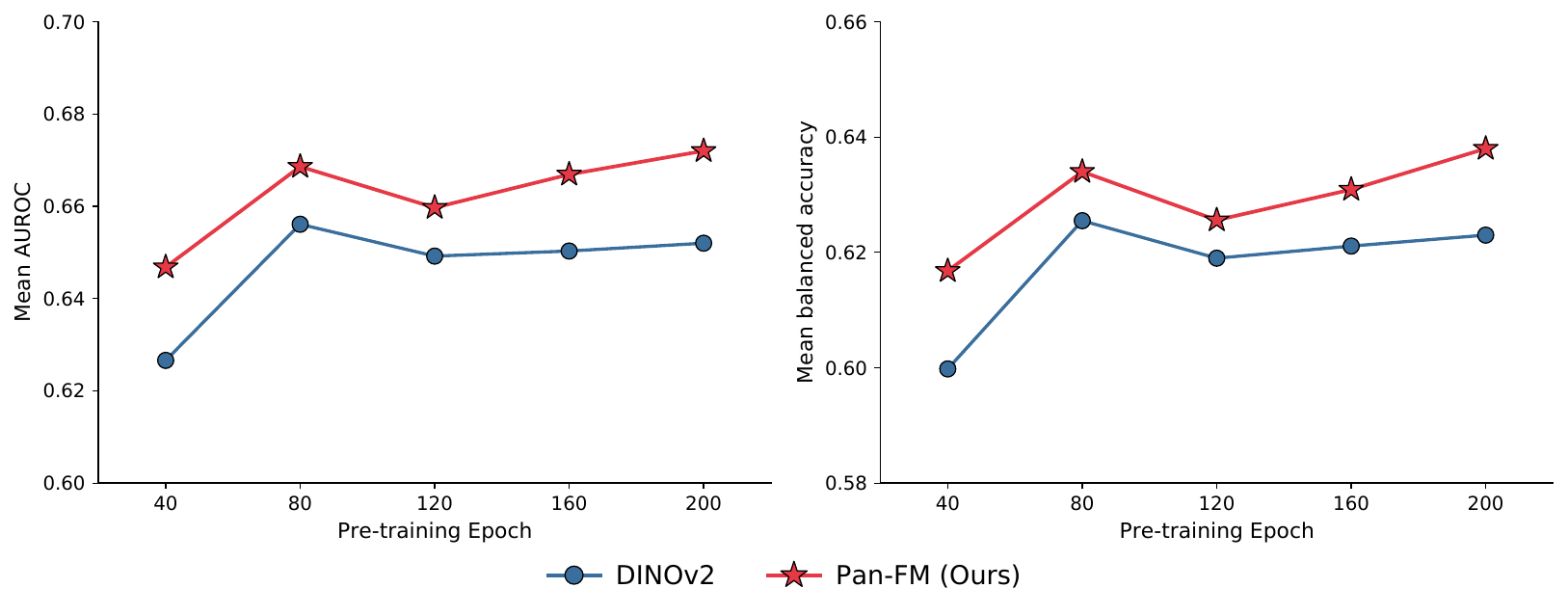} 
\end{center}
 \caption{\textbf{Linear-probing convergence across pre-training epochs.} For each checkpoint, we report the linear probing performance on the full held-out test set as the mean AUROC (left) and mean balanced accuracy (right) across the 13 categories. Pan-FM with SGM consistently outperforms the DINOv2 baseline throughout the whole pre-training stage.}
 \label{fig:pretrain_epoch}
\end{figure}

\subsection{Pre-training Convergence}
\label{sec:pretrain_covergence}

To assess how representation quality evolves during self-supervised
pre-training, we perform linear probing at every $40$ pre-training
epochs. At each checkpoint, the backbone is frozen and a linear
classifier is trained on top of the $\mathrm{[CLS]}$ token using the
training set, then evaluated on the full held-out test set. We report the mean AUROC and mean balanced accuracy
across the $13$ disease categories in Fig.~\ref{fig:pretrain_epoch}. At epoch $40$, Pan-FM already outperforms
the baseline by a noticeable margin ($0.647$ vs $0.627$ AUROC),
indicating that the SGM signal accelerates representation acquisition
from early in pre-training. Both methods improve until epoch $80$,
after which their behaviours diverge: the DINOv2 baseline plateaus at
$\sim\!0.652$ AUROC and shows essentially no further gain across the
remaining $120$ epochs, whereas Pan-FM continues to improve and
reaches $0.672$ AUROC and $0.638$ balanced accuracy at epoch $200$
($+2.0$ and $+1.5$ points over the baseline, respectively). The
sustained improvement of Pan-FM through the full $200$-epoch budget
indicates that saliency-guided masking continues to provide useful
learning gradient after random masking has saturated, by repeatedly
challenging the representation to recover from the loss of the most
informative organs.

\begin{figure}[t]
\begin{center}
   \includegraphics[width=1.0\linewidth]{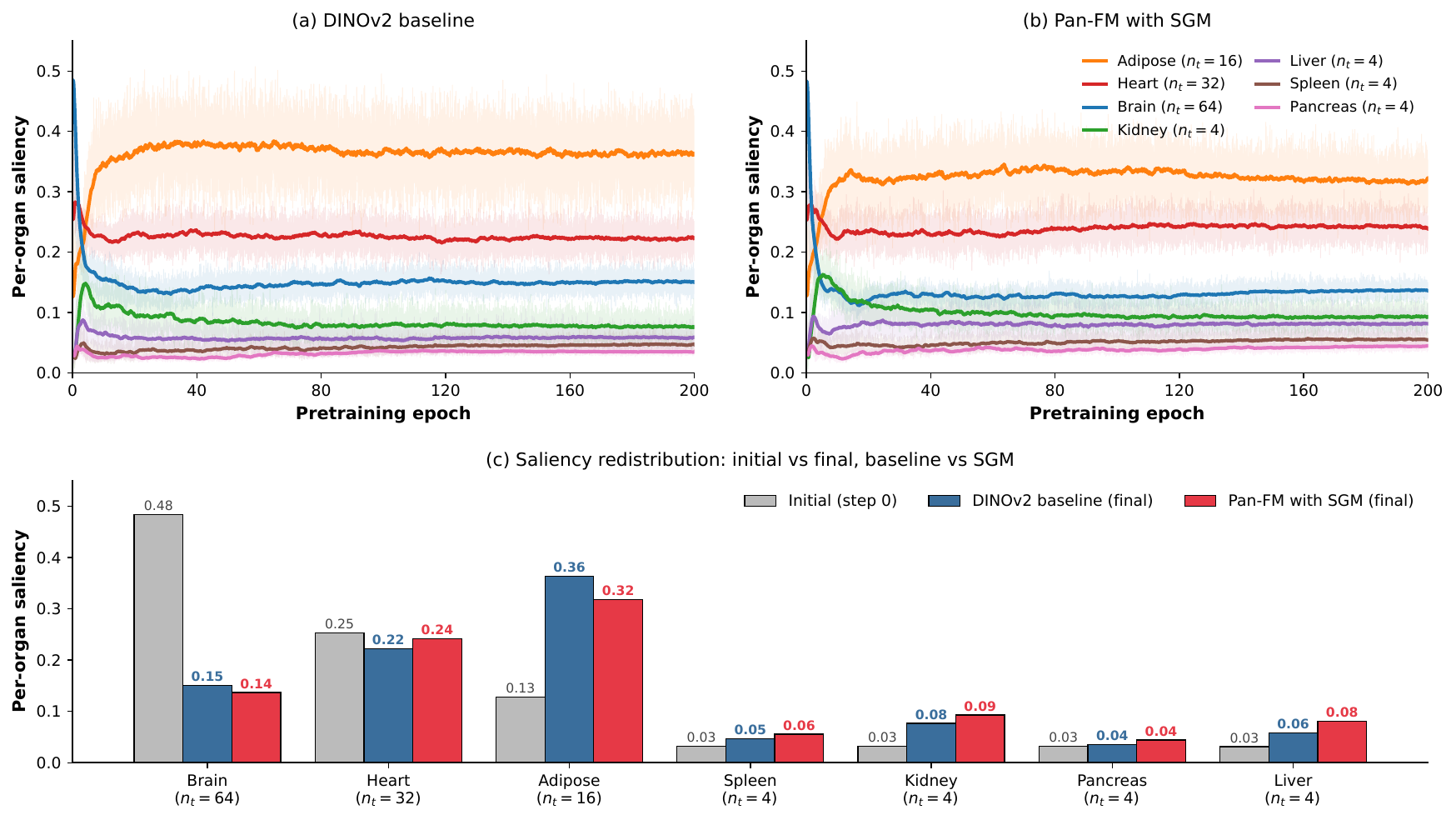} 
\end{center}
 \caption{\textbf{The dominant-organ shortcut bias and its mitigation
    by SGM.} Per-organ saliency is defined as the $\mathrm{[CLS]}$-token
    attention mass aggregated over each organ's tokens in the teacher
    backbone, computed on the full multi-organ input without any
    masking. The two runs share identical initialisation, optimisation
    schedule, and training data, and differ only in whether SGM is
    applied to the student. \textbf{(a, b)} Saliency trajectories
    across $200$ pretraining epochs for the DINOv2 baseline and Pan-FM,
    respectively; faint lines show raw values and bold lines show
    exponential-moving-average-smoothed trajectories. \textbf{(c)}
    Per-organ saliency at initialisation (step $0$, grey) compared to
    the final state of the baseline (blue) and Pan-FM (red). At
    initialisation, saliency is approximately proportional to each
    organ's token-count footprint ($n_t$). Self-supervised pretraining
    alone displaces brain but \emph{relocates} the shortcut to a new
    dominant organ (adipose, $36\%$), while SGM mitigates this 
bias by redistributing representation capacity across organs, rather
than merely moving the shortcut from one dominant organ to another.}
 \label{fig:saliency_compare}
\end{figure}

\subsection{Saliency Redistribution Under SGM}
\label{sec:organ_redistribution}

The dominant-organ shortcut learning bias (see Sec.~\ref{sec:naive_pretrain}) suggests that, without explicit
intervention, a multi-organ self-supervised model tends to allocate most of its representation capacity to only a few highly attended organs, while underusing the remaining organs. We directly examine this effect by tracking per-organ saliency throughout training. Specifically, we define organ saliency as the total $\mathrm{[CLS]}$-token attention mass
assigned to all tokens from a given organ in the teacher backbone,
computed using the full multi-organ input. We compare Pan-FM with a
DINOv2 baseline that uses the same initialization and training schedule
but applies uniform random masking.

At step $0$ (Fig.~\ref{fig:saliency_compare}\,c, grey bars), the two
models show almost identical saliency patterns. Attention is distributed
mainly according to the number of tokens per organ: the brain, with
$64$ tokens, receives $48\%$ of the total $\mathrm{[CLS]}$ attention,
whereas each $4$-token organ receives only about $3\%$. This indicates
that the model begins training with a strong token-count prior, which
provides a natural starting point for shortcut learning. 
After self-supervised pretraining, the DINOv2 baseline reduces its
reliance on the brain ($48\%\!\to\!15\%$), but the attention budget is
largely transferred to another dominant organ, adipose
($13\%\!\to\!36\%$). Thus, the shortcut bias is not eliminated; it is
simply shifted from one organ to another. In contrast, SGM uses the
teacher's saliency distribution to guide masking, so organs that become
dominant during training are more likely to be masked and therefore
discouraged from monopolizing the representation. As a result, adipose
saliency is reduced compared with the baseline ($32\%$, $-4\%$), and
more attention is redistributed to the small-footprint organs. 

At convergence, the four small-footprint organs together receive
$27\%$ of the total saliency under SGM, compared with $22\%$ for the
baseline and only $13\%$ at initialization, even though they account for
just $14\%$ of all tokens. 
These results provide direct evidence that SGM mitigates the dominant-organ shortcut
bias by redistributing representation capacity across organs, rather
than merely moving the shortcut from one dominant organ to another.

\begin{table}[h]
\centering
\caption{\textbf{Comparison of saliency proxies for SGM under linear probing.} Mean AUROC across 13 disease groups on the held-out test set, averaged over 10 independent runs (mean $\pm$ std). \textit{Standard} uses the full test set; \textit{Full Organs (7)} and \textit{Drop $k$ Organ(s)} are evaluated on the complete-7-organ subset; \textit{w/o $X$} removes only organ $X$ at test time. Best result per row in \textbf{bold}.}
\label{tab:saliency_proxy}
\setlength{\tabcolsep}{6pt}
\begin{tabular}{lccc}
\toprule
Setting & A1: Last-layer & A3: Rollout \cite{abnar2020quantifying} & \textbf{A2: All-layer avg.\ (Default)} \\
\midrule
Standard               & 0.668 $\pm$ 0.002 & 0.664 $\pm$ 0.002 & \textbf{0.672 $\pm$ 0.002} \\
Full Organs (7)        & 0.685 $\pm$ 0.003 & 0.673 $\pm$ 0.002 & \textbf{0.694 $\pm$ 0.002} \\
Drop 1 Organ           & 0.653 $\pm$ 0.002 & 0.653 $\pm$ 0.001 & \textbf{0.681 $\pm$ 0.002} \\
Drop 2 Organs          & 0.649 $\pm$ 0.002 & 0.647 $\pm$ 0.001 & \textbf{0.672 $\pm$ 0.002} \\
Drop 3 Organs          & 0.622 $\pm$ 0.004 & 0.628 $\pm$ 0.002 & \textbf{0.654 $\pm$ 0.002} \\
Drop 4 Organs          & 0.602 $\pm$ 0.002 & 0.601 $\pm$ 0.002 & \textbf{0.622 $\pm$ 0.002} \\
\midrule
w/o Brain              & 0.663 $\pm$ 0.003 & 0.670 $\pm$ 0.003 & \textbf{0.688 $\pm$ 0.002} \\
w/o Heart              & 0.685 $\pm$ 0.004 & 0.674 $\pm$ 0.003 & \textbf{0.695 $\pm$ 0.002} \\
w/o Adipose            & 0.631 $\pm$ 0.003 & 0.606 $\pm$ 0.003 & \textbf{0.662 $\pm$ 0.002} \\
w/o Liver              & 0.682 $\pm$ 0.002 & 0.670 $\pm$ 0.002 & \textbf{0.684 $\pm$ 0.002} \\
w/o Kidney             & 0.684 $\pm$ 0.003 & 0.671 $\pm$ 0.002 & \textbf{0.695 $\pm$ 0.003} \\
w/o Spleen             & 0.686 $\pm$ 0.003 & 0.680 $\pm$ 0.002 & \textbf{0.695 $\pm$ 0.002} \\
w/o Pancreas           & 0.676 $\pm$ 0.003 & 0.672 $\pm$ 0.002 & \textbf{0.697 $\pm$ 0.002} \\
\bottomrule
\end{tabular}
\end{table}

\subsection{Organ Saliency Proxy for SGM}
\label{sec:saliency_proxy}

The saliency-guided masking (SGM) module relies on a proxy function $\mathcal{A}(\cdot)$ that converts the teacher's attention maps into per-organ importance scores. Different proxies trade off between local sharpness (later layers focus on task-relevant tokens) and global integration (earlier layers preserve broader context). To validate our design choice, we compare three attention proxies for computing the per-token saliency map $\bar{\mathbf{a}} \in \mathbb{R}^{N}$:

\textbf{A1: Last-layer attention.} Only the final block's CLS-to-patch attention is used:
\begin{equation}
\bar{\mathbf{a}}^{\text{A1}} = \frac{1}{H}\sum_{h=1}^{H} \mathbf{a}_{L}^{h}.
\end{equation}
This proxy captures the highest-level semantic focus but discards information from intermediate layers.

\textbf{A2: All-layer average (default).} The CLS-to-patch attention is averaged across all $L$ blocks and $H$ heads, which follows the default formulation in (\ref{eq:average_atten_all_layer}):
\begin{equation}
\bar{\mathbf{a}}^{\text{A2}} = \frac{1}{LH}\sum_{l=1}^{L}\sum_{h=1}^{H} \mathbf{a}_{l}^{h}.
\end{equation}
This balances semantic focus from later layers with broader contextual signal from earlier ones.

\textbf{A3: Attention rollout~\cite{abnar2020quantifying}.} Following~\cite{abnar2020quantifying}, attention rollout recursively multiplies head-averaged attention matrices (with residual connections) across layers to estimate the effective information flow from CLS to each input token:
\begin{equation}
\tilde{\mathbf{A}}_l = 0.5\cdot \bar{\mathbf{A}}_l + 0.5\cdot \mathbf{I}, \qquad \mathbf{R} = \prod_{l=1}^{L}\tilde{\mathbf{A}}_l, \qquad \bar{\mathbf{a}}^{\text{A3}} = \mathbf{R}_{[\texttt{CLS},:]},
\end{equation}
where $\bar{\mathbf{A}}_l$ is the head-averaged attention at layer $l$. Rollout accounts for skip connections and compositional attention paths but can over-smooth saliency in deep networks.

\textbf{Results.} Table~\ref{tab:saliency_proxy} compares the three proxies under linear probing across all 13 evaluation settings, averaged over 10 independent runs. To ensure a fair comparison, we performed an independent grid search of the masking temperature $\tau \in \{0.15, 0.20, 0.25, 0.30, 0.35, 0.40, 0.45, 0.50\}$ for each proxy and report results at the optimal value: $\tau=0.30$ for A1, $\tau=0.25$ for A2, and $\tau=0.35$ for A3. A2 (all-layer average) consistently outperforms both alternatives in every setting, with margins that substantially exceed the run-to-run standard deviations. The gap is most pronounced under severe missingness: at \emph{Drop-3 Organs}, A2 reaches 0.654~AUROC versus 0.622 for A1 and 0.628 for A3 ($+3.2\%$ and $+2.6\%$ respectively), while at \emph{w/o Adipose} the gap reaches $+3.1\%$ over A1 and $+5.6\%$ over A3.

We attribute A1's underperformance to its narrow reliance on final-layer attention, where the teacher's [CLS] focus is already biased toward dominant organs that the model has learned to over-rely on—precisely the shortcut that SGM aims to disrupt. A3 partially mitigates this through layer-wise integration, but the iterative matrix multiplication in attention rollout~\cite{abnar2020quantifying} amplifies low-magnitude noise across our 12-layer ViT, leading to overly diffuse organ saliency estimates and even worse performance than A1 on several settings (e.g., \emph{w/o Adipose}). A2's simple averaging strikes the best balance: it aggregates signal across all layers without compounding noise, providing both stability (low variance across runs) and discriminative organ-level saliency.

\begin{figure}[t]
\begin{center}
   \includegraphics[width=1.0\linewidth]{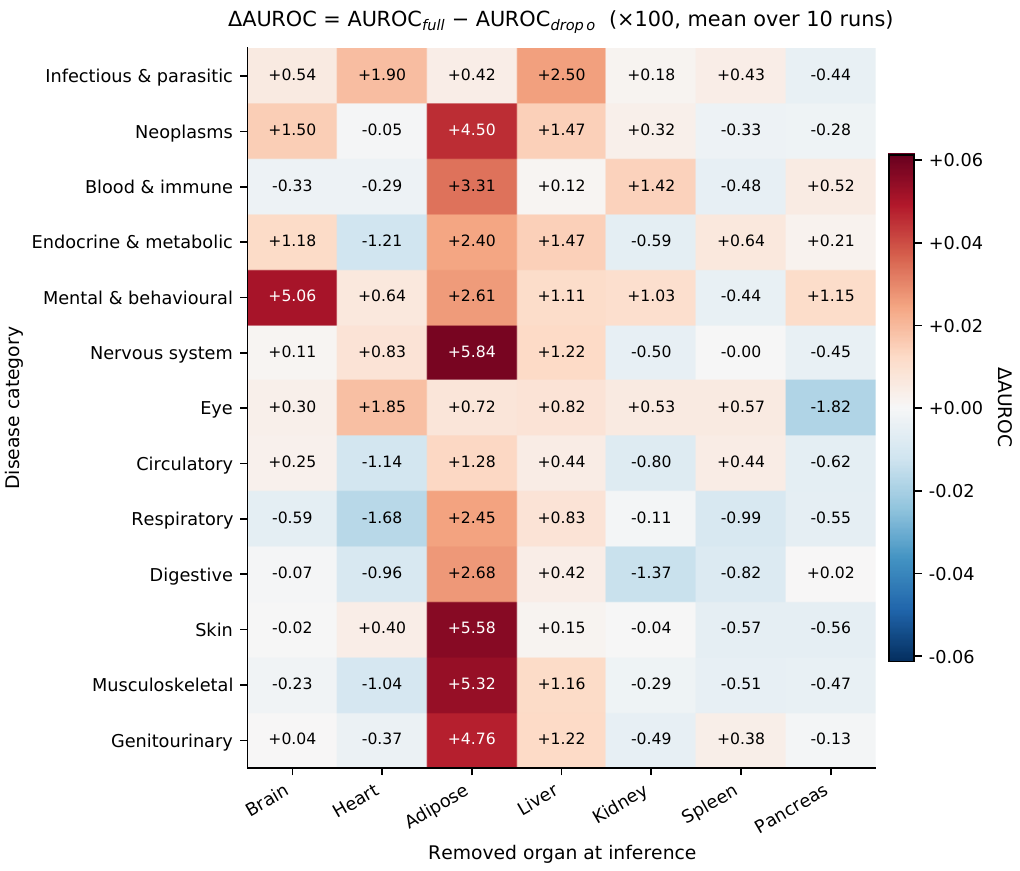} 
\end{center}
 \caption{\textbf{Leave-one-organ-out ablation}. Each cell reports $\Delta\text{AUROC} \times 100 = 100 \times (\text{AUROC}_{\text{full}} - \text{AUROC}_{\text{drop } o})$, averaged over 10 linear-probe training runs on top of the frozen pre-trained backbone. Probes are trained on the full training set and evaluated on the 7-organ-complete test subset. Positive values (red) indicate organ $o$ contributes to a specific disease $d$.}
 \label{fig:organ_removal_per_case}
\end{figure}

\subsection{Organ Importance Analysis}
\label{sec:organ_import_analy}
Fig.~\ref{fig:organ_removal_per_case} reveals several interesting patterns. First, \emph{adipose} features are near-universal contributors. Removing them degrades performance on the majority of disease categories, with the largest drops on nervous system ($\Delta = +5.84$), skin ($+5.58$), musculoskeletal ($+5.32$), and genitourinary ($+4.76$) diseases. This is consistent with the role of body composition as a systemic marker of metabolic and inflammatory state. Second, brain features are highly disease-specific. Their contribution concentrates on mental \& behavioural ($+5.06$) and, to a smaller extent, neoplasms ($+1.50$) and endocrine \& metabolic ($+1.18$) disorders. Third, the remaining organs show clear disease-specific specialisation. Heart contributes most to eye ($+1.85$) and infectious \& parasitic diseases ($+1.90$), reflecting the cardiovascular footprint of systemic infection and microvascular eye conditions; liver is the strongest contributor for infectious \& parasitic ($+2.50$) and a consistent secondary signal across endocrine, neoplastic, and metabolic categories ($+1.16$ to $+1.47$), in line with its central role in immune and metabolic homeostasis; kidney plays a notable role in blood \& immune disease ($+1.42$) and mental \& behavioural disorders ($+1.03$); spleen and pancreas contribute meaningfully to endocrine \& metabolic ($+0.64$, $+0.21$) and mental \& behavioural ($+1.15$) diseases.Overall, these results indicate that our pre-trained representation learns disease-specific reliance on specific organs rather than treating modalities uniformly.

\subsection{Effect of Pre-training Loss Components}
\label{sec:loss_ablation}

The full Pan-FM training objective combines two loss components: the DINO consistency loss $\mathcal{L}_{\text{DINO}}$ on the [CLS] token and the KoLeo regularization loss $\mathcal{L}_{\text{KoLeo}}$~\cite{dinov2}. To isolate the contribution of each, we re-train Pan-FM under two reduced objectives and compare against the full model: 1) {DINO only}: $\mathcal{L} = \mathcal{L}_{\text{DINO}}$, 2) {KoLeo only}: $\mathcal{L} = \mathcal{L}_{\text{KoLeo}}$, and 3) {Full (default)}: $\mathcal{L} = \mathcal{L}_{\text{DINO}} + \lambda \mathcal{L}_{\text{KoLeo}}$ 
All other configurations (backbone, optimizer, SGM, masking ratio, training epochs) are held fixed.

\textbf{Results.} Table~\ref{tab:loss_ablation} reports linear-probing AUROC across all 13 evaluation settings. Three observations emerge. \textbf{(i)} \emph{KoLeo alone fails to learn meaningful representations}, achieving only 0.551 mean AUROC, essentially at chance level on hardest settings (e.g., 0.521 at Drop-4). This is because KoLeo provides no supervisory target and merely pushes features apart on the unit sphere. \textbf{(ii)} \emph{DINO alone learns a competent representation} (0.624 mean AUROC), confirming that the teacher–student consistency objective is the primary driver of feature quality, but it lags the full model by 5.4 points on average. \textbf{(iii)} \emph{Adding KoLeo as a regularizer to DINO substantially improves performance} across every setting, with gains ranging from $+4.1\%$ (Standard) to $+5.3\%$ (Drop-3) and reaching $+7.0\%$ on \emph{w/o Heart}. We attribute this to KoLeo's role as an anti-collapse regularizer: by encouraging uniform feature spread on the embedding sphere, it prevents the model from collapsing onto a few dominant organs and complements DINO's invariance objective. The two losses are complementary: DINO encourages discriminative representations, while KoLeo promotes a well-distributed feature space; together, they yield stronger representations than either loss alone.

\begin{table}[h]
\centering
\caption{\textbf{Effect of loss components on Pan-FM under linear probing.} Mean AUROC across 13 disease groups on the held-out test set. \emph{Standard} uses the full test set; \emph{Full Organs (7)} and \emph{Drop $k$ Organ(s)} are evaluated on the complete-7-organ subset; \emph{w/o $X$} removes only organ $X$ at test time. Best result per row in \textbf{bold}.}
\label{tab:loss_ablation}
\vspace{-0.15cm}
\setlength{\tabcolsep}{8pt}
\begin{tabular}{lccc}
\toprule
Setting & KoLeo only & DINO only & \textbf{DINO + KoLeo (Default)} \\
\midrule
Standard               & 0.561 & 0.624 & \textbf{0.672} \\
Full Organs (7)        & 0.543 & 0.636 & \textbf{0.694} \\
Drop 1 Organ           & 0.564 & 0.628 & \textbf{0.681} \\
Drop 2 Organs          & 0.538 & 0.616 & \textbf{0.672} \\
Drop 3 Organs          & 0.525 & 0.603 & \textbf{0.654} \\
Drop 4 Organs          & 0.522 & 0.606 & \textbf{0.622} \\
\midrule
w/o Brain              & 0.559 & 0.611 & \textbf{0.688} \\
w/o Heart              & 0.546 & 0.626 & \textbf{0.695} \\
w/o Adipose            & 0.527 & 0.614 & \textbf{0.662} \\
w/o Liver              & 0.529 & 0.637 & \textbf{0.684} \\
w/o Kidney             & 0.579 & 0.642 & \textbf{0.695} \\
w/o Spleen             & 0.568 & 0.636 & \textbf{0.695} \\
w/o Pancreas           & 0.597 & 0.637 & \textbf{0.697} \\
\midrule
\textbf{Mean}          & 0.551 & 0.624 & \textbf{0.678} \\
\bottomrule
\end{tabular}
\end{table}

\subsection{Downstream Training Data Ratio}
\label{sec:training_data_ratio}
Table~\ref{tab:table_train_data_ratio} reports linear-probing performance across five label fractions (5\%-75\%) under both standard and missing-organ evaluation protocols. For training with various data ratio, we construct training sets by sampling both \emph{disease cases} and \emph{cognitively normal (CN)} subjects according to predefined label fractions. A linear classifier is trained with these sampled data on frozen backbone features. To ensure robustness, each experimental variant is repeated for 5 runs with different random seeds, and the average performance is reported. Evaluation is performed on a fixed held-out test set for fair comparison across all methods and data regimes. Pan-FM achieves consistent performance gains over DINOv2 across all downstream training data ratios under the linear probing setting. The improvements are particularly pronounced in the low-label regime, suggesting superior data efficiency and stronger representation quality. As the amount of labeled data increases, Pan-FM continues to maintain its advantage, demonstrating that its learned representations generalize well across both scarce and abundant supervision scenarios.

\begin{table}[ht]
\centering
\caption{Comprehensive performance comparison (Mean AUROC) across five downstream training data ratios (5\%, 10\%, 25\%, 50\%, 75\%) over 5 runs under linear probing. Bold values indicate the better performance between \textbf{DINOv2} and \textbf{Pan-FM} under each setting.}
\label{tab:table_train_data_ratio}
\resizebox{\textwidth}{!}{
\begin{tabular}{l c c c c c c c c c c}
\hline
\multirow{2}{*}{{Evaluation Setting}} 
& \multicolumn{2}{c}{{5\% Ratio}} 
& \multicolumn{2}{c}{{10\% Ratio}} 
& \multicolumn{2}{c}{{25\% Ratio}} 
& \multicolumn{2}{c}{{50\% Ratio}} 
& \multicolumn{2}{c}{{75\% Ratio}} \\ 
\cline{2-11}
& {DINOv2} & {Pan-FM} 
& {DINOv2} & {Pan-FM} 
& {DINOv2} & {Pan-FM} 
& {DINOv2} & {Pan-FM} 
& {DINOv2} & {Pan-FM} \\ 
\hline

\textit{Standard Protocols} & & & & & & & & & & \\
Standard
& 0.564 & \textbf{0.580} 
& 0.581 & \textbf{0.598} 
& 0.614 & \textbf{0.628} 
& 0.635 & \textbf{0.651} 
& 0.647 & \textbf{0.663} \\

Full Organs (7)
& 0.586 & \textbf{0.606} 
& 0.615 & \textbf{0.629} 
& 0.649 & \textbf{0.657} 
& 0.668 & \textbf{0.679} 
& 0.682 & \textbf{0.688} \\

\hline
\textit{Random Organ Dropout} & & & & & & & & & & \\
Drop 1 Organ 
& 0.573 & \textbf{0.587} 
& 0.588 & \textbf{0.620} 
& 0.608 & \textbf{0.644} 
& 0.640 & \textbf{0.659} 
& 0.655 & \textbf{0.677} \\

Drop 2 Organs 
& 0.578 & \textbf{0.581} 
& 0.594 & \textbf{0.606} 
& 0.614 & \textbf{0.633} 
& 0.638 & \textbf{0.653} 
& 0.647 & \textbf{0.667} \\

Drop 3 Organs 
& 0.568 & \textbf{0.576} 
& 0.579 & \textbf{0.595} 
& 0.593 & \textbf{0.622} 
& 0.609 & \textbf{0.637} 
& 0.617 & \textbf{0.648} \\

Drop 4 Organs 
& 0.561 & \textbf{0.564} 
& 0.561 & \textbf{0.582} 
& 0.574 & \textbf{0.600} 
& 0.595 & \textbf{0.606} 
& 0.597 & \textbf{0.618} \\

\hline
\textit{Specific Organ Dropout} & & & & & & & & & & \\
w/o Brain 
& 0.585 & \textbf{0.585} 
& 0.592 & \textbf{0.629} 
& 0.628 & \textbf{0.651} 
& 0.665 & \textbf{0.668} 
& 0.684 & \textbf{0.683} \\

w/o Heart 
& 0.592 & \textbf{0.606} 
& 0.602 & \textbf{0.632} 
& 0.632 & \textbf{0.663} 
& 0.659 & \textbf{0.680} 
& 0.684 & \textbf{0.691} \\

w/o Adipose 
& 0.571 & \textbf{0.576} 
& 0.585 & \textbf{0.611} 
& 0.607 & \textbf{0.614} 
& 0.620 & \textbf{0.645} 
& 0.626 & \textbf{0.660} \\

w/o Liver 
& 0.586 & \textbf{0.605} 
& 0.608 & \textbf{0.631} 
& 0.632 & \textbf{0.653} 
& 0.665 & \textbf{0.679} 
& 0.679 & \textbf{0.690} \\

w/o Kidney 
& 0.586 & \textbf{0.599} 
& 0.613 & \textbf{0.631} 
& 0.641 & \textbf{0.661} 
& 0.673 & \textbf{0.690} 
& 0.688 & \textbf{0.698} \\

w/o Spleen 
& 0.590 & \textbf{0.597} 
& 0.607 & \textbf{0.632} 
& 0.639 & \textbf{0.664} 
& 0.679 & \textbf{0.690} 
& 0.697 & \textbf{0.699} \\

w/o Pancreas 
& 0.583 & \textbf{0.600} 
& 0.604 & \textbf{0.634} 
& 0.631 & \textbf{0.665} 
& 0.661 & \textbf{0.691} 
& \textbf{0.677} & 0.676 \\

\hline
\end{tabular}
}
\end{table}

\section{Limitations and Future Work}
\label{sec:limitations}

Pan-FM is the first foundation model to jointly learn cross-organ representations across seven organ systems under realistic missing-organ scenarios, but the present work has several limitations that we acknowledge and view as directions for future research. First, our experiments are designed primarily as a \emph{conceptualization} of multi-organ representation learning rather than as a clinical-grade prediction system: we operate on low-dimensional, pre-extracted MRI-derived imaging phenotypes (IDPs) instead of raw voxel data, which constrains the representational capacity of the backbone and leaves direct pre-training on raw multi-organ imaging as a natural next step. Second, the UK Biobank is a general-population cohort whose disease labels are derived from inpatient records and ICD codes rather than disease-specific phenotyping, so absolute predictive performance is inherently modest. Our goal is to demonstrate the effectiveness of Pan-FM in handling realistic missing-organ scenarios and in improving over baseline approaches, rather than to establish state-of-the-art disease prediction. Validating Pan-FM on disease-specific cohorts (e.g., ADNI \cite{petersen2010alzheimer} for neurodegeneration) is therefore an important next step. Finally, the UK Biobank is predominantly white-British and middle-aged-to-elderly. Generalization to under-represented populations and other healthcare systems requires further validation.

\end{document}